\documentclass[10pt,twocolumn,letterpaper]{article}

\usepackage[table,xcdraw,dvipsnames]{xcolor}

\usepackage[final]{iccv}

\usepackage{multirow}
\usepackage{graphicx}
\usepackage{amsmath}
\usepackage{amssymb}
\usepackage{booktabs}
\usepackage{dashrule}
\usepackage{array}
\usepackage{makecell}
\usepackage{pifont}
\usepackage{placeins}
\usepackage{balance}
\usepackage{xr}  
\newcolumntype{P}[1]{>{\centering\arraybackslash}p{#1}}

\usepackage{gensymb}

\usepackage{float}
\usepackage{adjustbox}
\usepackage{longtable}

\usepackage{subcaption}
\usepackage{siunitx}

\usepackage{lipsum}

\newcommand\blfootnote[1]{%
  \begingroup
  \renewcommand\thefootnote{}\footnote{#1}%
  \addtocounter{footnote}{-1}%
  \endgroup
}

\def\paperID{6548} 
\def\confName{ICCV}
\def\confYear{2025}


\definecolor{cvprblue}{rgb}{0.21,0.49,0.74}

\definecolor{textgreen}{rgb}{0.4980392156862745, 0.788235294117647, 0.4980392156862745}
\definecolor{ourmediumblue}{rgb}{0.21568627450980393,0.49411764705882355,0.7215686274509804}
\definecolor{ourmediumred}{rgb}{0.8941176470588236,0.10196078431372549,0.10980392156862745}
\definecolor{ourmediumgreen}{rgb}{0.30196078431372547,0.6862745098039216,0.2901960784313726}

\definecolor{ourlightred}{rgb}{0.984313725490196, 0.5019607843137255, 0.4470588235294118}
\definecolor{ourlightblue}{rgb}{0.5019607843137255, 0.6941176470588235, 0.8274509803921568}
\definecolor{ourlightcyan}{rgb}{0.5529411764705883, 0.8274509803921568, 0.7803921568627451}

\definecolor{darkspringgreen}{rgb}{0.09, 0.45, 0.27}

\definecolor{cornellred}{rgb}{0.7, 0.11, 0.11}
\definecolor{darkcerulean}{rgb}{0.03, 0.27, 0.49}
\definecolor{amaranth}{rgb}{0.9, 0.17, 0.31}
\definecolor{americanrose}{rgb}{1.0, 0.01, 0.24}

\definecolor{blue(ryb)}{rgb}{0.01, 0.28, 1.0}
\definecolor{cadmiumred}{rgb}{0.89, 0.0, 0.13}
\definecolor{darkorange}{rgb}{1.0, 0.55, 0.0}
\definecolor{flame}{rgb}{0.89, 0.35, 0.13}

\newcommand{\txtred}[1]{\textcolor{amaranth}{#1}}
\newcommand{\txtgreen}[1]{\textcolor{textgreen}{#1}}
\newcommand{\txtblue}[1]{\textcolor{ourmediumblue}{#1}}

\newcommand{\txtdeepred}[1]{\textcolor{cornellred}{#1}}
\newcommand{\txtdeepblue}[1]{\textcolor{darkcerulean}{#1}}
\newcommand{\txtorange}[1]{\textcolor{flame}{#1}}

\newcommand{\checkmarkgreen}[1]{\textcolor{darkspringgreen}{#1}}

\newcommand{\datasetname}{ROADWork~}
\newcommand{\emphdatasetname}{ROADWork~}

\newcommand{\mycheckmark}[0]{\checkmarkgreen{\ding{52}}}

\newcommand\extrafootertext[1]{%
    \bgroup
    \renewcommand\thefootnote{\fnsymbol{footnote}}%
    \renewcommand\thempfootnote{\fnsymbol{mpfootnote}}%
    \footnotetext[0]{#1}%
    \egroup
}

\usepackage[pagebackref,breaklinks,colorlinks,citecolor=cvprblue]{hyperref}

\newcommand\red[1]{\textcolor{red}{#1}}
\newcommand{\norm}[1]{\left\lVert#1\right\rVert}

\newcommand{\pointToAppendix}[2]{Appendix #1}
\newcommand{\pointToMain}[2]{#2}

\begin{document}

\title{\textbf{\txtdeepred{ROADWork}}: A Dataset and Benchmark for Learning to \\ \txtdeepred{R}ecognize, \txtdeepred{O}bserve, \txtdeepred{A}nalyze and \txtdeepred{D}rive Through \textbf{\txtdeepred{Work}} Zones}


\author{
Anurag Ghosh \quad Shen Zheng \quad Robert Tamburo \quad Khiem Vuong \quad Juan Alvarez-Padilla \\ \quad Hailiang Zhu$^{*}$ \quad Michael Cardei$^{*}$  \quad Nicholas Dunn$^{*}$  \quad Christoph Mertz \quad Srinivasa G. Narasimhan \\
Carnegie Mellon University \\
\url{https://www.cs.cmu.edu/~roadwork/}
}

\maketitle
\blfootnote{$^{*}$Equal contribution.Work done at CMU.}

\begin{abstract}
Perceiving and autonomously navigating through work zones is a challenging and underexplored problem. Open datasets for this long-tailed scenario are scarce. We propose the \emphdatasetname dataset to learn to recognize, observe, analyze, and drive through work zones. State-of-the-art foundation models fail when applied to work zones. Fine-tuning models on our dataset significantly improves perception and navigation in work zones. With \emphdatasetname, we discover new work zone images with higher precision (\textbf{\txtgreen{+32.5\%}}) at a much higher rate (\textbf{\txtgreen{12.8$\times$}}) around the world. Open-vocabulary methods fail too, whereas fine-tuned detectors improve performance \textbf{(\textbf{\txtgreen{+32.2 AP}})}.
Vision-Language Models (VLMs) struggle to describe work zones, but fine-tuning substantially improves performance (\textbf{\txtgreen{+36.7 SPICE}}). 

Beyond fine-tuning, we show the value of simple techniques. Video label propagation provides additional gains (\textbf{\txtgreen{+2.6 AP}}) for instance segmentation. While reading work zone signs, composing a detector and text spotter via crop-scaling improves performance (\textbf{\txtgreen{+14.2\% 1-NED}}).  Composing work zone detections to provide context further reduces hallucinations (\textbf{\txtgreen{+3.9 SPICE}}) in VLMs. We predict navigational goals and compute drivable paths from work zone videos. Incorporating road work semantics ensures 53.6\% goals have angular error (AE) $< \ang{0.5}$ (\textbf{\txtgreen{+9.9 \%}}) and 75.3\% pathways have $AE < \ang{0.5}$ (\textbf{\txtgreen{+8.1 \%}}).
\vspace{-0.1in}
\end{abstract}
\vspace{-0.2in}
\section{Introduction}
\vspace{-0.05in}
\label{sec:intro}

\begin{figure}[t]
    \vspace{-0.25in}
    \centering 
    \includegraphics[width=\linewidth]{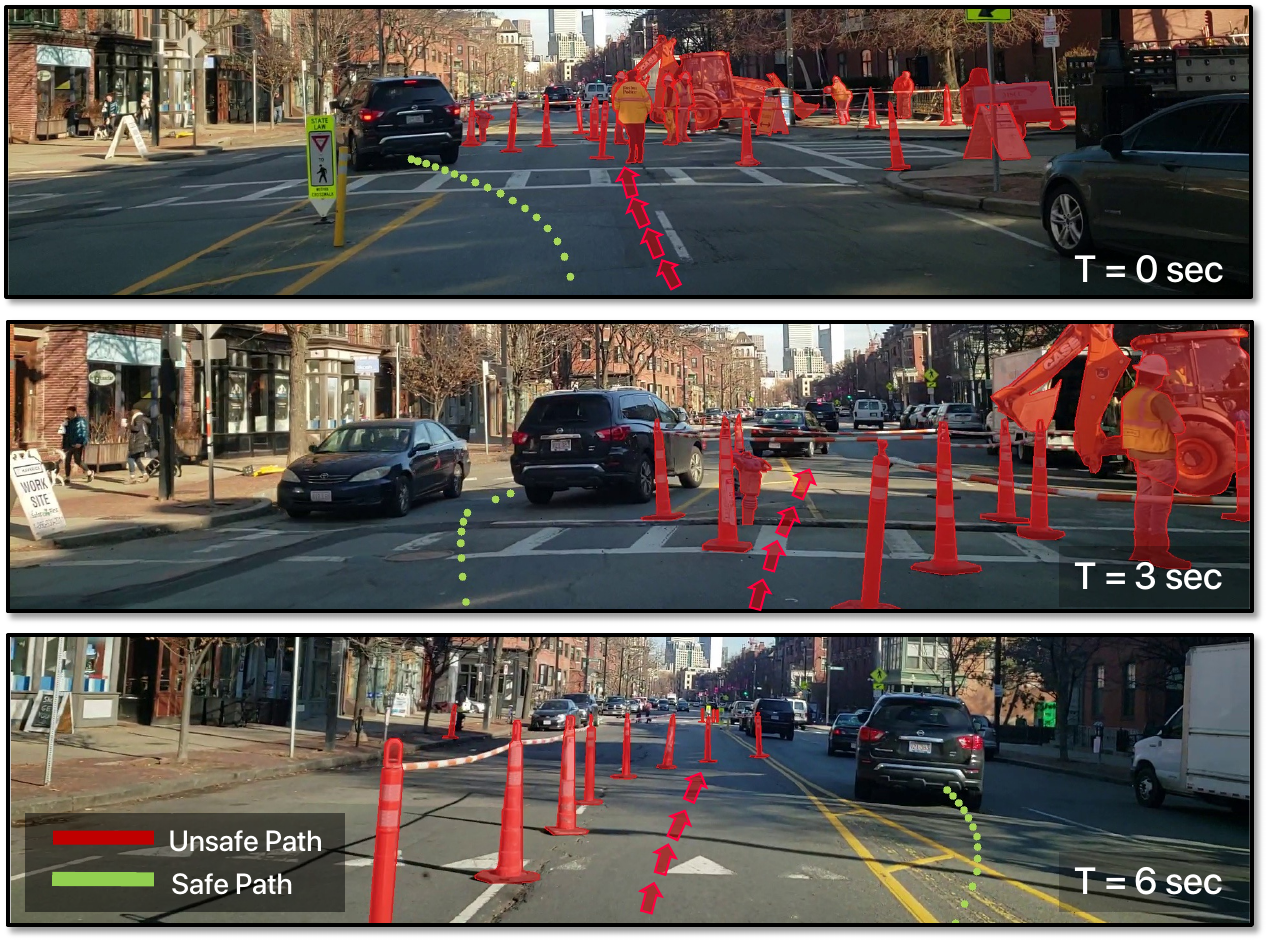}
    \vspace{-0.15in}
    \captionof{figure}{\small \textbf{Autonomous Driving In Work Zones.} Work zone objects block the road (in red), making some road regions unsafe (red arrows), necessitating the development of forecasting traversable paths from driven paths (green dots). Prior datasets do not comprehensively address this long-tailed challenge.
    Foundation models struggle with recognizing and interpreting work zone objects and signs, discovering new work zones, and analyzing work zones. We additionally formulate work zone navigation as a learnable task. The \emphdatasetname dataset and benchmark highlight key challenges in work zone perception and navigation, and we demonstrate simple techniques that improve performance on these tasks.} 
    \label{fig:teaser}
    \vspace{-0.2in}
\end{figure}

Navigating safely through work zones is a significant challenge for both autonomous and human-driven vehicles. In recent years, several reports suggest that self-driving cars are stumped by the simplest work zones~\cite{SelfDrivingChallenge2, news-wired2017construction, news-atlantic2023sf, news-pntv2023construction, news-cnn2021cones, news-nyt2023concrete, news-sfstandard2023hole, news-wired2022cruise, news-2024missionlocal}. In fact, even humans find driving through work zones challenging~\cite{Thapa2024, Morgan2010, Wang1996}. Since 2010, more than 700 people have died every year in the United States in accidents involving work zones~\cite{NSCInjuryFacts2023, NHTSAPeopleStats2021}. Although road work zones are relatively rare, learning to navigate them is essential for full vehicular autonomy. 

\begin{table*}[t!]
\centering
\resizebox{0.9\textwidth}{!}{
\begin{tabular}{lclccccccc}
\hline
  Datasets &
  Type &
  Dataset Size &
  \begin{tabular}[c]{@{}c@{}}Work Zone \\ Related Objects\end{tabular} &
  \begin{tabular}[c]{@{}c@{}}Number of \\ Cities\end{tabular} &
  \begin{tabular}[c]{@{}c@{}}Pixel-Level \\ Annotations\end{tabular} &
  \begin{tabular}[c]{@{}c@{}}Object-Level \\ Annotations\end{tabular} &
  \begin{tabular}[c]{@{}c@{}}Fine-grained \\ Annotations\end{tabular} &
  \begin{tabular}[c]{@{}c@{}}Scene \\ Descriptions\end{tabular} &
  \begin{tabular}[c]{@{}c@{}}Driven \\ Pathway\end{tabular} \\ \hline
 &
   &
  10K Images &
  None &
  4 US Cities &
  \mycheckmark &
   &
   &
   &
   \\
\multirow{-2}{*}{BDD100k~\cite{yu2020bdd100k}} &
  \multirow{-2}{*}{C} &
  100K Images &
  None &
   &
   &
  \mycheckmark &
   &
   &
   \\ \hline
NuScenes~\cite{nuscenes} &
  C &
  1000 Scenes (40K Frames) &
  5 Object Categories &
  Boston and Singapore &
   &
  \mycheckmark &
   &
   &
  \mycheckmark \\
Waymo~\cite{Waymodataset} &
  C &
  1150 Scenes &
  None &
  3 US Cities &
   &
  \mycheckmark &
   &
   &
  \mycheckmark \\
Mapillary~\cite{neuhold2017mapillary} &
  C &
  25K Images &
  2 Object Categories &
  Around The World &
  \mycheckmark &
  \mycheckmark &
   &
   &
   \\
Cityscapes~\cite{cordts2016cityscapes} &
  C &
  5000 Images &
  1 Object Category &
  27 Cities in Germany &
  \mycheckmark &
  \mycheckmark &
   &
   &
   \\ \hline
WildDash~\cite{zendel2018wilddash} &
  LT &
  1800 Images &
  None &
  Around The World &
  \mycheckmark &
   &
   &
   &
   \\
WildDash 2~\cite{zendel2022wilddash2} &
  LT &
  5000 Images &
  None &
  Around The World &
  \mycheckmark &
   &
   &
   &
   \\
 &
   &
   &
   &
  NuScenes, Karlsruhe &
   &
   &
   &
   &
   \\
\multirow{-2}{*}{CODA~\cite{li2022coda}} &
  \multirow{-2}{*}{LT} &
  \multirow{-2}{*}{1500 Scenes (5937 Images)} &
  \multirow{-2}{*}{6 Object Categories} &
  and few cities in China &
  \multicolumn{1}{l}{} &
  \multirow{-2}{*}{\mycheckmark} &
  \multicolumn{1}{l}{} &
  \multicolumn{1}{l}{} &
  \multicolumn{1}{l}{} \\ \hline
SegmentMeIfYouCan~\cite{chan2021segmentmeifyoucan} &
  AN &
  327 Images (Eval) &
  None &
  Unknown &
  \mycheckmark &
   &
   &
   &
   \\
BDD-Anomaly~\cite{hendrycks2019bddanomaly} &
  AN &
  810 Images (Eval) &
  None &
  Data from BDD100K &
  \mycheckmark &
   &
   &
   &
   \\ \hline
\rowcolor[HTML]{ECF4FF} 
\cellcolor[HTML]{ECF4FF} &
  \cellcolor[HTML]{ECF4FF} &
  4375 Videos, &
  15 Object Categories &
  18 US Cities &
  \cellcolor[HTML]{ECF4FF} &
  \cellcolor[HTML]{ECF4FF} &
  \cellcolor[HTML]{ECF4FF} &
  \cellcolor[HTML]{ECF4FF} &
  \cellcolor[HTML]{ECF4FF} \\
\rowcolor[HTML]{ECF4FF} 
\cellcolor[HTML]{ECF4FF} &
  \cellcolor[HTML]{ECF4FF} &
  9650 Images, &
  360 Unique Signs &
  In-The-Wild Images from &
  \cellcolor[HTML]{ECF4FF} &
  \cellcolor[HTML]{ECF4FF} &
  \cellcolor[HTML]{ECF4FF} &
  \cellcolor[HTML]{ECF4FF} &
  \cellcolor[HTML]{ECF4FF} \\
\rowcolor[HTML]{ECF4FF} 
\multirow{-3}{*}{\cellcolor[HTML]{ECF4FF}\textbf{\emphdatasetname (Ours)}} &
  \multirow{-3}{*}{\cellcolor[HTML]{ECF4FF}LT} &
  129K Images (Path.) &
   &
  Around The World &
  \multirow{-3}{*}{\cellcolor[HTML]{ECF4FF}\mycheckmark} &
  \multirow{-3}{*}{\cellcolor[HTML]{ECF4FF}\mycheckmark} &
  \multirow{-3}{*}{\cellcolor[HTML]{ECF4FF}\mycheckmark} &
  \multirow{-3}{*}{\cellcolor[HTML]{ECF4FF}\mycheckmark} &
  \multirow{-3}{*}{\cellcolor[HTML]{ECF4FF}\mycheckmark} \\ \hline
\end{tabular}%
}
\vspace{-0.1in}
\captionof{table}{\small \textbf{Comparing \emphdatasetname With Other Driving Datasets/Benchmarks.} Compared to common (\textit{C}) datasets~\cite{yu2020bdd100k, nuscenes, neuhold2017mapillary, cordts2016cityscapes}, \emphdatasetname has a wider variety of annotations for a rare driving scenario while being similar in size. Compared to long-tailed (\textit{LT}) and anomalous (\textit{AN}) datasets~\cite{zendel2018wilddash, zendel2022wilddash2, li2022coda, chan2021segmentmeifyoucan, hendrycks2019bddanomaly}, \emphdatasetname is larger with  rich pixel, object and scene-level annotations across a broader range of tasks. Others focus on one or two aspects of self-driving but \emphdatasetname holistically covers work zone perception and navigation. }
\label{tbl:dataset-comparison}
\vspace{-0.25in}
\end{table*}

Despite the research potential, few studies focus on this domain. Work zones fall into the long-tail of scene distributions for vehicular autonomy. Mining data for such long-tail situations is expensive, difficult, and labor-intensive~\cite{liang2024aide}. Even large data sources contain few informative and diverse long-tailed samples. Thus, such data and methods are often trade secrets~\cite{tesla-patent, cruise-long-tail}. For perspective, in this work, we semi-automatically filtered only 5078 keyframes (less than 0.01\%) from nearly 45 million frames in one data source.


Both common~\cite{yu2020bdd100k, neuhold2017mapillary, cordts2016cityscapes} and long-tailed datasets~\cite{li2022coda, zendel2018wilddash, zendel2022wilddash2} contain few work zones. Among common datasets, our analysis estimates that BDD100K~\cite{yu2020bdd100k} and Mapillary~\cite{neuhold2017mapillary} collectively contain fewer than 1000 work zone images. Although long-tailed datasets such as CODA~\cite{li2022coda} contain bounding boxes for conventional work zone objects such as cones and work vehicles, they do not cover all work zone objects and do not provide any instance segmentations, fine-grained object attributes, scene descriptions, or drivable paths. Others such as WildDash~\cite{zendel2018wilddash, zendel2022wilddash2} contain images with heavy weather, lighting and geographic variations but largely annotate common Cityscapes~\cite{cordts2016cityscapes} categories. Previous works contain little research on work zones, focusing mainly on individual issues and edge cases~\cite{mathibela2013roadwork, gumpp2009recognition, graf2012probabilistic, wimmer2009automatic, pannen2020keep, ferguson2015construction, ferguson2015mapping, kim2024rosa, shi2021work}.

Because work zones are neglected and under-represented in existing datasets, even after training with massive amounts of data and compute, foundation models such as open vocabulary detectors~\cite{zhou2022detecting, wang2023detecting, liu2024grounding}, promptable segmentation models~\cite{kirillov2023segany, ravi2024sam2}, scene text models~\cite{ronen2022glass} and vision language models~\cite{li2022blip, liu2023llava, li2024llavanext-strong}, do not perform well in this scenario. The key to improving self-driving in work zones lies in obtaining work zone driving data - even simple baselines such as Mask R-CNN~\cite{he2017mask} outperform foundation models~\cite{zhou2022detecting, wang2023detecting} when trained on a well-curated dataset. 

\begin{figure*}[t!]
\centering
\includegraphics[width=1\linewidth]{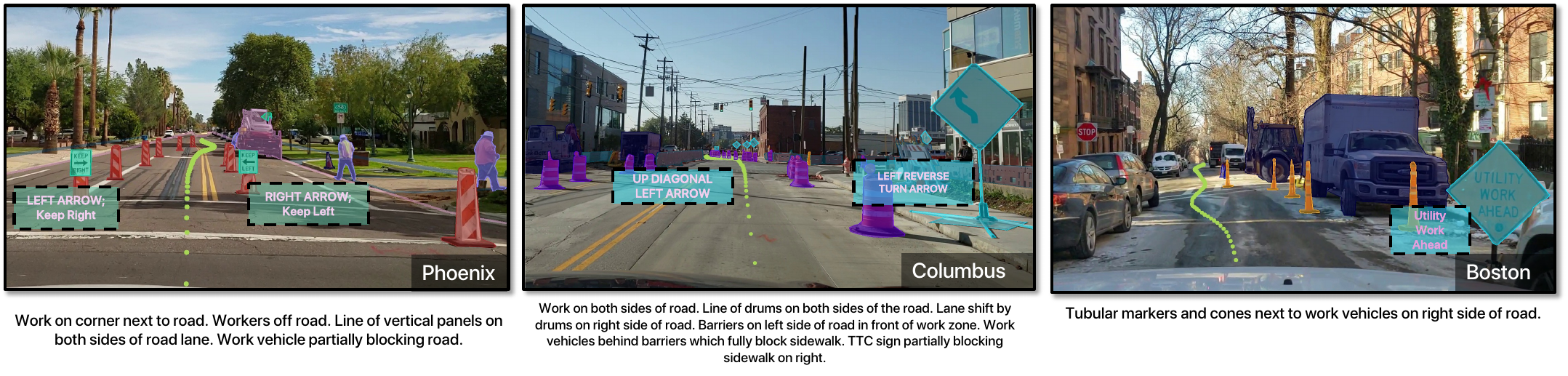}
\vspace{-0.3in}
\captionof{figure}{\small \textbf{The \datasetname Dataset} consists of work zone videos and images. We have segmented 15 object categories such as workers, vehicles and barriers. We provide object attributes for signs and arrow boards to enable fine-grained understanding. Our work zone scene descriptions analyze the scene globally and one passable trajectory automatically estimated from the associated video sequences learns how to drive through work zones. See \pointToAppendix{\ref{sup:dataset}}{B} for more details.}
\label{fig:roadwork-zone-examples}
\vspace{-0.2in}
\end{figure*}

Thus, a concerted and structured approach is needed to address this challenge. We present a novel \emphdatasetname dataset and benchmark for addressing self-driving in work zones. In total, our dataset contains 4375 thirty-second videos, 9650 richly annotated keyframes, and 129017 images with automatically computed pathways from more than 5000 work zones. We cover 18 US cities and in-the-wild images from around the world. Our dataset has 15 types of objects, 360 types of Temporary Traffic Control (TTC) signs and boards, scene descriptions and 2D/3D traversable paths (Figure~\ref{fig:roadwork-zone-examples}).  Table~\ref{tbl:dataset-comparison} provides a comparison with other datasets. \emphdatasetname contains richer annotations for a greater number of tasks, is larger than other long-tailed datasets, and covers many aspects of perception and navigation in work zones. To the best of our knowledge, no large-scale public dataset targets work zones. For detailed related work, see \pointToAppendix{\ref{sup:related_work}}{A}.

\noindent
\textbf{What kind of data is needed for driving in work zones?} No two work zones are truly alike, with unique arrangements of barriers, cones, and signs. Navigating these work zones requires context-specific understanding of atypical object configurations that standard driving datasets fail to capture. We propose that understanding and navigating work zones mirrors human cognition, where seeing differs from observing, and knowing differs from understanding~\cite{su2011glancing}. We address: \textbf{\txtdeepred{Recognizing}} work zones and their constituent objects, \textbf{\txtdeepred{Observing}} fine-grained details (e.g. sign text) that provide navigation instructions, \textbf{\txtdeepred{Analyzing}} global scene context to capture spatial relationships, and \textbf{\txtdeepred{Driving}} through work zones via pathway prediction.

\noindent
Our contributions are summarized below:
\begin{itemize}
    \item We mine thousands of hours of driving videos to curate the \emphdatasetname dataset containing 4375 videos, 9650 richly annotated images, and 129017 images with automatically computed pathways from over 5000 work zones across 18 US cities and around the world. We formulate major challenges as problems reflecting human cognition: recognizing work zones and their objects, observing fine-grained details such as sign text, analyzing global scene context, and driving safely through these zones.

    \item Foundation models perform poorly on work zone tasks. For example, open vocabulary segmentation models achieve only \textbf{\txtred{2.9-4.2 $AP$}}. We dramatically improve perception and navigation with our data (\textbf{\txtgreen{39.0 $AP$}}).

    \item Beyond fine-tuning, we show the value of simple techniques: (a) video label propagation improves detection (\textbf{\txtgreen{+2.8 $AP$}}), (b) crop-rescaling improves reading sign texts (\textbf{\txtgreen{+14.2\% 1-NED}}), (c) composing work zone detections with VLMs reduces hallucinations in scene descriptions (\textbf{\txtgreen{+3.9 SPICE}}), and (d) incorporating work zone semantics improves pathway prediction (\textbf{\txtgreen{+9.9\%}} in goals with angular error $< \ang{0.5}$). Fine-tuning foundation models on the \emphdatasetname dataset does not hurt performance on common driving situations allowing label unification. The \emphdatasetname dataset enables studying other long-tailed  scenarios such as geographic domain adaptation.

\end{itemize}

\section{\textbf{\txtdeepred{ROADWork}} Dataset}
\label{sec:dataset}

\vspace{-0.05in}

\noindent
The \emphdatasetname dataset contains road work navigation videos and images with pixel-level annotations, object-level annotations, fine-grained object annotations, scene level annotations, and 2D/3D driven pathways. The annotations encompass multiple levels mirroring human cognition: (a) \textbf{\txtdeepred{R}}ecognizing work zones and detection of constituent objects using bounding boxes and segmentation labels, (b) \textbf{\txtdeepred{O}}bserving fine-grained details about the work zones using arrow board states, labeled text, arrows, and graphics on traffic signs, (c) \textbf{\txtdeepred{A}}nalyzing the work zones enabled by detailed and consistent human description about type of work along with spatial understanding about blockage and activity, and (d) the traversable \textbf{\txtdeepred{D}}riving path computed automatically from driving videos. Figure~\ref{fig:roadwork-zone-examples} shows examples of work zones and annotations from our dataset.

\datasetname consists of 4375 videos, 9650 richly annotated images, and 129017 images with passable trajectories. The videos were used to  propagate labels~\cite{ravi2024sam2} to 11959 additional keyframes to train our models. Like other recent papers for scaling data~\cite{kirillov2023segany}, we use bootstrapping to collect long-tailed work zone data. Our first bootstrapping stage involved manually capturing 2338 work zone images by driving in Pittsburgh to train our initial models. In our second bootstrapping stage, we manually curated 5078 keyframes after automatically mining nearly 100K potential images from Michelin Mobility Intelligence (MMI) Open Dataset~\cite{roadbotics-open} containing 45 million frames. From the selected keyframes, we obtained 4375 thirty-second videos of driving through work zones. All data is captured using smartphones with known intrinsics mounted inside the vehicle. In total, we estimate that our data covers around 5000 work zones (see \pointToAppendix{\ref{sup:dataset}}{B}) from the first two data curation stages. Most of the results presented in the paper correspond to training using data from the first two stages, unless otherwise stated. In our third bootstrapping stage, we expanded to in-the-wild data sources. We mined, discovered and annotated 969 work zone images from BDD100K~\cite{yu2020bdd100k} and Mapillary~\cite{neuhold2017mapillary} for in-the-wild evaluation. We also semi-automatically discovered images in bad weather and night conditions. We discovered 465 images by driving in Pittsburgh and other rural US areas, along with 800 images from a commuter bus that followed a fixed route in Pittsburgh over the course of two years.

\begin{figure}
    \centering
    \includegraphics[width=0.98\linewidth]{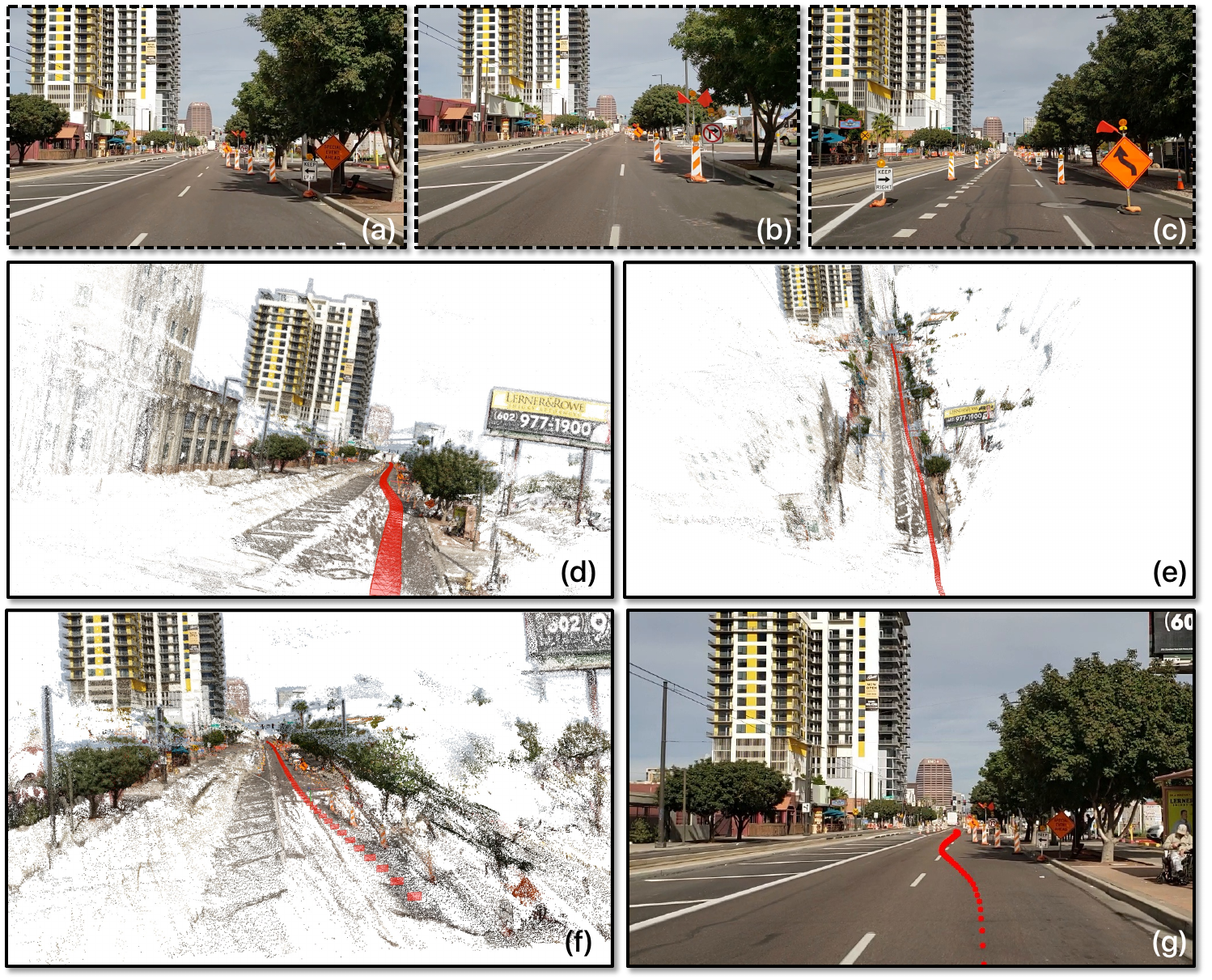}
    \vspace{-0.1in}    
    \caption{\textbf{Generating Driving Trajectory.} Using driving video frames (a-c) as input, we estimated camera poses using COLMAP~\cite{schoenberger2016sfm} (d-f), where camera poses are depicted as viewing frustums shown in red. These poses were projected onto the estimated ground plane to form the driven trajectory, then back-projected onto the frames as the future driving trajectory (g).} 
    \label{fig:pathway-3d}
    \vspace{-0.2in}    
\end{figure}

From the videos, we automatically extracted the actual path trajectory taken by the ego-vehicle. For this, we implemented a 3D reconstruction pipeline using off-the-shelf-modules (described in \pointToAppendix{\ref{sup:addl-recog-results}}{B.3}). We then obtained driven pathways by projecting the camera poses first onto the ground plane and then back onto the image. To ensure correctness, we manually verified all associated keyframes. Figure~\ref{fig:pathway-3d} visualizes some of these steps. This process resulted in 1936 unique sequences with passable trajectories for 129017 frames. A complete description of the acquisition, annotations, data cleaning protocols, and key statistics about \emphdatasetname are in~\pointToAppendix{\ref{sup:dataset}}{B}. 


\section{\textbf{\txtdeepred{Recognizing}} Work Zones}
\label{subsec:recognize}
\vspace{-0.05in}

\begin{figure}[t!]
\centering
\includegraphics[width=\linewidth]{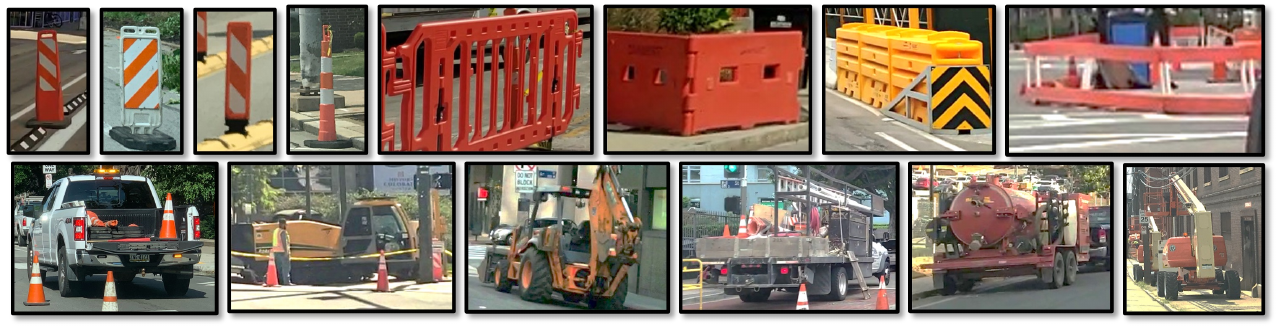}
\vspace{-0.2in}
\captionof{figure}{\small \textbf{Challenges And Variations In Work Zone Objects.} There exists a significant geographical variation in work zone objects and work vehicles in the \emphdatasetname dataset. For instance, the appearance of barriers, vertical panels, and tabular markers (first row) varies across cities. Similarly, work vehicles (second row) demonstrate large variations, as they are specialized for specific tasks within the work zones.}
\vspace{-0.1in}
\label{fig:work-objects-mosaic}
\end{figure}

\noindent
``Recognizing'' work zones is underexplored, with little research in this domain. We define ``recognition'' as being able to identify a work zone and objects present in it -- no matter the geography or conditions. Figure~\ref{fig:work-objects-mosaic} shows some of the associated challenges. The \emphdatasetname Dataset enables us to understand and improve these recognition challenges in the era of foundation models.


\noindent \textbf{Detecting And Segmenting Work Zone Objects.} Foundation models for open vocabulary scene understanding~\cite{zhou2022detecting, zhang2023simple, xu2023open, zhang2024omgllava} have shown impressive zero-shot generalization on many datasets. No closed vocabulary detector exists for recognizing work zones objects, leading us to consider these open vocabulary foundation models. Although earlier datasets considered few work zone objects such as cones~\cite{neuhold2017mapillary, nuscenes}, they did not consider all types of work zone objects, such as ``Arrow Board''.

\begin{table}[t]
\centering
\resizebox{\linewidth}{!}{
\begin{tabular}{lcccccc}
\hline
 &
  \multicolumn{1}{l}{$AP$} &
  \multicolumn{1}{l}{$AP50$} &
  \multicolumn{1}{l}{$AP75$} &
  \multicolumn{1}{l}{$AP_s$} &
  \multicolumn{1}{l}{$AP_m$} &
  \multicolumn{1}{l}{$AP_{l}$} \\ \hline
\multicolumn{7}{c}{\textbf{Foundation Models  (Zero-Shot)}}                             \\ \hline
OpenSeeD (COCO+O365)~\cite{zhang2023simple} & 2.9  & 5.8  & 2.5  & 1.3  & 2.8  & 4.0  \\
Detic (LVIS+I21K)~\cite{zhou2022detecting}  & 4.2  & 6.3  & 4.6  & 2.1  & 5.5  & 6.0  \\ 
OMG-LLaVA (Many Datasets)~\cite{zhang2024omgllava} & 0.2	& 0.3 &	0.2	& 0	& 0.1 & 0.8 \\ \hline
\multicolumn{7}{c}{\textbf{Supervised with \emphdatasetname Dataset}}                 \\ \hline
Mask R-CNN~\cite{he2017mask}                & 29.9 & 48.3 & 32.7 & 16.0 & 32.9 & 43.8 \\
Mask DINO~\cite{li2023maskdino}             & 36.4 & 55.2 & 40.7 & 20.2 & 37.4 & 52.2 \\
\rowcolor[HTML]{ECF4FF} 
Mask DINO~\cite{li2023maskdino} w/ Video Prop.~\cite{ravi2024sam2} &
  \textbf{39.0} &
  \textbf{58.8} &
  \textbf{43.4} &
  \textbf{22.1} &
  \textbf{40.0} &
  \textbf{55.0} \\ \hline
\end{tabular}%
}
\vspace{-0.1in}
\caption{\textbf{Segmenting Work Zone Objects.} We train instance segmentation~\cite{he2017mask, li2023maskdino} on \emphdatasetname using a coarse vocabulary. Open vocabulary methods~\cite{zhang2023simple, zhou2022detecting, liu2024grounding} fail to recognize work zone objects. In contrast, supervised models perform significantly better (\textbf{\txtgreen{+32.2 $AP$}}), and label propagation through video~\cite{ravi2024sam2} further improves performance (\textbf{\txtgreen{+2.6 $AP$}}).}
\vspace{-0.2in}
\label{tbl:segm-coarse-grained}
\end{table}

We investigate whether popular open vocabulary instance segmentation methods, Detic~\cite{zhou2022detecting} and OpenSeeD~\cite{zhang2023simple} generalize to our data. We also investigate whether OMG-LLaVA~\cite{zhang2024omgllava}, a foundation model trained for pixel level tasks, can reason about and segment work zone objects. Table~\ref{tbl:segm-coarse-grained} shows our results. Although few of the object categories are common concepts, such as ``Fence'' and ``Police Officer,'' open vocabulary methods~\cite{zhou2022detecting, zhang2023simple} fail to detect and segment most objects, such as ``Temporary Traffic Control (TTC) Message Board'' and ``Arrow Board''. This is irrespective of the large-scale training source. 
Note that all methods additionally leverage CLIP~\cite{radford2021learning}, which is trained on a 400 million image-text paired dataset, which likely under-represents work zones. For comparison, we train coarse vocabulary work zone object detectors by fine-tuning Mask R-CNN~\cite{he2017mask} and Mask DINO~\cite{li2023maskdino} on our dataset, producing significantly improved detection (\textbf{\txtgreen{+25.7 $AP$}} and \textbf{\txtgreen{+32.2 $AP$}}, respectively). However, many work zone categories are rare, e.g. the \emphdatasetname annotations itself has fewer than 1000 instances of ``Arrow Board''. To increase the number of instances, we use the Segment Anything 2 (SAM2)~\cite{ravi2024sam2} model to propagate segmentations through \emphdatasetname videos -- we prompt all ground truth masks as input for an annotated keyframe to SAM2. Then we propagate masks for the next 60 frames and uniformly add every 10th frame to our augmented dataset. We obtain 11959 more labeled frames using 1947 training videos from \emphdatasetname, in addition to 5318 images in the training set. Training on this augmented dataset yields further improvements (\textbf{\txtgreen{+2.6 $AP$}}).


In summary, work zone objects are woefully underrepresented in existing foundation models and datasets, highlighting the crucial role of \emphdatasetname dataset. We perform analysis of open-vocabulary detectors~\cite{liu2024grounding} and state-of-the-art object detectors in \pointToAppendix{\ref{sup:addl-recog-results}}{C.1}, which follow similar trends. In \pointToAppendix{\ref{sup:addl-recog-results}}{C.1}, we also show that these foundation models can be fine-tuned on our data with no or minimal loss in performance on common driving datasets.


\begin{table}[t!]
\centering
\resizebox{0.8\linewidth}{!}{
\begin{tabular}{lccc|ccc}
\hline
              & $AP$ & $AP50$ & $AP75$ & $AP$ & $AP50$ & $AP75$ \\ \cline{2-7} 
\multirow{-2}{*}{\textbf{Supervision}} & \multicolumn{3}{c|}{SAM~\cite{kirillov2023segany}} & \multicolumn{3}{c}{SAM2~\cite{ravi2024sam2}}  \\ \hline
Bbox          & 24.7 & 47.9   & 27.3   & 26.4 & 47.7   & 26.4   \\
Bbox + 5 pts  & 25.4 & 46.5   & 25.3   & 26.2 & 47.4   & 26.5   \\
Bbox + 10 pts & 25.8 & 47.2   & 25.9   & 25.0 & 46.8   & 24.6   \\ \hline \hline
\rowcolor[HTML]{ECF4FF} 
\textbf{Ground Truth}                     & \textbf{29.9}   & \textbf{48.3}   & \textbf{32.7}  & \textbf{29.9} & \textbf{48.3} & \textbf{32.7} \\ \hline
\end{tabular}%
}
\vspace{-0.1in}
\caption{\textbf{Are Manual Segmentations Still Necessary?} We train instance segmentation models~\cite{he2017mask} with varying levels of weak supervision from the \emphdatasetname dataset. We use SAM~\cite{kirillov2023segany} and SAM2~\cite{ravi2024sam2} to generate pseudo-ground truth masks from ground truth boxes and points. Manual ground truth masks (\textit{last row}) outperform SAM at tighter IoU thresholds (\textbf{\txtgreen{+5.4 $AP75$}}) and for rare categories such as ``Arrow Board'' (\textbf{\txtgreen{+26.9 $AP$}}). SAM2 pseudo-ground truth masks, while better than SAM psuedo-ground truth overall (\textbf{\txtgreen{+3.5 $AP$}}), perform worse at tighter thresholds (\textbf{\txtgreen{+6.3 $AP75$}}) when compared to manual ground-truth masks.}
\label{tbl:sam-results}
\vspace{-0.2in}
\end{table}

\noindent \textbf{Are Manual Segmentations Still Necessary?} Given foundation models for promptable segmentation~\cite{kirillov2023segany, ravi2024sam2} have shown competitive zero-shot generalization on many datasets, are manual segmentations even needed anymore? 

We use Segment Anything Model (SAM)~\cite{kirillov2023segany} and Segment Anything Model 2 (SAM2)~\cite{ravi2024sam2} to perform zero-shot instance segmentation. We use ground truth boxes as input to generate pseudo-ground truth masks for the \emphdatasetname dataset. We also simulate $K = \{5, 10\}$ positive and negative points from manual ground-truth masks as input prompts~\cite{cheng2022pointly}. We then train detectors (in this case, Mask R-CNN~\cite{he2017mask}) using manual and pseudo-ground truth masks. 

From Table~\ref{tbl:sam-results}, manual ground truth masks improve performance by \textbf{\txtgreen{+5.2 $AP$}} over pseudo-ground truth masks from SAM. This gap is more pronounced for rare objects, example, ``Arrow Board'' (\textbf{\txtgreen{+26.9 $AP$}}) and ``TTC Message Board'' (\textbf{\txtgreen{+8.7 $AP$}}). Pseudo-ground truth masks are comparable for common classes such as ``Worker'' (\textbf{\txtgreen{+1.2 $AP$}}) and ``Police Officer'' (\textbf{\red{-2.1 $AP$}}). Performance gap at tighter thresholds is higher (\textbf{\txtgreen{+5.4 $AP75$}}). Using additional annotated points as weak supervision reduced the overall gap to \textbf{\txtgreen{+4.1 $AP$}}. Performance of SAM2 is not much better than SAM.  Ground truth masks are still better than SAM2 (\textbf{\txtgreen{+3.5 $AP$}}) but the gap is smaller than SAM. SAM2 is worse than SAM at tighter IOU thresholds (\textbf{\txtgreen{+6.7 $AP$}}). In this case, additional point supervision affects performance at tighter thresholds, increasing the gap to \textbf{\txtgreen{+8.1 $AP$}} with ground truth masks. Thus, manual ground truth masks are still necessary.


\begin{table}[t!]
\resizebox{\linewidth}{!}{
\begin{tabular}{llcccc}
\hline
                               &             & \multicolumn{2}{c}{BDD100K~\cite{yu2020bdd100k}} & \multicolumn{2}{c}{Mapillary~\cite{neuhold2017mapillary}} \\ \cline{3-6} 
\multirow{-2}{*}{Method} &
  \multirow{-2}{*}{Supervision} &
  \multicolumn{1}{l}{\#Discovered} &
  \multicolumn{1}{l}{Precision} &
  \multicolumn{1}{l}{\#Discovered} &
  \multicolumn{1}{l}{Precision} \\ \hline
Detic~\cite{zhou2022detecting} & I21K + LVIS & 32                    & 52.4 \%                  & 125                       & 42.9 \%                       \\
\rowcolor[HTML]{ECF4FF} 
Mask R-CNN~\cite{he2017mask}   & COCO + ROADWork    & \textbf{411}          & \textbf{84.9 \%}         & \textbf{558}              & \textbf{77.0 \%}              \\ \hline
\end{tabular}%
}
\vspace{-0.1in}
\caption{\textbf{Characterizing and Discovering Work Zones.} The \emphdatasetname dataset improves discovery of new work zones compared to open-vocabulary foundation models. Using the same classification rule, we employ a coarse-vocabulary detector~\cite{he2017mask} trained on \emphdatasetname dataset. We discovered \textbf{\txtgreen{12.8$\times$}} and \textbf{\txtgreen{4.5$\times$}} more work zone images than  Detic~\cite{zhou2022detecting}, improving precision by \textbf{\txtgreen{+32.5\%}} and \textbf{\txtgreen{+34.0\%}} on BDD100K~\cite{yu2020bdd100k} and Mapillary~\cite{neuhold2017mapillary}.}
\vspace{-0.1in}
\label{tbl:discover}
\end{table}

\begin{figure}
    \centering
    \vspace{-0.1in}
    \includegraphics[width=0.9\linewidth]{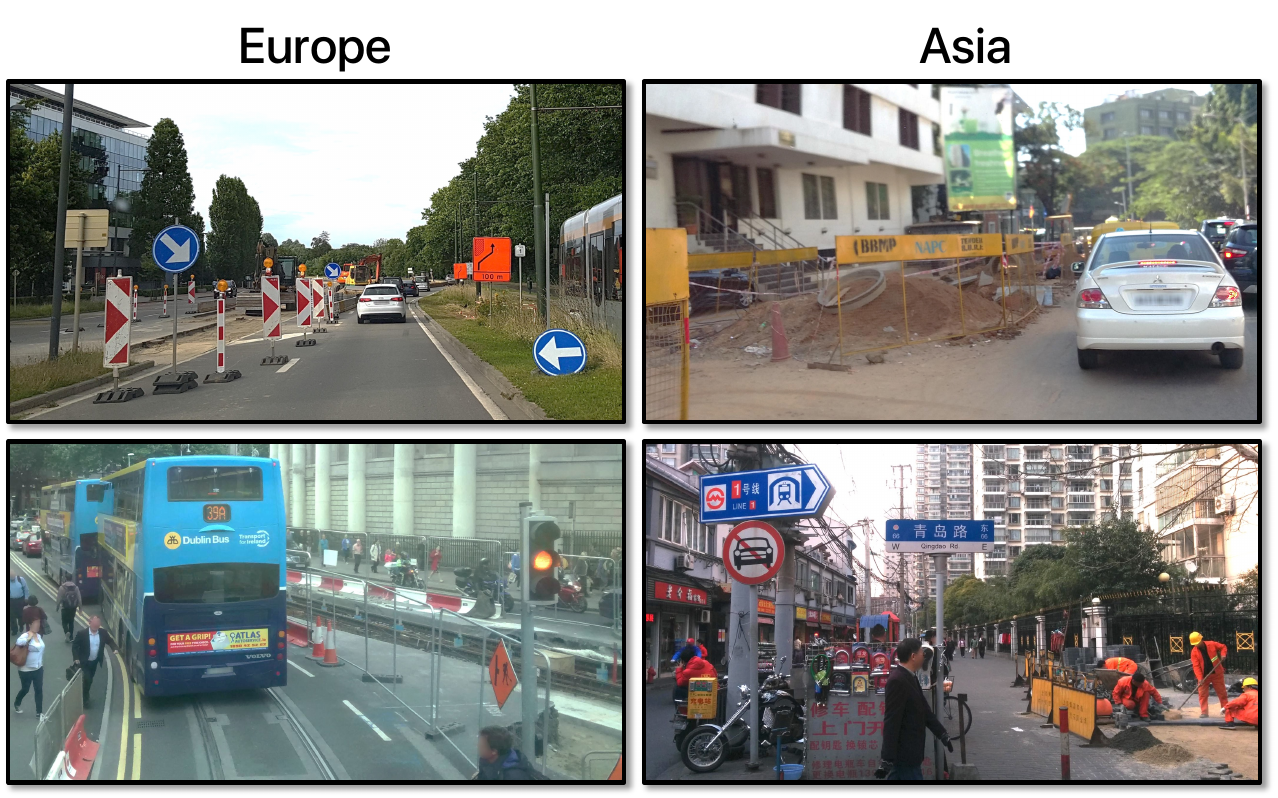}
    \vspace{-0.1in}
    \caption{\textbf{Work Zones Discovered Around The World.} Mapillary~\cite{neuhold2017mapillary} dataset contains driving images from around the world. Despite the \emphdatasetname training dataset images being restricted to the U.S., we discovered work zones captured in Europe and Asia.}
    \vspace{-0.2in}
    \label{fig:discovered-mapillary}
\end{figure}

\noindent \textbf{Characterizing And Discovering Work Zones.} During dataset curation, we manually filtered work zone images (See Section~\ref{sec:dataset}) — a process too cumbersome to scale to millions of images at lower precision levels. Thus, automatically discovering new work zones from vast amounts of unlabeled data is important. It is also useful to understand whether models trained on the \emphdatasetname dataset generalize to new data sources. However, the existence of a work zone object (e.g., a work vehicle) alone does not make a scene a work zone. \textit{We classify a work zone as any activity on the road, defined by specific objects and human actors, that redefines the road network and traffic rules.} A proxy for work zone activity is its ``intensity,'' measured by the number and relative area of work zone objects in the scene.

\begin{figure}[t!]
    \centering
    \includegraphics[width=0.9\linewidth]{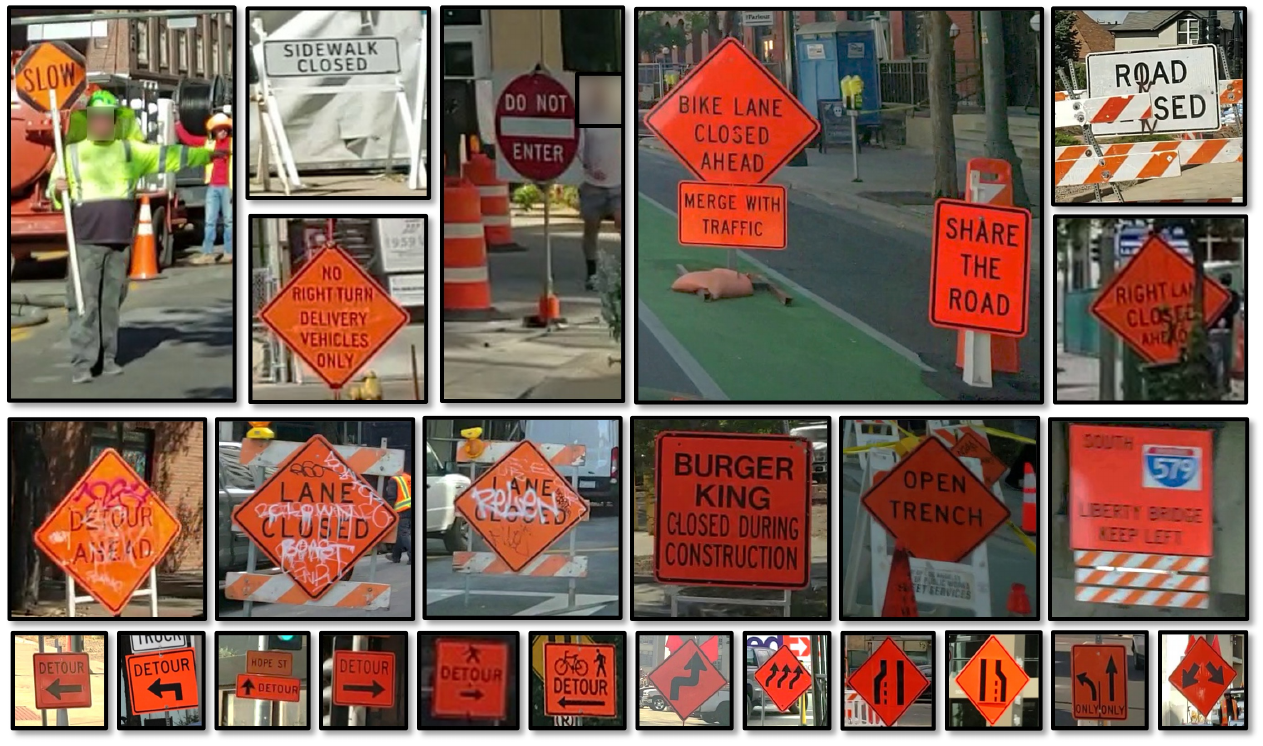}
    \vspace{-0.1in}
    \caption{\textbf{Challenges In Fine-Grained Observations Like Signs.} TTC signs contain ``text'', ``graphics'', or both. We annotated 62 types of graphics and 360 text types across 8242 sign instances. These signs convey specific information (e.g., for pedestrians or delivery vehicles), require interpretation with other signs, and are may be vandalized. They also impact long-term navigation (e.g., driving towards Burger King and reading ``Burger King Closed During Construction``). Thus, interpreting signs is not only challenging but also requires understanding the broader scene.}
    \vspace{-0.2in}
    \label{fig:sign-mosaic}
\end{figure}

We discover work zones in existing self-driving datasets devising a simple classification rule: an image is a work zone \textit{iff} at least three instances of work zone objects from two unique categories that occupy at least $10\%$ of the image area. We chose Mask R-CNN for discovery due to its good accuracy-latency trade-offs~\cite{li2020towards, huang2017speed} for processing millions of images. Compared to foundation model Detic~\cite{zhou2022detecting}, Mask R-CNN trained on \emphdatasetname discovers work zone images with higher precision in BDD100K~\cite{yu2020bdd100k} and Mapillary~\cite{neuhold2017mapillary} datasets. We found fewer than 1000 roadwork images, representing only \txtdeepblue{$0.4\%$} and \txtdeepblue{$2.3\%$} of these datasets respectively. Surprisingly, while the \emphdatasetname dataset contains images only from US cities, the model discovered work zones worldwide (Figure~\ref{fig:discovered-mapillary}). Furthermore, detectors trained on the \emphdatasetname dataset perform significantly better than open-vocabulary methods on discovered images in a variety of other conditions along with BDD100K and Mapillary; see \pointToAppendix{\ref{sup:addl-recog-results}}{C.1}.

\noindent \textbf{\textit{Summary.}} The \emphdatasetname dataset significantly helps to recognize and discover work zones. Further, In \pointToAppendix{\ref{sup:addl-recog-results}}{C.1}, open-vocabulary detectors~\cite{liu2024grounding} fine-tuned on the \emphdatasetname data still generalize to common driving scenes, while not degrading performance on Cityscapes~\cite{cordts2016cityscapes} (\textbf{\txtgreen{+0.2 $AP$}}) compared to a pretrained model. \emphdatasetname enables study of other interesting long-tailed recognition problems. In \pointToAppendix{\ref{sup:other-interesting-scenarios}}{C.1.1}, we show that there exist large domain gaps (\textbf{\txtred{+7.4 $AP$}}) between cities even with SOTA adaptation methods~\cite{kennerley20232pcnet} (\textbf{\txtred{-13.8 $AP_{50}$}} for ``barricade'' compared to upper-bound). We also show results for label unification. Detector performance degrades (\textbf{\txtred{-5.0 $AP$}}) when models are jointly trained on datasets with common and rare object labels. The state-of-the-art label-unification algorithm~\cite{wang2023detecting} marginally improves performance (\textbf{\txtgreen{+2.0 $AP$}}).


\section{\textbf{\txtdeepred{Observing}} Work Zones}
\label{subsec:observe}

\noindent
As noted in Section~\ref{sec:dataset}, recognizing objects alone is insufficient; fine-grained local observations and understanding is essential, such as sign interpretation and human intent recognition. We define this challenge as ``observing''. For example, consider temporary traffic control (TTC) signs in Figure~\ref{fig:sign-mosaic}. Sign interpretation is valuable for both short-term navigation and long-term planning. For the purposes of our discussion, we break down sign interpretation into two sub-problems: recognizing the graphical elements and reading the text. We discuss challenges in current methods.

\noindent \textbf{Segmenting Fine-Grained Signs.} ``Graphics'' refers to the pictorial diagrams that are frequently present in signs, such as arrows or stick figures that depict humans. \emphdatasetname  contains 62 types of signs board graphics. Frequency distribution of these graphics is long-tailed and we show some examples in Figure~\ref{fig:sign-mosaic}. In the coarse vocabulary considered in Section~\ref{subsec:recognize}, all TTC signs were binned into the general ``TTC sign'' category. Here, we consider a fine-grained vocabulary, where each sign-board graphics is a separate category, expanding the label vocabulary to 49 classes. However, scarce categories (sign graphics with less than 5 instances) were binned into ``Other'' category. 

As discussed in Section~\ref{subsec:recognize} we employ methods like Detic~\cite{zhou2022detecting}, OpenSeeD~\cite{xu2023open} and OMG-LLaVA~\cite{zhang2024omgllava}. We fine-tune instance segmentation models~\cite{he2017mask, li2023maskdino} on \emphdatasetname with fine-grained vocabulary for comparison. Our observations are in Table~\ref{tbl:segm-fine-grained}.All the foundation models~\cite{zhou2022detecting, zhang2023simple, zhang2024omgllava} are dismal in segmenting these signs while Mask DINO~\cite{li2023maskdino} performs considerably better (\textbf{\txtgreen{+28.9 $AP$}}). Following Section~\ref{subsec:recognize}, we adopt a similar strategy to exploit \emphdatasetname videos and to propagate ground-truth labels  using SAM2~\cite{ravi2024sam2} and obtain an additional improvement of \textbf{\txtgreen{+2.8 $AP$}}. However, overall performance of Mask R-CNN was still low (\textbf{\txtgreen{17.6 $AP$}}) compared to Mask DINO. We use simple copy-paste augmentation~\cite{ghiasi2021simple} which improves rare category segmentation. It further improved the accuracy of our Mask R-CNN~\cite{he2017mask} model by \textbf{\txtgreen{+7.1 $AP$}}.

\begin{table}[t!]
\centering
\resizebox{\linewidth}{!}{
\begin{tabular}{lcccccc}
\hline
 &
  \multicolumn{1}{l}{$AP$} &
  \multicolumn{1}{l}{$AP50$} &
  \multicolumn{1}{l}{$AP75$} &
  \multicolumn{1}{l}{$AP_s$} &
  \multicolumn{1}{l}{$AP_m$} &
  \multicolumn{1}{l}{$AP_{l}$} \\ \hline
\multicolumn{7}{c}{\textbf{Foundation Models (Zero-Shot)}}                                                         \\ \hline
OpenSEED (COCO+O365)~\cite{zhang2023simple}      & 1.0  & 2.0  & 0.9  & 0.5           & 1.1           & 2.1  \\
Detic (LVIS+I21K)~\cite{zhou2022detecting}       & 1.1  & 1.5  & 1.3  & 0.3           & 1.8           & 2.2  \\ 
OMG-LLaVA (Many Datasets)~\cite{zhang2024omgllava} & 0	& 0.1 & 0.1 & 	0	& 0	& 0.2 \\
\hline
\multicolumn{7}{c}{\textbf{Supervised with \emphdatasetname Dataset}}                                        \\ \hline
Mask R-CNN~\cite{he2017mask}                     & 17.6 & 26.2 & 19.5 & 11.3          & 21.2          & 36.0 \\
Mask R-CNN w/ Copy Paste~\cite{ghiasi2021simple} & 25.0 & 34.6 & 27.8 & 11.0          & 26.9          & 46.8 \\
Mask DINO~\cite{li2023maskdino}                  & 30.0 & 40.9 & 33.4 & \textbf{20.6} & \textbf{34.4} & 50.1 \\
\rowcolor[HTML]{ECF4FF} 
Mask DINO~\cite{li2023maskdino} w/ Video Prop.~\cite{ravi2024sam2} &
  \textbf{32.8} &
  \textbf{43.4} &
  \textbf{36.4} &
  18.7 &
  32.7 &
  \textbf{60.2} \\ \hline
\end{tabular}%
}
\vspace{-0.1in}
\caption{\textbf{Segmenting Fine-Grained Signs.} We expand the ``TTC sign'' category to classify different kinds of TTC signs into separate categories.  Open vocabulary methods~\cite{xu2023open, zhou2022detecting} fail to detect most of these objects with this expanded vocabulary. Supervised Mask DINO~\cite{li2023maskdino} improves considerably (\textbf{\txtgreen{+23.9 $AP$}}). Additionally, simple copy-paste~\cite{zhang2023simple} ensures simpler Mask R-CNN~\cite{he2017mask} baseline compares favorably with SOTA transformers.}
\vspace{-0.15in}
\label{tbl:segm-fine-grained}
\end{table}

\begin{table}[t!]
\centering
\resizebox{\linewidth}{!}{
\begin{tabular}{lccccc}
\hline
 &
  \multicolumn{1}{l}{\textbf{Edit. (1 - NED)}} &
  \multicolumn{1}{l}{\textbf{Word Acc.}} &
  \multicolumn{1}{l}{\textbf{Ch. Recall}} &
  \multicolumn{1}{l}{\textbf{Ch. Precision}} &
  \multicolumn{1}{l}{\textbf{Ch. F1}} \\ \hline
\multicolumn{6}{c}{\textbf{Foundation Models (Zero-Shot)}}                               \\ \hline
Gemini 2.5 Flash~\cite{comanici2025gemini} & 22.7           & 4.5               & \textbf{94.1}                & 14.6                   & 25.2 \\ 
GPT-4o~\cite{hurst2024gpt4o}           & 40.3           & 11.9              & 84.7                & 27.3                   & 41.3 \\ \hline
\multicolumn{6}{c}{\textbf{Specialized Text Spotting Model (Zero-Shot)}}                               \\ \hline
Glass~\cite{ronen2022glass} & 65.4          & 56.0          & 65.8          & \textbf{98.4} & 78.9          \\
Glass (w/ 1x crops)         & 76.8          & 56.3          & 78.9          & 84.2          & 81.4          \\
\rowcolor[HTML]{ECF4FF} 
Glass (w/ 3x crops)         & \textbf{79.6} & \textbf{59.6} & 80.6 & 86.1          & \textbf{83.2} \\ \hline
\multicolumn{6}{c}{\textbf{Small Signs Subset (Area less than $32\times32$)}}                               \\ \hline
Glass~\cite{ronen2022glass} & 19.6          & 13.2          & 18.1          & \textbf{93.1} & 30.3          \\
Glass (w/ 1x crops)         & 76.7          & 20.9          & 42.2          & 70.4          & 52.8          \\
\rowcolor[HTML]{ECF4FF} 
Glass (w/ 3x crops)         & \textbf{81.3} & \textbf{32.3} & \textbf{48.7} & 78.4          & \textbf{60.0} \\ \hline
\end{tabular}%
}
\vspace{-0.1in}
\caption{\textbf{Reading Sign Text.} The pretrained text spotting method Glass~\cite{ronen2022glass} performs well on nearby signs, but it fails to detect and read distant text (See Figure~\ref{fig:sign-text-errors}). Closed foundation models like GPT-4o~\cite{hurst2024gpt4o} and Gemini 2.5~\cite{comanici2025gemini} are worse. A simple crop-rescale strategy composing a work zone detector with Glass~\cite{ronen2022glass} improves text spotting and recognition by \textbf{\txtgreen{+14.2\% (1 - NED)}}. 
}
\label{tbl:text-spotting}
\vspace{-0.15in}
\end{table}

\begin{figure}
    \centering
    \includegraphics[width=0.95\linewidth]{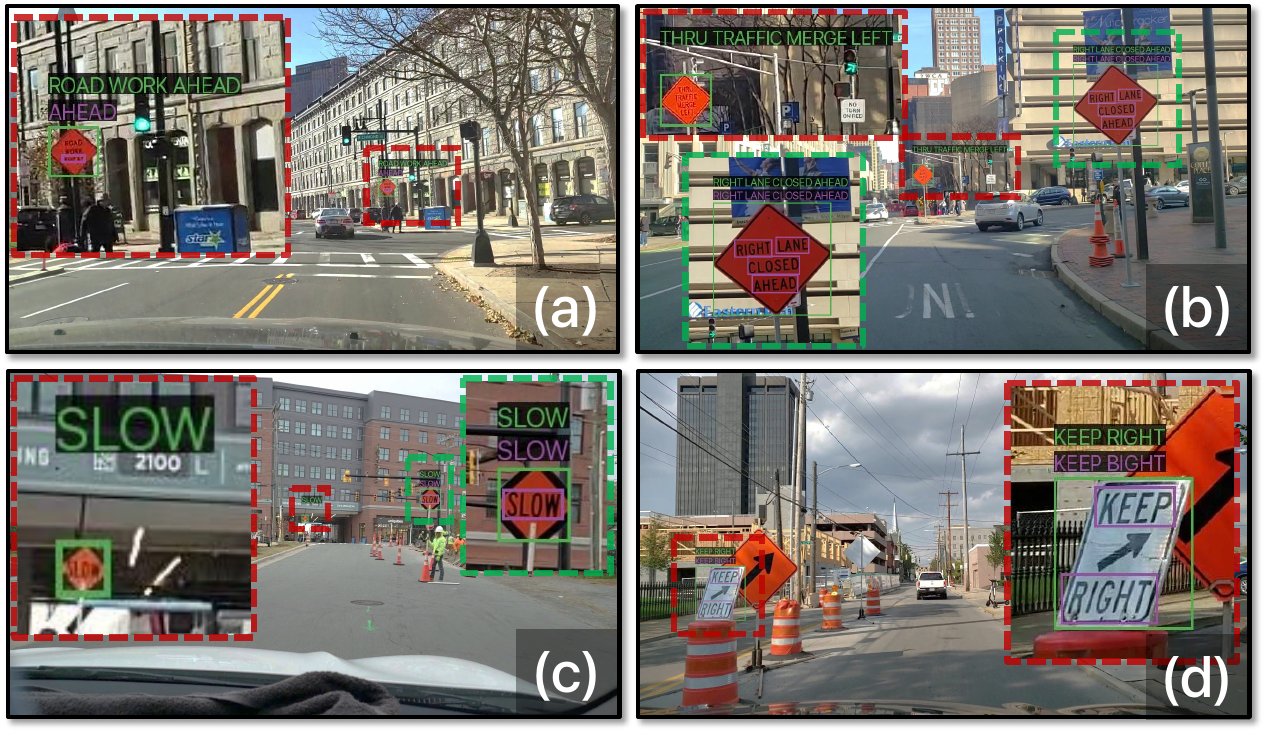}
    \vspace{-0.15in}
    \caption{\textbf{Reading Sign Text.} Glass~\cite{ronen2022glass} reads nearby signs   but fails for distant signs (a-c). Dirt, occlusions and vandalism are issues for close-by signs too. In (d), prediction is ``Keep \textit{B}ight'' .}
    \vspace{-0.15in}
    \label{fig:sign-text-errors}
\end{figure}

\begin{figure*}[t!]
    \centering
    \includegraphics[width=0.95\textwidth]{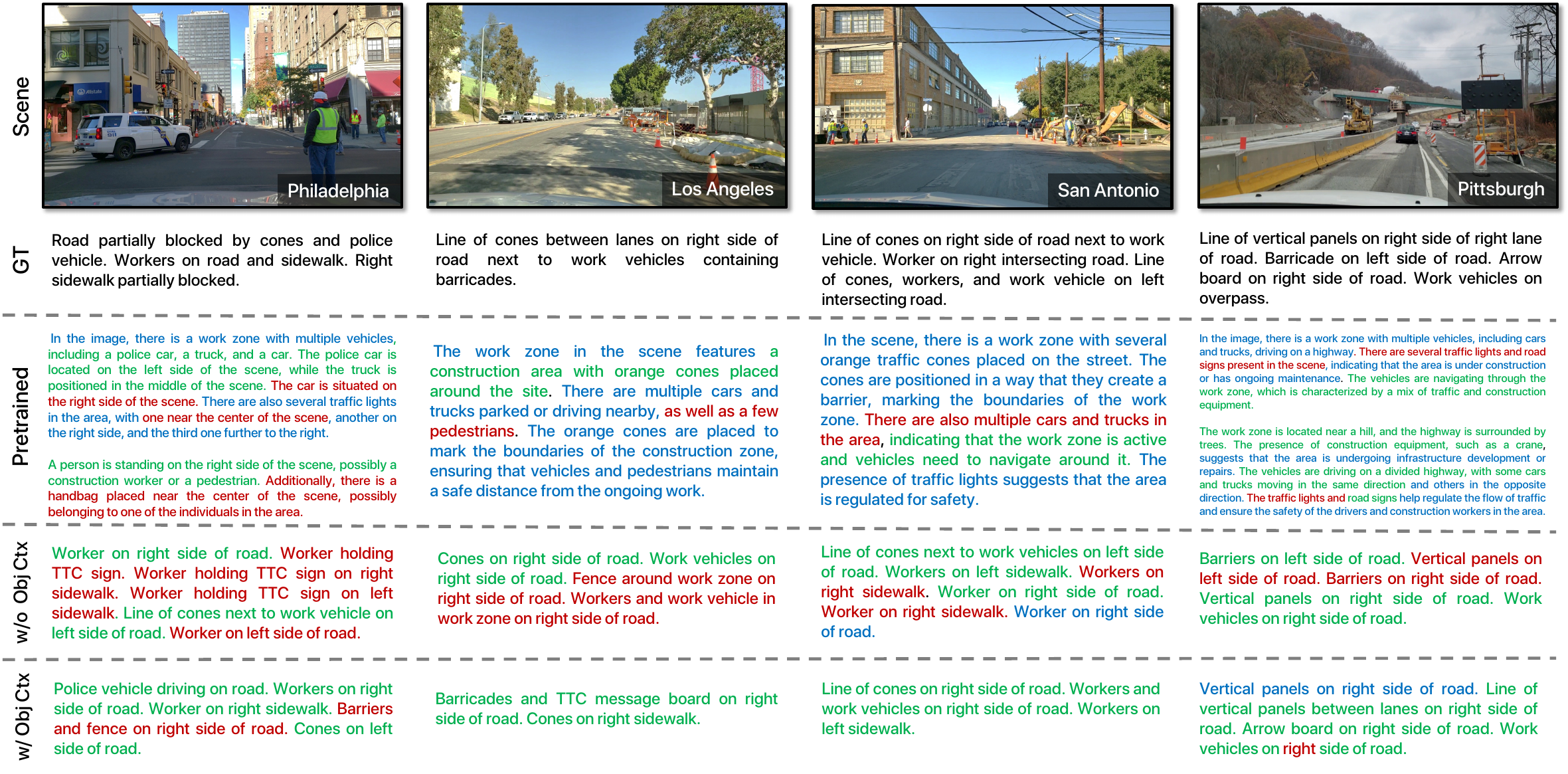}
    \vspace{-0.1in}
    \caption{\textbf{Generating Work Zone Descriptions.} We present examples of work zone descriptions generated by a pretrained LLaVA~\cite{liu2024improved} and a LoRA~\cite{hu2022lora} fine-tuned LLaVA (with and without additional object context). The generated descriptions can be classified as \textbf{\txtgreen{factual}},  \textbf{\txtdeepred{incorrect}} or \textbf{\txtblue{uninformative}}. As we can observe, while pretrained LLaVA correctly identifies certain scene elements, it also generates many uninformative or inaccurate sentences. Fine-tuning LLaVA without object context reduces uninformative outputs but introduces hallucinated scene elements that do not exist. Incorporating LLaVA with our workzone detector to provide object context significantly improves faithfulness; however, it still misses some scene elements and spatial interactions (See San Antonio example's ground truth).}
    \label{fig:llava-captions}
    \vspace{-0.2in}
\end{figure*}

\noindent \textbf{Reading Sign Text.}  \emphdatasetname contains 360 different TTC sign texts. Sign texts are unconstrained and provide navigational instructions. Spotting and reading them accurately is important. We consider foundation models like GPT-4o~\cite{hurst2024gpt4o} and Gemini 2.5 Flash~\cite{comanici2025gemini} and a popular scene text spotting method, Glass~\cite{ronen2022glass}. We evaluate the spotter on TTC signs, and ignore other text in the scene. Foundation Models are unable to localize the text regions accurately, which is consistent with their underperformance on standard tasks~\cite{ramachandran2025well}. We then try Glass~\cite{ronen2022glass}, trained on the TextOCR~\cite{singh2021textocr} dataset. Glass accurately spots and reads most TTC signs, but fails on small signs far away (see Figure~\ref{fig:sign-text-errors}). Our annotations do not include character-level annotations required by Glass~\cite{ronen2022glass} so we did not fine-tune it on our data. 

We report sentence, word, and character level metrics for text spotting~\cite{ronen2022glass, chng2019icdar2019}. From Table~\ref{tbl:text-spotting}, we observe that while the precision of Glass~\cite{ronen2022glass} is high, the recall is low. Thus, we adopted a simple cropping strategy, composing Glass~\cite{ronen2022glass} with a coarse vocabulary work zone object detector~\cite{he2017mask} (Section~\ref{subsec:recognize}). We infer the text spotter only on TTC sign crops. Our strategy significantly increased recall (\textbf{\txtgreen{+13.1\%}}) at the expense of precision. Increasing the scale of the crops improved all metrics, such as word accuracy (\textbf{\txtgreen{+4.5\%}}).  There is a large performance gap for small signs -- around 20\% to 30\% lower. The crop rescale strategy significantly improves word accuracy (\textbf{\txtgreen{+21.3\%}}) for such signs. 

\noindent
\noindent \textbf{\textit{Summary.}} The \emphdatasetname dataset enables studying foundation models to interpret sign graphics and spot sign text. Addressing failures, simple strategies such as copy-paste~\cite{ghiasi2021simple}, video label propagation~\cite{ravi2024sam2} and crop-rescaling image regions all improve downstream performance.

\vspace{-0.1in}
\section{\textbf{\txtdeepred{Analyzing}} Work Zones}
\label{subsec:analyze}

\noindent
Sections~\ref{subsec:recognize} and~\ref{subsec:observe} focused on object recognition and fine-grained understanding and observations. However, global scene analysis is equally crucial for navigating work zones.
In this section, we study how well foundation models generate global work zone descriptions and use ``analyzing'' to refer to generating scene descriptions that may guide downstream planning.





\noindent
\textbf{Generating Work Zone Descriptions.} Vision-Language Models (VLM) have shown strong performance in scene understanding in various scenarios~\cite{radford2021learning, li2022blip, liu2023llava, liu2024improved}. However, work zones are underrepresented in their training data. Our experiments show that while pre-trained BLIP~\cite{li2022blip} and LLaVA~\cite{liu2024improved, li2024llavanext-strong} fail to generate informative work zone descriptions, fine-tuning on \emphdatasetname enables them to produce accurate and contextually relevant descriptions. 

\begin{table}[t!]
\resizebox{\linewidth}{!}{
\vspace{-0.1in}
\begin{tabular}{lccccc}
\hline
 \textbf{Method} &
  \multicolumn{1}{l}{\textbf{BLEU@4}} &
  \multicolumn{1}{l}{\textbf{METEOR}} &
  \multicolumn{1}{l}{\textbf{ROUGE}} &
  \multicolumn{1}{l}{\textbf{CIDEr}} &
  \multicolumn{1}{l}{\textbf{SPICE}} \\ \hline
\multicolumn{6}{c}{\textbf{Pretrained Foundation Models}}                                 \\ \hline
BLIP-361M~\cite{li2022blip}         & 0.2  & 4.3  & 8.6  & 1.6   & 3.9  \\
LLaVA-1.5-7B~\cite{liu2024improved} & 0.4  & 11.0 & 9.4  & 0     & 9.9  \\
GPT-4o~\cite{hurst2024gpt4o} & 0.9             & 16.4            & 14.4 & 0 & -  \\
Gemini 2.5 Flash~\cite{comanici2025gemini} & 0.6             & 13.2            & 11.7 & 0 & - \\ \hline
\multicolumn{6}{c}{\textbf{Fine-tuned Foundation Models}}                                  \\ \hline
BLIP-361M~\cite{li2022blip}         & 20.7 & 21.5 & 43.7 & 83.5  & 41.3 \\
LLaVA-1.5-7B~\cite{liu2024improved} & 27.0 & 24.7 & 48.0 & 112.1 & 42.7 \\ \hline
\multicolumn{6}{c}{\textbf{Fine-tuned w/ Object Context}}                \\ \hline
\rowcolor[HTML]{ECF4FF} 
~\cite{liu2024improved} + Pred. Objs &
  \textbf{31.1} &
      \textbf{27.7} &
  \textbf{50.9} &
  \textbf{140.3} &
  \textbf{46.6} \\ \hline \hline
~\cite{liu2024improved} + GT Objs   & 32.5 & 28.5 & 52.5 & 166.0 & 49.9 \\ \hline
\end{tabular}%
}
\vspace{-0.1in}
\caption{\textbf{Generating Work Zone Descriptions.} Pretrained VLMs, BLIP~\cite{li2022blip} and LLaVA-1.5-7B~\cite{liu2024improved}, do not generate informative descriptions about work zones. Finetuning 
 with LoRA~\cite{hu2022lora} on \emphdatasetname dramatically improves performance. LLaVA-1.5-7B~\cite{liu2024improved} significantly outperforms BLIP~\cite{li2022blip}, though larger LLaVA models perform similarly (See \pointToAppendix{~\ref{sup:addl-analyzing-work-zones}}{C.2}).  Using object context enhances description faithfulness, but a performance gap remains between ground truth and uncertain object predictions.}
\label{tbl:workzone-captions}
\vspace{-0.2in}
\end{table}

Table~\ref{tbl:workzone-captions} shows improvements with fine-tuning. LLaVA-7B~\cite{liu2024improved} outperforms BLIP-361M~\cite{li2022blip} on all metrics by a large margin (e.g. \txtgreen{\textbf{+32.8 SPICE}}) -- effectively utilizing the learned priors of a bigger model. Apart from fine-tuning, we added work zone object context by appending detection predictions from a trained work zone object detector, which further improves performance (\txtgreen{\textbf{+3.9 SPICE}}). Using ground truth objects as context establishes an upper bound, and a gap exists compared to providing uncertain object predictions (\red{\textbf{-3.3 SPICE}}). Figure~\ref{fig:llava-captions} shows some generated descriptions from pre-trained LLaVA~\cite{liu2024improved}, with untrue and irrelevant sentences. In comparison, training on \emphdatasetname dramatically improves description quality, and providing object context reduces hallucinations. 

\noindent \textbf{\textit{Summary.}} VLMs fail to describe work zones without the \emphdatasetname dataset. Beyond fine-tuning, providing object context significantly improves performance.  \pointToAppendix{\ref{sup:addl-analyzing-work-zones}}{C.2} shows that larger foundation models, \eg LLaVA-NEXT~\cite{li2024llavanext-strong}, also perform poorly, highlighting the need for the \emphdatasetname dataset. Fine-tuning VLMs on our dataset preserves the overall performance and even improves COCO-Captions~\cite{chen2015microsoft} (\textbf{\txtgreen{4.7 $\rightarrow$ 12.6 BLEU@4}}).

\noindent \section{\textbf{\txtdeepred{Driving}} Through Work Zones}
\label{subsec:drive}

\noindent
In Sections~\ref{subsec:recognize},~\ref{subsec:observe} and~\ref{subsec:analyze}, we studied challenges in recognizing, observing, and analyzing work zones. In this section, we discuss the challenges in driving through work zones and show how to predict goals and pathways in the image space to address these challenges.



\noindent
\textbf{Pathway Prediction In Work Zone Images.}  Work zones demand long-term planning beyond bird's eye view (BEV)~\cite{claussmann2019review} representations, which are common for driving scenarios~\cite{teng2023motion, mcnaughton2011motion, hu2023planning, dauner2023parting} such as highways. Rare work zone objects and their spatial interactions are difficult to capture using BEV alone. Additionally, navigational goals in work zones are inherently multi-modal and cannot be effectively constrained  within a fixed temporal window. 
Thus, we represent goals and pathways as spatial probability distributions in the image space, similar to long-horizon human path planning~\cite{mangalam2021goals}. Perception cost maps for BEV planners~\cite{lu2014layered, ferguson2008motion, salzmann2020trajectron} can be derived from these distributions. 


\begin{table}[]
\centering
\resizebox{\linewidth}{!}{
\begin{tabular}{lcccc|cccc}
\hline
\multirow{3}{*}{\textbf{$AE \% < \theta$}} & \multicolumn{4}{c|}{Goal}                   & \multicolumn{4}{c}{Pathway}                 \\ \cline{2-9} 
 & \multicolumn{2}{c}{All Curv.} & \multicolumn{2}{c|}{High Curv.} & \multicolumn{2}{c}{All Curv.} & \multicolumn{2}{c}{High Curv.} \\ \cline{2-9} 
 & C  & \textbf{\txtdeepred{R}}  & C   & \textbf{\txtdeepred{R}}   & C  & \textbf{\txtdeepred{R}}  & C   & \textbf{\txtdeepred{R}}  \\ \hline
$\theta = {\ang{0.5}}$                     & 43.7 & \textbf{53.6} & 31.2 & \textbf{38.4} & 67.2 & \textbf{75.3} & 67.1 & \textbf{75.4} \\
$\theta = {\ang{1}}$                       & 63.5 & \textbf{73.8} & 53.8 & \textbf{60.5} & 82.8 & \textbf{87.2} & 82.7 & \textbf{87.0} \\
$\theta = {\ang{2}}$                       & 79.0 & \textbf{85.9} & 73.8 & \textbf{83.0} & 90.5 & \textbf{93.0} & 90.4 & \textbf{93.0} \\
$\theta = {\ang{5}}$                       & 89.6 & \textbf{93.2} & 91.4 & \textbf{96.8} & 96.0 & \textbf{97.3} & 96.0 & \textbf{97.3} \\
$\theta = {\ang{10}}$                      & 94.5 & \textbf{97.0} & 95.7 & \textbf{98.9} & 98.5 & \textbf{99.3} & 98.5 & \textbf{99.3} \\ \hline
\end{tabular}%
}
\vspace{-0.1in}
\caption{\textbf{Pathway Prediction In Work Zones.} YNet~\cite{mangalam2021goals} models are trained on \emphdatasetname dataset pathway data using pretrained segmentation models from Cityscapes (C) and \emphdatasetname (\textbf{\txtdeepred{R}}). Our metric measures the percentage of predictions where the angular error (AE) between the predicted and ground-truth pathway falls below a given threshold, considering the image's field of view ($\sim$50\degree). The model trained with \emphdatasetname segmentations consistently outperforms the Cityscapes-trained model at all thresholds. We also report results on a subset of high-curvature paths, hypothesizing that navigating irregular work zone paths is more challenging. As expected, $AE \% < \theta$ of predicted goal is lower (\textbf{\txtred{-15.4\%}}) for this subset at tight thresholds ($\ang{0.5}$ to $\ang{2}$).}
\vspace{-0.2in}
\label{tbl:pathways-expt-aa}
\end{table}

\begin{figure}
    \centering
    \includegraphics[width=0.87\linewidth]{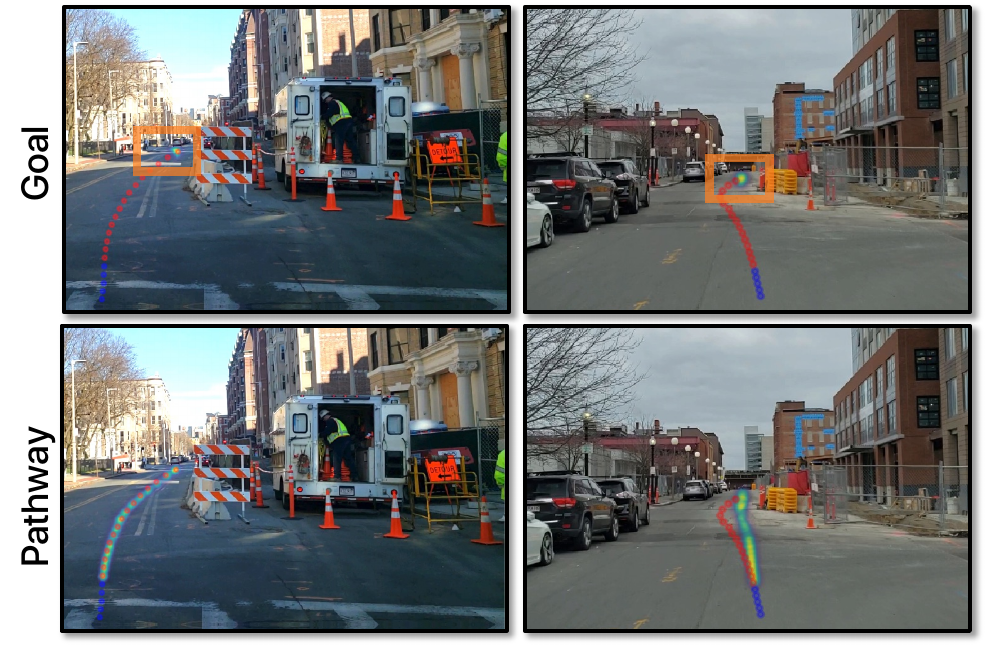}
    \vspace{-0.1in}    
    \caption{\textbf{Pathway Prediction in Work Zones.} We show goal and trajectory heatmaps predicted by YNet~\cite{mangalam2021goals}. The model inputs are the image and the \textbf{\textcolor{blue(ryb)}{observed pathway}}. We display the \textbf{\textcolor{cadmiumred}{future pathway}} (both computed from driving videos). The top row shows the predicted goal heatmaps, and the bottom row shows the predicted pathway heatmaps conditioned on a sampled goal. The predicted goal heatmap (highlighted with an \textbf{\txtorange{orange box}}) closely aligns with the ground-truth, and the pathway appears plausible.}
    \label{fig:pathway-good}
    \vspace{-0.2in}    
\end{figure}

We employ YNet~\cite{mangalam2021goals}. We pre-train the segmentation model that YNet employs using the Cityscapes~\cite{cordts2016cityscapes} dataset which contains no work zone objects and with the \emphdatasetname dataset. Both models are trained on pathways from our dataset. Pixel-level metrics~\cite{mangalam2021goals} do not account for the field of view of images (See \pointToAppendix{\ref{sup:addl-results}}{D}). Thus, we employ Percentage of Paths within Angular Error threshold $\theta$ ($AE\% < \theta$) as our metric. Table~\ref{tbl:pathways-expt-aa} presents our results. Our model, employing work zone semantic segmentations is better at all $\theta$ thresholds, than the model employing Cityscapes semantic segmentations. For example, at threshold $\ang{0.5}$, our model improves AE\% for goals by \textbf{\txtgreen{+9.9\%}} and AE\% for pathways by \textbf{\txtgreen{+8.1\%}} (see Figure~\ref{fig:pathway-good} for examples). We hypothesize navigating irregular work zone paths that deviate from a straight line is more difficult. Thus we also report results on subsets of paths thresholded at different curvatures. AE\% at $\ang{0.5}$ for goals on paths with high curvature is lower by \red{-15.4\%} than all paths. Our dataset demonstrates that curved paths are indeed more difficult. 

\noindent \textbf{\textit{Summary.}} The \emphdatasetname dataset enables the prediction of long-term navigation goals and paths in work zones. We introduced a novel and important task for navigating work zones, while providing evaluation metrics and baselines. More analysis and details can be found in \pointToAppendix{\ref{sup:addl-driving-results}}{C.3}.

\vspace{-0.1in}
\section{Conclusion}
\label{sec:conclusion}
\vspace{-0.05in}

We introduced the \emphdatasetname dataset to address scene understanding and navigation in the challenging yet critical problem of self-driving in work zones. By mining thousands of hours of driving data, we curated nearly 5000 annotated work zones. Foundation models struggle on many work zone tasks. We highlighted several challenges, from work zone object recognition, to fine-grained scene understanding, to work zone description, to long-horizon pathway prediction. Our findings demonstrate the effectiveness of simple strategies such as video label propagation, copy-pasting, crop-rescaling, and providing object context as well as emphasize the need for better data-scarce algorithms. Our dataset, primarily sourced from smartphones, lacks multiple calibrated cameras seen on self-driving platforms. Not all traversable paths are captured, and 3D perception in work zones remains an open challenge.
Despite these limitations, we hope our dataset inspires further research into this overlooked challenge. 

\noindent \textbf{Acknowledgements:} This work was supported by a research contract from General Motors Research-Israel, NSF Grant CNS-2038612, a US DOT grant 69A3551747111 through the Mobility21 UTC and grants 69A3552344811 and 69A3552348316 through the Safety21 UTC. We thank N. Dinesh Reddy, Shefali Srivastava, Neha Boloor, Tiffany Ma for insightful discussions.






{\small
\bibliographystyle{ieeenat_fullname}
\bibliography{egbib}

@String(CVPR= {IEEE Conf. Comput. Vis. Pattern Recog.})

@String(ICCV= {Int. Conf. Comput. Vis.})

@String(ECCV= {Eur. Conf. Comput. Vis.})

@String(ICLR = {Int. Conf. Learn. Represent.})

@String(CVPR  = {CVPR})

@String(ICCV  = {ICCV})

@String(ECCV  = {ECCV})

@String(ICLR  = {ICLR})

@inproceedings{xu2023open,
  title={Open-vocabulary panoptic segmentation with text-to-image diffusion models},
  author={Xu, Jiarui and Liu, Sifei and Vahdat, Arash and Byeon, Wonmin and Wang, Xiaolong and De Mello, Shalini},
  booktitle={CVPR},
  year={2023}
}

@article{su2011glancing,
  title={Glancing and then looking: on the role of body, affect, and meaning in cognitive control},
  author={Su, Li and Bowman, Howard and Barnard, Philip},
  journal={Frontiers in psychology},
  year={2011},
}

@misc{news-cnn2021cones,
      title  = "Traffic cones confused a Waymo self-driving car. Then things got worse.",
      author = {McFarland, Matt},
      howpublished = "\url{https://www.cnn.com/2021/05/17/tech/waymo-arizona-confused/}",
      year   = "2021"
}

@misc{news-wired2022cruise,
    title="An Autonomous Car Blocked a Fire Truck Responding to an Emergency",
    author= {Marshall, Aarian},
    howpublished = "\url{https://www.wired.com/story/cruise-fire-truck-block-san-francisco-autonomous-vehicles/}",
    year = "2022"
}

@misc{news-atlantic2023sf,
    title="San Francisco Has a Problem With Robotaxis",
    author={Zipper, David},
    howpublished = "\url{https://www.theatlantic.com/ideas/archive/2023/08/robotaxis-san-francisco-self-driving-car/674956/}",
    year = "2023"
}

@misc{news-nyt2023concrete,
    title="Driverless Car Gets Stuck in Wet Concrete in San Francisco",
    author={Levenson, Michael},
    howpublished = "\url{https://www.nytimes.com/2023/08/17/us/driverless-car-accident-sf.html}",
    year = "2023"
}

@misc{news-wired2017construction,
    title="Why Self-Driving Cars *Can't Even* With Construction Zones",
    author={Marshall, Aarian},
    howpublished = "\url{https://www.wired.com/2017/02/self-driving-cars-cant-even-construction-zones/}",
    year = "2017"
}

@misc{news-pntv2023construction,
    title="Robotaxi Causes Traffic Jam After Attempting To Drive Through Construction Site; Elon Musk Responds",
    author={Nation TV, Power},
    howpublished = "\url{https://www.powernationtv.com/post/robotaxi-traffic-elon-musk-responds}",
    year = "2023"
}

@misc{news-sfstandard2023hole,
    title="Driverless Waymo car almost digs itself into hole—literally",
    author={Zigoris, Julie},
    howpublished = "\url{https://sfstandard.com/2023/01/15/driverless-waymo-car-digs-itself-into-hole-literally/}",
    year = "2023"
}

@inproceedings{kalluri2023geonet,
  title={GeoNet: Benchmarking Unsupervised Adaptation across Geographies},
  author={Kalluri, Tarun and Xu, Wangdong and Chandraker, Manmohan},
  booktitle={CVPR},
  year={2023}
}

@inproceedings{wang2020train,
  title={Train in germany, test in the usa: Making 3d object detectors generalize},
  author={Wang, Yan and Chen, Xiangyu and You, Yurong and Li, Li Erran and Hariharan, Bharath and Campbell, Mark and Weinberger, Kilian Q and Chao, Wei-Lun},
  booktitle={CVPR},
  year={2020}
}

@misc{roadbotics-open,
    title="RoadBotics Open Data Set",
    author={Michelin Mobility Intelligence},
    howpublished = "\url{https://www.roadbotics.com/2021/03/15/roadbotics-open-data-set/}",
    year = "2021"
}

@misc{NSCInjuryFacts2023,
author = {{National Safety Council Injury Facts}},
title = {{Motor Vehicle Safety Issues}},
year = {Visited on March 1, 2024},
url = {https://injuryfacts.nsc.org/motor-vehicle/motor-vehicle-safety-issues/work-zones/}
}

@misc{NHTSAPeopleStats2021,
author = {{U.S. Department of Transportation, National Highway Traffic Safety Administration}},
title = {{FARS Data: People Killed in Construction or Maintenance Zones}},
year = {Visited on March 1, 2024},
url = {https://www-fars.nhtsa.dot.gov/People/PeopleAllVictims.aspx}
}

@article{SelfDrivingChallenge2,
author = {Dixit, Vinayak and Chand, Sai and Nair, Divya},
year = {2016},
title = {Autonomous Vehicles: Disengagements, Accidents and Reaction Times},
journal = {PLOS ONE},
}

@article{Morgan2010,
title = {Driver responses to differing urban work zone configurations},
journal = {Accident Analysis \& Prevention},
year = {2010},
author = {J.F. Morgan and A.R. Duley and P.A. Hancock}
}

@article{Wang1996,
author = {Jun Wang and Warren E. Hughes and Forrest M. Council and Jeffrey F. Paniati},
title ={Investigation of Highway Work Zone Crashes: What We Know and What We Don't Know},
journal = {Transportation Research Record},
year = {1996},
}

@article{Thapa2024,
title = {Assessing driver behavior in work zones: A discretized duration approach to predict speeding},
journal = {Accident Analysis \& Prevention},
year = {2024},
author = {Diwas Thapa and Sabyasachee Mishra and Asad Khattak and Muhammad Adeel}
}

@inproceedings{he2017mask,
  title={Mask r-cnn},
  author={He, Kaiming and Gkioxari, Georgia and Doll{\'a}r, Piotr and Girshick, Ross},
  booktitle={ICCV},
  year={2017}
}

@inproceedings{ghiasi2021simple,
  title={Simple copy-paste is a strong data augmentation method for instance segmentation},
  author={Ghiasi, Golnaz and Cui, Yin and Srinivas, Aravind and Qian, Rui and Lin, Tsung-Yi and Cubuk, Ekin D and Le, Quoc V and Zoph, Barret},
  booktitle={CVPR},
  year={2021}
}

@inproceedings{zhou2022detecting,
  title={Detecting twenty-thousand classes using image-level supervision},
  author={Zhou, Xingyi and Girdhar, Rohit and Joulin, Armand and Kr{\"a}henb{\"u}hl, Philipp and Misra, Ishan},
  booktitle={ECCV},
  year={2022},
}

@inproceedings{zhang2023simple,
  title={A simple framework for open-vocabulary segmentation and detection},
  author={Zhang, Hao and Li, Feng and Zou, Xueyan and Liu, Shilong and Li, Chunyuan and Yang, Jianwei and Zhang, Lei},
  booktitle={ICCV},
  year={2023}
}

@inproceedings{li2024llada,
    title={Driving Everywhere with Large Language Model Policy Adaptation},
    author={Li, Boyi and Wang, Yue and Mao, Jiageng and Ivanovic, Boris and Veer, Sushant and Leung, Karen and Pavone, Marco},
    booktitle={CVPR},
    year={2024},
}

@article{kirillov2023segany,
  title={Segment Anything},
  author={Kirillov, Alexander and Mintun, Eric and Ravi, Nikhila and Mao, Hanzi and Rolland, Chloe and Gustafson, Laura and Xiao, Tete and Whitehead, Spencer and Berg, Alexander C. and Lo, Wan-Yen and Doll{\'a}r, Piotr and Girshick, Ross},
  journal={arXiv:2304.02643},
  year={2023}
}

@inproceedings{cheng2022pointly,
  title={Pointly-supervised instance segmentation},
  author={Cheng, Bowen and Parkhi, Omkar and Kirillov, Alexander},
  booktitle={CVPR},
  year={2022}
}

@inproceedings{ronen2022glass,
  title={Glass: Global to local attention for scene-text spotting},
  author={Ronen, Roi and Tsiper, Shahar and Anschel, Oron and Lavi, Inbal and Markovitz, Amir and Manmatha, R},
  booktitle={ECCV},
  year={2022},
}

@inproceedings{li2022blip,
  title={Blip: Bootstrapping language-image pre-training for unified vision-language understanding and generation},
  author={Li, Junnan and Li, Dongxu and Xiong, Caiming and Hoi, Steven},
  booktitle={ICML},
  year={2022},
}

@inproceedings{liu2023llava,
      title={Visual Instruction Tuning}, 
      author={Liu, Haotian and Li, Chunyuan and Wu, Qingyang and Lee, Yong Jae},
      booktitle={NeurIPS},
      year={2023},
}

@article{mao2023gpt,
  title={Gpt-driver: Learning to drive with gpt},
  author={Mao, Jiageng and Qian, Yuxi and Zhao, Hang and Wang, Yue},
  journal={arXiv preprint arXiv:2310.01415},
  year={2023}
}

@article{wang2023drive,
  title={Drive anywhere: Generalizable end-to-end autonomous driving with multi-modal foundation models},
  author={Wang, Tsun-Hsuan and Maalouf, Alaa and Xiao, Wei and Ban, Yutong and Amini, Alexander and Rosman, Guy and Karaman, Sertac and Rus, Daniela},
  journal={arXiv preprint arXiv:2310.17642},
  year={2023}
}

@inproceedings{schoenberger2016sfm,
    author={Sch\"{o}nberger, Johannes Lutz and Frahm, Jan-Michael},
    title={Structure-from-Motion Revisited},
    booktitle={CVPR},
    year={2016},
}

@inproceedings{nuscenes,
  title={nuscenes: A multimodal dataset for autonomous driving},
  author={Caesar, Holger and Bankiti, Varun and Lang, Alex H and Vora, Sourabh and Liong, Venice Erin and Xu, Qiang and Krishnan, Anush and Pan, Yu and Baldan, Giancarlo and Beijbom, Oscar},
  booktitle={CVPR},
  year={2020}
}

@InProceedings{Waymodataset, author = {Sun, Pei and Kretzschmar, Henrik and Dotiwalla, Xerxes and Chouard, Aurelien and Patnaik, Vijaysai and Tsui, Paul and Guo, James and Zhou, Yin and Chai, Yuning and Caine, Benjamin and Vasudevan, Vijay and Han, Wei and Ngiam, Jiquan and Zhao, Hang and Timofeev, Aleksei and Ettinger, Scott and Krivokon, Maxim and Gao, Amy and Joshi, Aditya and Zhang, Yu and Shlens, Jonathon and Chen, Zhifeng and Anguelov, Dragomir}, title = {Scalability in Perception for Autonomous Driving: Waymo Open Dataset}, booktitle = {CVPR}, year = {2020} 
}

@inproceedings{yu2020bdd100k,
  title={Bdd100k: A diverse driving dataset for heterogeneous multitask learning},
  author={Yu, Fisher and Chen, Haofeng and Wang, Xin and Xian, Wenqi and Chen, Yingying and Liu, Fangchen and Madhavan, Vashisht and Darrell, Trevor},
  booktitle={CVPR},
  year={2020}
}

@inproceedings{chang2019argoverse,
  title={Argoverse: 3d tracking and forecasting with rich maps},
  author={Chang, Ming-Fang and Lambert, John and Sangkloy, Patsorn and Singh, Jagjeet and Bak, Slawomir and Hartnett, Andrew and Wang, De and Carr, Peter and Lucey, Simon and Ramanan, Deva and others},
  booktitle={CVPR},
  year={2019}
}

@inproceedings{mathibela2013roadwork,
  title={A roadwork scene signature based on the opponent colour model},
  author={Mathibela, Bonolo and Posner, Ingmar and Newman, Paul},
  booktitle={IROS},
  year={2013},
}

@inproceedings{gumpp2009recognition,
  title={Recognition and tracking of temporary lanes in motorway construction sites},
  author={Gumpp, Thomas and Nienhuser, Dennis and Liebig, Rebecca and Zollner, J Marius},
  booktitle={Intelligent Vehicles Symposium},
  year={2009},
}

@inproceedings{graf2012probabilistic,
  title={Probabilistic estimation of temporary lanes at road work zones},
  author={Graf, Regine and Wimmer, Andreas and Dietmayer, Klaus CJ},
  booktitle={ITSC},
  year={2012},
}

@inproceedings{wimmer2009automatic,
  title={Automatic detection and classification of safety barriers in road construction sites using a laser scanner},
  author={Wimmer, Andreas and Weiss, Thorsten and Flogel, Francesco and Dietmayer, Klaus},
  booktitle={Intelligent Vehicles Symposium},
  year={2009},
}

@inproceedings{pannen2020keep,
  title={How to keep HD maps for automated driving up to date},
  author={Pannen, David and Liebner, Martin and Hempel, Wolfgang and Burgard, Wolfram},
  booktitle={ICRA},
  year={2020},
}

@article{burnett2023boreas,
  title={Boreas: A multi-season autonomous driving dataset},
  author={Burnett, Keenan and Yoon, David J and Wu, Yuchen and Li, Andrew Z and Zhang, Haowei and Lu, Shichen and Qian, Jingxing and Tseng, Wei-Kang and Lambert, Andrew and Leung, Keith YK and others},
  journal={IJRR},
  year={2023},
}

@inproceedings{cordts2016cityscapes,
  title={The cityscapes dataset for semantic urban scene understanding},
  author={Cordts, Marius and Omran, Mohamed and Ramos, Sebastian and Rehfeld, Timo and Enzweiler, Markus and Benenson, Rodrigo and Franke, Uwe and Roth, Stefan and Schiele, Bernt},
  booktitle={CVPR},
  year={2016}
}

@inproceedings{bijelic2020seeing,
  title={Seeing through fog without seeing fog: Deep multimodal sensor fusion in unseen adverse weather},
  author={Bijelic, Mario and Gruber, Tobias and Mannan, Fahim and Kraus, Florian and Ritter, Werner and Dietmayer, Klaus and Heide, Felix},
  booktitle={CVPR},
  year={2020}
}

@inproceedings{sakaridis2021acdc,
  title={ACDC: The adverse conditions dataset with correspondences for semantic driving scene understanding},
  author={Sakaridis, Christos and Dai, Dengxin and Van Gool, Luc},
  booktitle={ICCV},
  year={2021}
}

@inproceedings{sakaridis2019guided,
  title={Guided curriculum model adaptation and uncertainty-aware evaluation for semantic nighttime image segmentation},
  author={Sakaridis, Christos and Dai, Dengxin and Gool, Luc Van},
  booktitle={ICCV},
  year={2019}
}

@inproceedings{neuhold2017mapillary,
  title={The mapillary vistas dataset for semantic understanding of street scenes},
  author={Neuhold, Gerhard and Ollmann, Tobias and Rota Bulo, Samuel and Kontschieder, Peter},
  booktitle={ICCV},
  year={2017}
}

@inproceedings{geiger2012we,
  title={Are we ready for autonomous driving? the kitti vision benchmark suite},
  author={Geiger, Andreas and Lenz, Philip and Urtasun, Raquel},
  booktitle={CVPR},
  year={2012},
}

@inproceedings{WaymoMotion, author={Ettinger, Scott and Cheng, Shuyang and Caine, Benjamin and Liu, Chenxi and Zhao, Hang and Pradhan, Sabeek and Chai, Yuning and Sapp, Ben and Qi, Charles R. and Zhou, Yin and Yang, Zoey and Chouard, Aur'elien and Sun, Pei and Ngiam, Jiquan and Vasudevan, Vijay and McCauley, Alexander and Shlens, Jonathon and Anguelov, Dragomir}, 
title={Large Scale Interactive Motion Forecasting for Autonomous Driving: The Waymo Open Motion Dataset}, 
booktitle= {ICCV}, 
year={2021}
}

@inproceedings{shi2021work,
  title={Work zone detection for autonomous vehicles},
  author={Shi, Weijing and Rajkumar, Ragunathan Raj},
  booktitle={ITSC},
  year={2021},
}

@inproceedings{mathibela2012can,
  title={Can priors be trusted? learning to anticipate roadworks},
  author={Mathibela, Bonolo and Osborne, Michael A and Posner, Ingmar and Newman, Paul},
  booktitle={ITSC},
  year={2012},
}

@InProceedings{hu2023planning,
    author    = {Hu, Yihan and Yang, Jiazhi and Chen, Li and Li, Keyu and Sima, Chonghao and Zhu, Xizhou and Chai, Siqi and Du, Senyao and Lin, Tianwei and Wang, Wenhai and Lu, Lewei and Jia, Xiaosong and Liu, Qiang and Dai, Jifeng and Qiao, Yu and Li, Hongyang},
    title     = {Planning-Oriented Autonomous Driving},
    booktitle = {CVPR},
    year      = {2023},
}

@inproceedings{kennerley20232pcnet,
  title={2PCNet: Two-Phase Consistency Training for Day-to-Night Unsupervised Domain Adaptive Object Detection},
  author={Kennerley, Mikhail and Wang, Jian-Gang and Veeravalli, Bharadwaj and Tan, Robby T},
  booktitle={CVPR},
  year={2023}
}

@inproceedings{varma2019idd,
  title={IDD: A dataset for exploring problems of autonomous navigation in unconstrained environments},
  author={Varma, Girish and Subramanian, Anbumani and Namboodiri, Anoop and Chandraker, Manmohan and Jawahar, CV},
  booktitle={WACV},
  year={2019},
}

@inproceedings{zheng2024instance,
  title={Instance-Warp: Saliency Guided Image Warping for Unsupervised Domain Adaptation},
  author={Zheng, Shen and Ghosh, Anurag and Narasimhan, Srinivasa G},
  booktitle={WACV},
  year={2025}
}

@inproceedings{hoyer2022daformer,
  title={Daformer: Improving network architectures and training strategies for domain-adaptive semantic segmentation},
  author={Hoyer, Lukas and Dai, Dengxin and Van Gool, Luc},
  booktitle={CVPR},
  year={2022}
}

@inproceedings{singh2021textocr,
  title={Textocr: Towards large-scale end-to-end reasoning for arbitrary-shaped scene text},
  author={Singh, Amanpreet and Pang, Guan and Toh, Mandy and Huang, Jing and Galuba, Wojciech and Hassner, Tal},
  booktitle={CVPR},
  year={2021}
}

@inproceedings{radford2021learning,
  title={Learning transferable visual models from natural language supervision},
  author={Radford, Alec and Kim, Jong Wook and Hallacy, Chris and Ramesh, Aditya and Goh, Gabriel and Agarwal, Sandhini and Sastry, Girish and Askell, Amanda and Mishkin, Pamela and Clark, Jack and others},
  booktitle={ICML},
  year={2021},
}

@inproceedings{hu2022lora,
  title={LoRA: Low-Rank Adaptation of Large Language Models},
  author={Hu, Edward J and Wallis, Phillip and Allen-Zhu, Zeyuan and Li, Yuanzhi and Wang, Shean and Wang, Lu and Chen, Weizhu and others},
  booktitle={ICLR},
  year={2022}
}

@inproceedings{vuong2024walt3d,
  title={WALT3D: Generating Realistic Training Data from Time-Lapse Imagery for Reconstructing Dynamic Objects Under Occlusion},
  author={Vuong, Khiem and Reddy, N Dinesh and Tamburo, Robert and Narasimhan, Srinivasa G.},
  booktitle={CVPR},
  year={2024}
}

@incollection{visvalingam2017line,
  title={Line generalization by repeated elimination of points},
  author={Visvalingam, Maheswari and Whyatt, James D},
  booktitle={Landmarks in Mapping},
  year={2017},
}

@inproceedings{cheng2022masked,
  title={Masked-attention mask transformer for universal image segmentation},
  author={Cheng, Bowen and Misra, Ishan and Schwing, Alexander G and Kirillov, Alexander and Girdhar, Rohit},
  booktitle={CVPR},
  year={2022}
}

@inproceedings{mangalam2021goals,
  title={From goals, waypoints \& paths to long term human trajectory forecasting},
  author={Mangalam, Karttikeya and An, Yang and Girase, Harshayu and Malik, Jitendra},
  booktitle={ICCV},
  year={2021}
}

@inproceedings{kim2018textual,
  title={Textual explanations for self-driving vehicles},
  author={Kim, Jinkyu and Rohrbach, Anna and Darrell, Trevor and Canny, John and Akata, Zeynep},
  booktitle={ECCV},
  year={2018}
}

@inproceedings{xu2020explainable,
  title={Explainable object-induced action decision for autonomous vehicles},
  author={Xu, Yiran and Yang, Xiaoyin and Gong, Lihang and Lin, Hsuan-Chu and Wu, Tz-Ying and Li, Yunsheng and Vasconcelos, Nuno},
  booktitle={CVPR},
  year={2020}
}

@misc{ferguson2015construction,
  title={Construction zone object detection using light detection and ranging},
  author={Ferguson, David Ian and Haehnel, Dirk and Mahon, Ian},
  year={2015},
  note={US Patent 9,199,641}
}

@misc{ferguson2015mapping,
  title={Mapping active and inactive construction zones for autonomous driving},
  author={Ferguson, David I and Burnette, Donald Jason},
  year={2015},
  note={US Patent 9,141,107}
}

@article{ferguson2008motion,
  title={Motion planning in urban environments},
  author={Ferguson, Dave and Howard, Thomas M and Likhachev, Maxim},
  journal={Journal of Field Robotics},
  year={2008},
}

@inproceedings{lu2014layered,
  title={Layered costmaps for context-sensitive navigation},
  author={Lu, David V and Hershberger, Dave and Smart, William D},
  booktitle={IROS},
  year={2014},
}

@inproceedings{liang2020garden,
  title={The garden of forking paths: Towards multi-future trajectory prediction},
  author={Liang, Junwei and Jiang, Lu and Murphy, Kevin and Yu, Ting and Hauptmann, Alexander},
  booktitle={CVPR},
  year={2020}
}

@inproceedings{salzmann2020trajectron,
  title={Trajectron++: Dynamically-feasible trajectory forecasting with heterogeneous data},
  author={Salzmann, Tim and Ivanovic, Boris and Chakravarty, Punarjay and Pavone, Marco},
  booktitle={ECCV},
  year={2020},
}

@inproceedings{gupta2018social,
  title={Social gan: Socially acceptable trajectories with generative adversarial networks},
  author={Gupta, Agrim and Johnson, Justin and Fei-Fei, Li and Savarese, Silvio and Alahi, Alexandre},
  booktitle={CVPR},
  year={2018}
}

@article{teng2023motion,
  title={Motion planning for autonomous driving: The state of the art and future perspectives},
  author={Teng, Siyu and Hu, Xuemin and Deng, Peng and Li, Bai and Li, Yuchen and Ai, Yunfeng and Yang, Dongsheng and Li, Lingxi and Xuanyuan, Zhe and Zhu, Fenghua and others},
  journal={T-IV},
  year={2023}
}

@inproceedings{mcnaughton2011motion,
  title={Motion planning for autonomous driving with a conformal spatiotemporal lattice},
  author={McNaughton, Matthew and Urmson, Chris and Dolan, John M and Lee, Jin-Woo},
  booktitle={ICRA},
  year={2011}
}

@inproceedings{dauner2023parting,
  title={Parting with misconceptions about learning-based vehicle motion planning},
  author={Dauner, Daniel and Hallgarten, Marcel and Geiger, Andreas and Chitta, Kashyap},
  booktitle={CoRL},
  year={2023}
}

@article{claussmann2019review,
  title={A review of motion planning for highway autonomous driving},
  author={Claussmann, Laurene and Revilloud, Marc and Gruyer, Dominique and Glaser, S{\'e}bastien},
  journal={T-ITS},
  year={2019}
}

@article{tian2024drivevlm,
  title={DriveVLM: The Convergence of Autonomous Driving and Large Vision-Language Models},
  author={Tian, Xiaoyu and Gu, Junru and Li, Bailin and Liu, Yicheng and Hu, Chenxu and Wang, Yang and Zhan, Kun and Jia, Peng and Lang, Xianpeng and Zhao, Hang},
  journal={arXiv preprint arXiv:2402.12289},
  year={2024}
}

@inproceedings{chng2019icdar2019,
  title={Icdar2019 robust reading challenge on arbitrary-shaped text-rrc-art},
  author={Chng, Chee Kheng and Liu, Yuliang and Sun, Yipeng and Ng, Chun Chet and Luo, Canjie and Ni, Zihan and Fang, ChuanMing and Zhang, Shuaitao and Han, Junyu and Ding, Errui and others},
  booktitle={ICDAR},
  year={2019},
}

@article{mmdetection,
  title   = {{MMDetection}: Open MMLab Detection Toolbox and Benchmark},
  author  = {Chen, Kai and Wang, Jiaqi and Pang, Jiangmiao and Cao, Yuhang and
             Xiong, Yu and Li, Xiaoxiao and Sun, Shuyang and Feng, Wansen and
             Liu, Ziwei and Xu, Jiarui and Zhang, Zheng and Cheng, Dazhi and
             Zhu, Chenchen and Cheng, Tianheng and Zhao, Qijie and Li, Buyu and
             Lu, Xin and Zhu, Rui and Wu, Yue and Dai, Jifeng and Wang, Jingdong
             and Shi, Jianping and Ouyang, Wanli and Loy, Chen Change and Lin, Dahua},
  journal= {arXiv preprint arXiv:1906.07155},
  year={2019}
}

@inproceedings{lin2014microsoft,
  title={Microsoft coco: Common objects in context},
  author={Lin, Tsung-Yi and Maire, Michael and Belongie, Serge and Hays, James and Perona, Pietro and Ramanan, Deva and Doll{\'a}r, Piotr and Zitnick, C Lawrence},
  booktitle={ECCV},
  year={2014},
}

@article{ravi2024sam2,
  title={SAM 2: Segment Anything in Images and Videos},
  author={Ravi, Nikhila and Gabeur, Valentin and Hu, Yuan-Ting and Hu, Ronghang and Ryali, Chaitanya and Ma, Tengyu and Khedr, Haitham and R{\"a}dle, Roman and Rolland, Chloe and Gustafson, Laura and Mintun, Eric and Pan, Junting and Alwala, Kalyan Vasudev and Carion, Nicolas and Wu, Chao-Yuan and Girshick, Ross and Doll{\'a}r, Piotr and Feichtenhofer, Christoph},
  journal={arXiv preprint arXiv:2408.00714},
  url={https://arxiv.org/abs/2408.00714},
  year={2024}
}

@inproceedings{cheng2021boundary,
  title={Boundary IoU: Improving object-centric image segmentation evaluation},
  author={Cheng, Bowen and Girshick, Ross and Doll{\'a}r, Piotr and Berg, Alexander C and Kirillov, Alexander},
  booktitle={CVPR},
  year={2021}
}

@inproceedings{kim2024rosa,
  title={RoSA Dataset: Road construct zone Segmentation for Autonomous Driving},
  author={Kim, Jinwoo and An, Kyounghwan and Lee, Donghwan},
  booktitle={ECCV 2024 Workshop on Multimodal Perception and Comprehension of Corner Cases in Autonomous Driving},
  year={2024}
}

@inproceedings{liang2024aide,
  title={Aide: An automatic data engine for object detection in autonomous driving},
  author={Liang, Mingfu and Su, Jong-Chyi and Schulter, Samuel and Garg, Sparsh and Zhao, Shiyu and Wu, Ying and Chandraker, Manmohan},
  booktitle={CVPR},
  year={2024}
}

@article{chan2021segmentmeifyoucan,
  title={SegmentMeIfYouCan: A Benchmark for Anomaly Segmentation},
  author={Chan, Robin Kien-Wei and Lis, Krzysztof and Uhlemeyer, Svenja and Blum, Hermann and Honari, Sina and Siegwart, Roland and Fua, Pascal and Salzmann, Mathieu and Rottmann, Matthias},
  journal={NeurIPS Track on Datasets and Benchmarks},
  year={2021}
}

@inproceedings{li2022coda,
  title={Coda: A real-world road corner case dataset for object detection in autonomous driving},
  author={Li, Kaican and Chen, Kai and Wang, Haoyu and Hong, Lanqing and Ye, Chaoqiang and Han, Jianhua and Chen, Yukuai and Zhang, Wei and Xu, Chunjing and Yeung, Dit-Yan and others},
  booktitle={ECCV},
  year={2022},
}

@inproceedings{zendel2018wilddash,
  title={Wilddash-creating hazard-aware benchmarks},
  author={Zendel, Oliver and Honauer, Katrin and Murschitz, Markus and Steininger, Daniel and Dominguez, Gustavo Fernandez},
  booktitle={ECCV},
  year={2018}
}

@inproceedings{zendel2022wilddash2,
  title={Unifying panoptic segmentation for autonomous driving},
  author={Zendel, Oliver and Sch{\"o}rghuber, Matthias and Rainer, Bernhard and Murschitz, Markus and Beleznai, Csaba},
  booktitle={CVPR},
  year={2022}
}

@article{hendrycks2019bddanomaly,
  title={Scaling out-of-distribution detection for real-world settings},
  author={Hendrycks, Dan and Basart, Steven and Mazeika, Mantas and Zou, Andy and Kwon, Joe and Mostajabi, Mohammadreza and Steinhardt, Jacob and Song, Dawn},
  journal={arXiv preprint arXiv:1911.11132},
  year={2019}
}

@inproceedings{chen2023diffusiondet,
  title={Diffusiondet: Diffusion model for object detection},
  author={Chen, Shoufa and Sun, Peize and Song, Yibing and Luo, Ping},
  booktitle={ICCV},
  year={2023}
}

@inproceedings{zhangdino,
  title={DINO: DETR with Improved DeNoising Anchor Boxes for End-to-End Object Detection},
  author={Zhang, Hao and Li, Feng and Liu, Shilong and Zhang, Lei and Su, Hang and Zhu, Jun and Ni, Lionel and Shum, Heung-Yeung},
  booktitle={ICLR},
  year={2023}
}

@misc{li2024llavanext-strong,
    title={LLaVA-NeXT: Stronger LLMs Supercharge Multimodal Capabilities in the Wild},
    url={https://llava-vl.github.io/blog/2024-05-10-llava-next-stronger-llms/},
    author={Li, Bo and Zhang, Kaichen and Zhang, Hao and Guo, Dong and Zhang, Renrui and Li, Feng and Zhang, Yuanhan and Liu, Ziwei and Li, Chunyuan},
    month={May},
    year={2024}
}

@inproceedings{wang2023detecting,
  title={Detecting everything in the open world: Towards universal object detection},
  author={Wang, Zhenyu and Li, Yali and Chen, Xi and Lim, Ser-Nam and Torralba, Antonio and Zhao, Hengshuang and Wang, Shengjin},
  booktitle={CVPR},
  year={2023}
}

@inproceedings{zhao2020object,
  title={Object detection with a unified label space from multiple datasets},
  author={Zhao, Xiangyun and Schulter, Samuel and Sharma, Gaurav and Tsai, Yi-Hsuan and Chandraker, Manmohan and Wu, Ying},
  booktitle={ECCV},
  year={2020},
}

@misc{news-2024missionlocal,
    title="Waymo rolls toward San Francisco Airport. A showdown is brewing.",
    author={Eskenazi, Joe},
    howpublished = "\url{https://missionlocal.org/2024/12/waymo-rolls-toward-san-francisco-airport-showdown-brewing/}",
    year = "2024"
}

@inproceedings{liu2024improved,
  title={Improved baselines with visual instruction tuning},
  author={Liu, Haotian and Li, Chunyuan and Li, Yuheng and Lee, Yong Jae},
  booktitle={CVPR},
  year={2024}
}

@misc{tesla-patent,
  title={Clip search with multimodal queries},
  author={Wilson, Matthew and Zaman, Tim and Tran, Long},
  url={https://patents.google.com/patent/US20240419724A1},
  year={2025},
  note={Tesla, US Patent US20240419724A1}
}

@misc{cruise-long-tail,
  title={Cruise’s Continuous Learning Machine Predicts the Unpredictable on San Francisco Roads},
  author={Harris, Sean},
  url={https://medium.com/cruise/cruise-continuous-learning-machine-30d60f4c691b},
  year={2020},
}

@inproceedings{li2020towards,
  title={Towards streaming perception},
  author={Li, Mengtian and Wang, Yu-Xiong and Ramanan, Deva},
  booktitle={ECCV},
  year={2020},
}

@inproceedings{huang2017speed,
  title={Speed/accuracy trade-offs for modern convolutional object detectors},
  author={Huang, Jonathan and Rathod, Vivek and Sun, Chen and Zhu, Menglong and Korattikara, Anoop and Fathi, Alireza and Fischer, Ian and Wojna, Zbigniew and Song, Yang and Guadarrama, Sergio and others},
  booktitle={CVPR},
  year={2017}
}

@inproceedings{liu2024grounding,
  title={Grounding dino: Marrying dino with grounded pre-training for open-set object detection},
  author={Liu, Shilong and Zeng, Zhaoyang and Ren, Tianhe and Li, Feng and Zhang, Hao and Yang, Jie and Jiang, Qing and Li, Chunyuan and Yang, Jianwei and Su, Hang and others},
  booktitle={ECCV},
  year={2024}
}

@inproceedings{li2023maskdino,
  title={Mask dino: Towards a unified transformer-based framework for object detection and segmentation},
  author={Li, Feng and Zhang, Hao and Xu, Huaizhe and Liu, Shilong and Zhang, Lei and Ni, Lionel M and Shum, Heung-Yeung},
  booktitle={CVPR},
  year={2023}
}

@article{ren2016faster,
  title={Faster R-CNN: Towards real-time object detection with region proposal networks},
  author={Ren, Shaoqing and He, Kaiming and Girshick, Ross and Sun, Jian},
  journal={TPAMI},
  year={2016},
}

@article{ilharco2022patching,
  title={Patching open-vocabulary models by interpolating weights},
  author={Ilharco, Gabriel and Wortsman, Mitchell and Gadre, Samir Yitzhak and Song, Shuran and Hajishirzi, Hannaneh and Kornblith, Simon and Farhadi, Ali and Schmidt, Ludwig},
  journal={NeurIPS},
  year={2022}
}

@article{chen2015microsoft,
  title={Microsoft coco captions: Data collection and evaluation server},
  author={Chen, Xinlei and Fang, Hao and Lin, Tsung-Yi and Vedantam, Ramakrishna and Gupta, Saurabh and Doll{\'a}r, Piotr and Zitnick, C Lawrence},
  journal={arXiv preprint arXiv:1504.00325},
  year={2015}
}

@article{hurst2024gpt4o,
  title={Gpt-4o system card},
  author={Hurst, Aaron and Lerer, Adam and Goucher, Adam P and Perelman, Adam and Ramesh, Aditya and Clark, Aidan and Ostrow, AJ and Welihinda, Akila and Hayes, Alan and Radford, Alec and others},
  journal={arXiv preprint arXiv:2410.21276},
  year={2024}
}

@article{comanici2025gemini,
  title={Gemini 2.5: Pushing the Frontier with Advanced Reasoning, Multimodality, Long Context, and Next Generation Agentic Capabilities},
  author={Comanici, Gheorghe and Bieber, Eric and Schaekermann, Mike and Pasupat, Ice and Sachdeva, Noveen and Dhillon, Inderjit and Blistein, Marcel and Ram, Ori and Zhang, Dan and Rosen, Evan and others},
  journal={arXiv preprint arXiv:2507.06261},
  year={2025}
}

@article{zhang2024omgllava,
  title={Omg-llava: Bridging image-level, object-level, pixel-level reasoning and understanding},
  author={Zhang, Tao and Li, Xiangtai and Fei, Hao and Yuan, Haobo and Wu, Shengqiong and Ji, Shunping and Loy, Chen Change and Yan, Shuicheng},
  journal={NeurIPS},
  year={2024}
}

@article{ramachandran2025well,
  title={How Well Does GPT-4o Understand Vision? Evaluating Multimodal Foundation Models on Standard Computer Vision Tasks},
  author={Ramachandran, Rahul and Garjani, Ali and Bachmann, Roman and Atanov, Andrei and Kar, O{\u{g}}uzhan Fatih and Zamir, Amir},
  journal={arXiv preprint arXiv:2507.01955},
  year={2025}
}

@inproceedings{klingner2013street,
  title={Street view motion-from-structure-from-motion},
  author={Klingner, Bryan and Martin, David and Roseborough, James},
  booktitle={ICCV},
  year={2013}
}

@inproceedings{berton2023eigenplaces,
  title={Eigenplaces: Training viewpoint robust models for visual place recognition},
  author={Berton, Gabriele and Trivigno, Gabriele and Caputo, Barbara and Masone, Carlo},
  booktitle={ICCV},
  year={2023}
}

@inproceedings{pan2024global,
  title={Global structure-from-motion revisited},
  author={Pan, Linfei and Bar{\'a}th, D{\'a}niel and Pollefeys, Marc and Sch{\"o}nberger, Johannes L},
  booktitle={ECCV},
  year={2024},
}

@inproceedings{lindenberger2023lightglue,
  title={Lightglue: Local feature matching at light speed},
  author={Lindenberger, Philipp and Sarlin, Paul-Edouard and Pollefeys, Marc},
  booktitle={ICCV},
  year={2023}
}

@inproceedings{vuong2024toward,
  title={Toward planet-wide traffic camera calibration},
  author={Vuong, Khiem and Tamburo, Robert and Narasimhan, Srinivasa G.},
  booktitle={WACV},
  year={2024}
}

@inproceedings{vuong2025aerialmegadepth,
  title={AerialMegaDepth: Learning Aerial-Ground Reconstruction and View Synthesis},
  author={Vuong, Khiem and Ghosh, Anurag and Ramanan, Deva and Narasimhan, Srinivasa and Tulsiani, Shubham},
  booktitle={CVPR},
  year={2025}
}

@inproceedings{berton2025megaloc,
  title={Megaloc: One retrieval to place them all},
  author={Berton, Gabriele and Masone, Carlo},
  booktitle={CVPR},
  year={2025}
}

@article{douze2024faiss,
  title={The faiss library},
  author={Douze, Matthijs and Guzhva, Alexandr and Deng, Chengqi and Johnson, Jeff and Szilvasy, Gergely and Mazar{\'e}, Pierre-Emmanuel and Lomeli, Maria and Hosseini, Lucas and J{\'e}gou, Herv{\'e}},
  journal={arXiv preprint arXiv:2401.08281},
  year={2024}
}

@inproceedings{detone2018superpoint,
  title={Superpoint: Self-supervised interest point detection and description},
  author={DeTone, Daniel and Malisiewicz, Tomasz and Rabinovich, Andrew},
  booktitle={Proceedings of the IEEE conference on computer vision and pattern recognition workshops},
  pages={224--236},
  year={2018}
}
}

\newpage
\clearpage
\newpage

\appendix


\section{Related Works}
\label{sup:related_work}

\begin{figure}[t]
    \centering 
    \includegraphics[width=\linewidth]{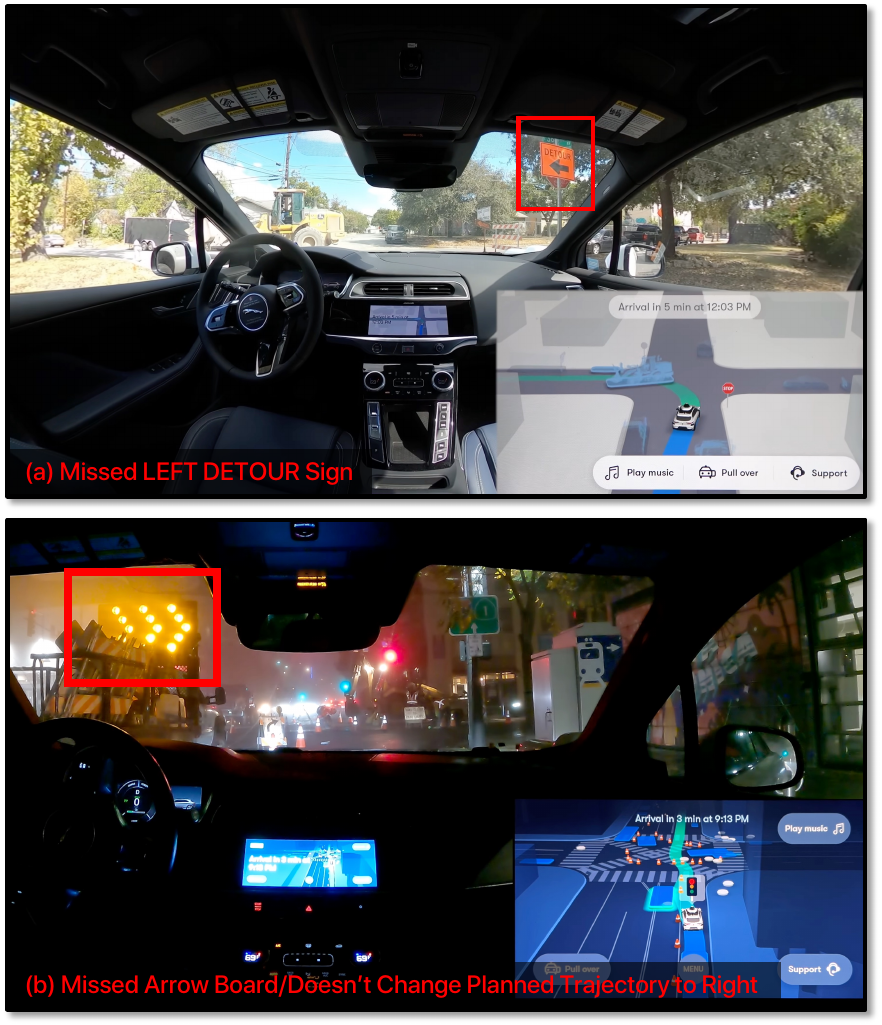}
    \vspace{-0.1in}
    \captionof{figure}{\small \textbf{Examples of work zone failures in a commercial self-driving vehicle.} While obtaining detailed failure reports of self-driving cars is infeasible, customers of these companies regularly post failure cases on social media. (a) The car failed to recognize and observe a sign that mentions ``DETOUR'' and has a left arrow graphic (\href{https://www.youtube.com/watch?v=2ihN0IkIMPg}{Link}). (b) The car fails to recognize and observe the Arrow Board, then fails to analyze the situation and finally does not change the predicted pathway in response (\href{https://www.youtube.com/watch?v=wWZGZWuUx-Y}{Link}).} 
    \label{fig:waymo-errors}
\end{figure}

\noindent
\textbf{Long-Tail Scenarios in Autonomous Driving.}  Driving datasets have evolved from KITTI~\cite{geiger2012we} to more diverse collections like BDD100K~\cite{yu2020bdd100k}, nuScenes~\cite{nuscenes}, Mapillary~\cite{neuhold2017mapillary} and Cityscapes~\cite{cordts2016cityscapes}, incorporating advanced sensor suites~\cite{chang2019argoverse, Waymodataset, WaymoMotion}. However, these datasets provide limited representation of long-tailed scenarios such as work zones - for instance, nuScenes contains only 19 driven sequences with work zones~\cite{shi2021work} out of 1000 scenes. Commercial self-driving vehicle deployments, while impressive in common situations, also find it difficult to navigate work zones, see Figure~\ref{fig:waymo-errors} for some failure examples collected from social media.

\begin{figure*}[t]
\centering
\includegraphics[width=\textwidth]{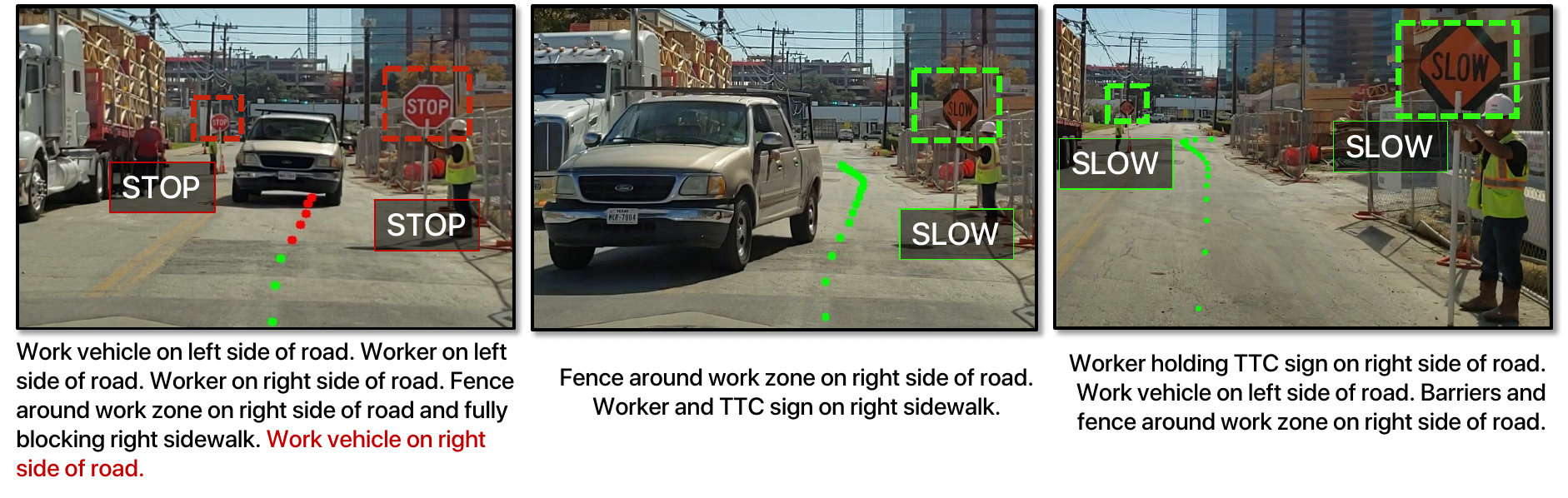}
\captionof{figure}{\small \textbf{Recognition is Not Enough for Navigating Work Zones.} Work zones are dynamic and rare occurrences, thus it is challenging to navigate through them. Depicted is a work zone navigation sequence with sign text detected by Glass~\cite{ronen2022glass}, work zone descriptions generated by fine-tuned LLaVA-1.5A~\cite{liu2024improved} (incorrect description indicated in red) and car trajectory estimated via COLMAP~\cite{schoenberger2016sfm}. Observe that initially the worker is holding a ``STOP'' sign, but later switches to a ``SLOW'' sign as the truck passes, indicating that the road is open for traversal by the ego-vehicle. \textit{This example shows mere object recognition is not enough for navigation; continuous fine grained scene observation and global scene analysis are both necessary.}}
\label{fig:observe-analyze-motivation}
\end{figure*}

Prior research on long-tailed driving scenarios has largely focused on scene understanding. Datasets like CODA~\cite{li2022coda} (with 1500 scenes containing long-tailed objects), WildDash~\cite{zendel2018wilddash, zendel2022wilddash2} (with global weather and lighting variations), SegmentMeIfYouCan~\cite{chan2021segmentmeifyoucan}, and BDD-Anomaly~\cite{hendrycks2019bddanomaly} focus almost exclusively on recognition, rather than holistically addressing perception and navigation in scenarios like work zones. Another well-studied long-tailed scenario is driving in adverse weather. Despite data collection challenges, specialized datasets exist for fog~\cite{bijelic2020seeing, sakaridis2021acdc}, night~\cite{yu2020bdd100k, sakaridis2019guided}, and snow~\cite{bijelic2020seeing, burnett2023boreas}, although these also primarily target recognition. Figure~\ref{fig:observe-analyze-motivation} illustrates why recognition alone is insufficient for self-driving in work zones. 

Work zones are complex, dynamic environments requiring multi-level understanding, yet they've received little attention due to the challenges in data mining~\cite{liang2024aide} and task formulation~\cite{shi2021work}. To our best knowledge, no large-scale public dataset has specifically addressed work zones before our contribution. While the MMI Open Dataset~\cite{roadbotics-open} provides raw videos collected for road inspection, we develop scenario taxonomies and annotated work zones to create the ROADWork Dataset.

\noindent \textbf{Work Zones in Autonomous Driving.} Prior research has addressed isolated work zone edge cases. For example,~\cite{gumpp2009recognition} recognize safety barriers using a laser scanner  while~\cite{graf2012probabilistic} attempt to determine which lane lines define a valid lane in work zones. Later works~\cite{mathibela2013roadwork, shi2021work} attempted to classify and localize work zones, while others updated HD maps~\cite{mathibela2012can, pannen2020keep} with additional work zone information. Concurrent work~\cite{kim2024rosa} has proposed segmenting construction areas in videos to detect continuous zones from a distance. However, no prior work systematically categorizes work zones, formulates tasks, or curates data for autonomous driving in these environments.

\noindent
\textbf{Language and Navigation in Work Zones.} Unseen scenarios, such as newly appearing work zones along a route, pose a major challenge for autonomous driving. Work zones are a classic example of navigation in open-ended driving scenes, requiring a higher level of semantic generalization. Linguistic representations can help generalization, enabling introspective explanations~\cite{kim2018textual} that improve action predictions~\cite{xu2020explainable}.

Recently, Vision-Language Models (VLMs)~\cite{liu2023llava} (and Large Language Models (LLMs)) have been increasingly applied to scene understanding, demonstrating state-of-the-art generalization and reasoning capabilities. Recent efforts~\cite{wang2023drive, mao2023gpt, tian2024drivevlm, li2024llada} have leveraged these VLMs and LLMs to redefine scene understanding and subsequently, motion planning. Navigating work zones require both visuospatial and linguistic abilities. To address this, we propose a work zone description benchmark to aid global scene understanding in workzones.

For navigation in work zones, we argue that long horizon trajectory forecasting is essential, as traditional structural cues like lanes may be unreliable. Prior works~\cite{mangalam2021goals, liang2020garden, gupta2018social} explored a related setting: long horizon human trajectory forecasting. Inspired by this line of work, we propose a new pathway prediction problem and baselines to address tackle this challenge.

\section{\txtdeepred{ROADWork} Dataset Description}
\label{sup:dataset}

\begin{figure*}[t]
    \centering
    \includegraphics[width=0.95\textwidth]{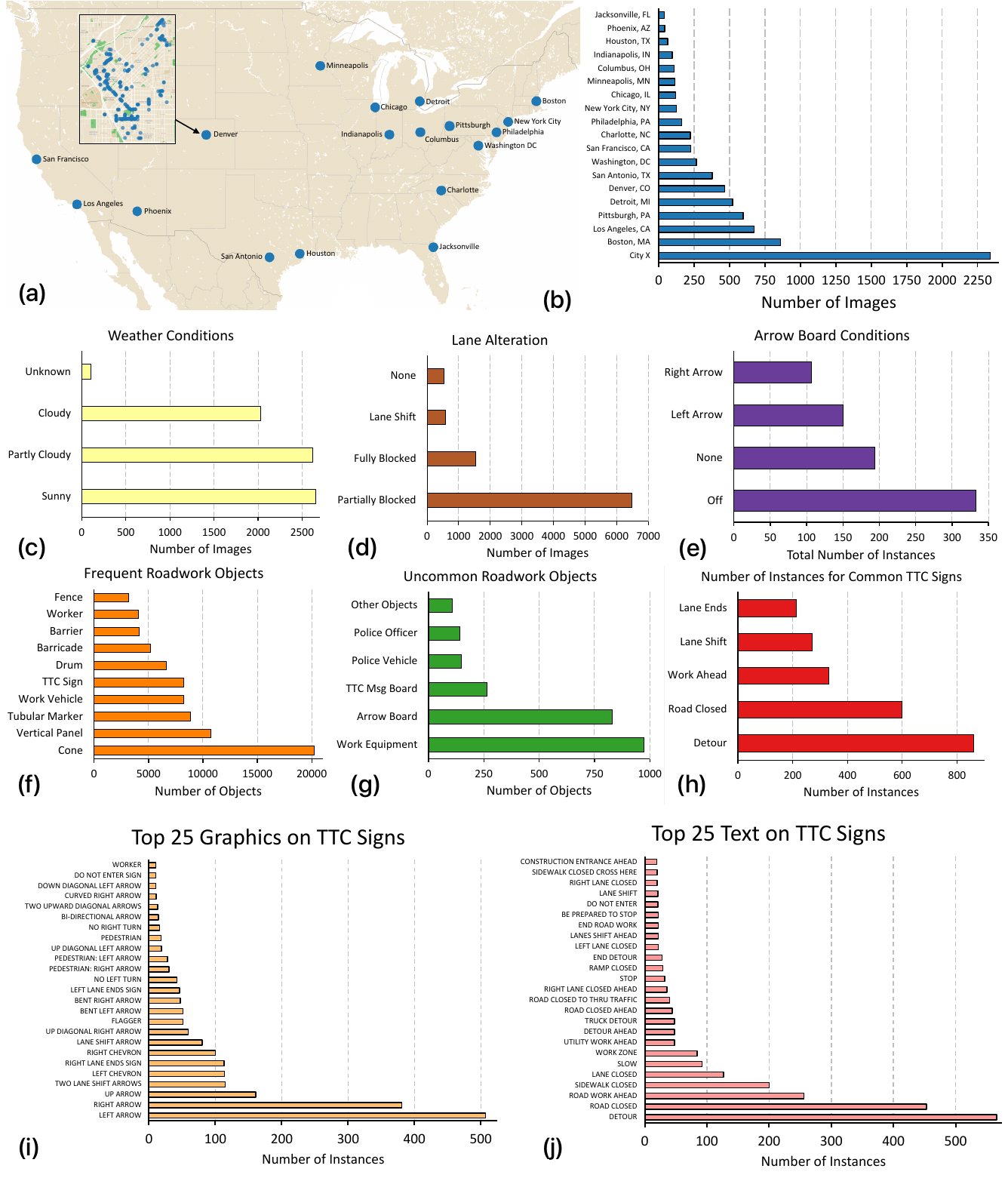}
    \caption{\small \textbf{\emphdatasetname Dataset Statistics.} (a) U.S. cities represented in the dataset, with geotagged images shown for Denver, Colorado. (b) Number of dataset images for each city. (c) Distribution of weather conditions. (d) Distribution of road-network alterations for work zones. (e) Arrow board conditions, where ``None'' indicates that the arrow board's LEDs are not visible. (f) Distribution of frequent roadwork objects, which are of the order of thousands of total instances. (g) Distribution of uncommon roadwork objects which have a few hundred instances. (h) Distribution of the most common TTC signs (both text and graphics), which have a few hundred instances each. (i-j) Distribution of the top 25 observed TTC signs by graphics and text.}
\label{fig:img_stats}
\end{figure*}

\noindent
We describe specific information regarding annotations protocol, data cleaning and processing procedures and other relevant details.

\subsection{Image Acquisition}

\noindent Visual data were acquired from cameras mounted inside a vehicle while driving through 18 US cities, resulting in 9650 images from three sources: \textit{(a)} images that we captured in Pittsburgh \textit{(b)} images that were semi-automatically extracted from the Michelin Mobility Intelligence (MMI) Open Dataset (formerly RoadBotics)~\cite{roadbotics-open} \textit{(c)} Images that were discovered in Mapillary~\cite{neuhold2017mapillary}, BDD100K~\cite{yu2020bdd100k} and other other data sources by our models trained on data from the first two sources.  

\noindent
\textbf{Main Data Sources.} To collect the first data subset of the main dataset, we drove on urban, suburban, and rural roads in Pittsburgh and captured 2,338 ($32\%$) images with an iPhone 14 Pro Max paired with a Bluetooth remote trigger. Next, images from other U.S. cities were sourced from videos in the MMI Open Dataset. A combination of Detic~\cite{zhou2022detecting} and a cone detector trained on NuScenes~\cite{nuscenes} were used to mine frames presumed to contain roadwork zones with detector confidence at $25\%$ -- ensuring high recall with the expense of low precision. This process yielded approximately 100000 candidate images. We then manually selected 5078 (68\%) images containing unique road objects or roadwork zones, prioritizing individual scene diversity. The distribution of images across U.S. cities is shown in Figure~\ref{fig:img_stats}.

\begin{figure*}[t]
    \centering
    \includegraphics[width=\textwidth]{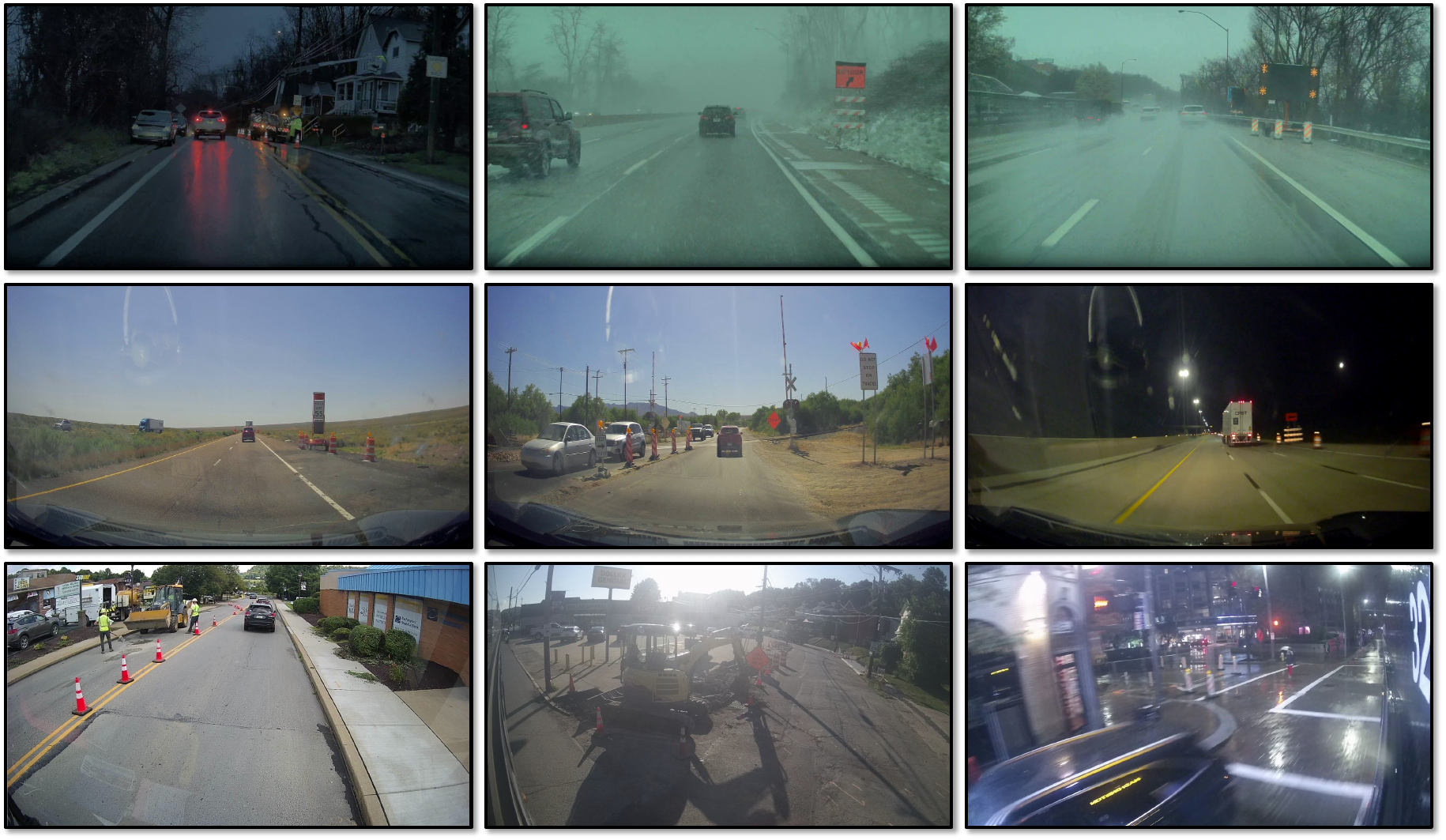}
    \caption{\small \textbf{ROADWork: Additional In-The-Wild Discovered and Annotated Work Zone Images.} Following the discovery process described in Section\pointToMain{~\ref{subsec:recognize}}{~3} of the main manuscript, we discovered and annotated an additional 1265 work zones images from a variety of sources apart from 969 images discovered in BDD100K~\cite{yu2020bdd100k} and Mapillary~\cite{neuhold2017mapillary}. \textbf{(a)} The top row depicts Vehicle-Pittsburgh subset images we discovered from driving in Pittsburgh (around 157 images). The subset consists of work zone images taken in bad weather and night. \textbf{(b)} The middle row  depicts Vehicle-Rural subset images we driving on rural areas and highways in the US (308 images). The subset consists of work zone images taken in both day and night. \textbf{(c)} The bottom row depicts images discovered from a Bus that was driven on a fixed route in Pittsburgh. We discovered 272 images captured from the front camera, while 528 images were captured from other cameras mounted on the bus. Images were captured in all conditions, including bad weather and night.}
    \label{fig:discovered-arizona-tesla}
\end{figure*}

\noindent \textbf{Discovered Data Sources.} In Section 3 of the main manuscript, we described our model and the work zone classification rule that we use to discover images. We discovered work zone images from common driving datasets (a) 558 images from Mappilary~\cite{neuhold2017mapillary} and (b) 411 images from BDD100K~\cite{yu2020bdd100k}. Additionally, we exploit other data sources to curate 1265 images into various subsets containing work zones (See Figure~\ref{fig:discovered-arizona-tesla} for examples). These subsets were further manually filtered to remove redundant images. We describe the subsets below,

\begin{itemize}
\item \textbf{Vehicle-Pittsburgh Discovered Subset.} We drove a vehicle in Pittsburgh during various weather and lighting conditions, collecting approximately 157 images with work zones. This subset specifically includes examples captured in rain, fog, snow, and at night.
\item \textbf{Vehicle-Rural Discovered Subset.} We collected 308 images by driving on rural roads and highways across multiple U.S. states. This subset includes work zones on two-lane roads, interstate highways, and in small towns, captured during both day and night conditions. This subset was captured using a dashcam, and shows significant radial distortion.
\item \textbf{Bus-Pittsburgh Discovered Subset.} We obtained 800 work zone images from a commuter bus that followed a fixed route in Pittsburgh over the course of two years. This includes 272 images from the front-facing camera and 528 images from side-mounted cameras with unqiue viewpoints, capturing work zones in all weather conditions and times of day. 
\end{itemize}

\subsection{Annotations}

\noindent \textbf{Scene Tags.} We labeled images with scene tags to capture weather, time of day, travel alterations, road environment, and whether the work zone is active (See Table~\ref{tbl:types}). The presence of roadwork objects in a scene does not necessarily indicate an active work zone, e.g., \textit{a cone in a parking lot}. Work zones are labeled as active work zone, not active work zone, or unsure. An active work zone includes roadwork as well as any activity that could potentially impact vehicles or pedestrians mobility. To qualify, objects must be located on a road or sidewalk where a vehicle or pedestrian could travel. Approximately 80\% images were labeled as active work zones.

\noindent \textbf{Scene Descriptions.} The associated descriptions detail key work zone elements, their locations, and relationships within the scene. They specify the approximate locations of work zones and objects on the road or sidewalk while also conveying the relative positioning of objects in relation to the work zone and other scene elements. To ensure consistency, all descriptions were written by a single annotator using a standardized vocabulary.

\noindent \textbf{Object Annotations.} We identified 15 categories of objects commonly found in work zones. These include objects that define temporary traffic control pathways, such as cones and tubular markers, fences, barriers, and drums. Additionally, we annotated objects that help navigation, including temporary traffic control (TTC) signs, TTC message boards and arrow boards. We also annotated Workers, Work Vehicles, Police officers and Police Vehicles, since they influence and direct traffic in work zones. See Table~\ref{tbl:types} for the full list of annotated work zone objects.

\begin{table}[t]
\centering
\resizebox{1.05\linewidth}{!}{
\begin{tabular}{@{}ll|llll@{}}
\toprule
\multicolumn{2}{c|}{\textbf{Object Categories}} & \textbf{Weather}       & \textbf{Alteration} & \textbf{Time}     & \textbf{Env.} \\ \midrule
Cone            & Tubular Marker       & Partly Cloudy & Fully Blocked     & Dark     & Urban    \\
Fence           & Vertical Panel       & Sunny         & Lane Shift        & Light    & Suburban \\
Worker          & Work Equipment       & Unknown       & Partially Blckd. & Twilight & Highway  \\
Work Vehicle    & Arrow Board          & Wet           & Other             & Unknown   & Rural    \\
TTC Sign        & TTC Msg. Board       & Cloudy        & None              & Other    & Unknown  \\
Drum            & Police Vehicle       & Fog or Mist   &                   &          & Other    \\
Barricade       & Police Officer       & Ice           &                   &          &          \\
Barrier         & Othr Rdwork Objs & Other         &                   &          &          \\
\bottomrule
\end{tabular}
}
\caption{\textbf{Work Zone Object Categories and Scene Level Tags.} The left side lists manually annotated object categories, while the right side present scene-level tags that describe various work zone properties.}
\label{tbl:types}
\end{table}

The object annotation workflow combined automatic and manual labeling, followed by manual verification. To reduce annotation effort, we used Detic~\cite{zhou2022detecting} with a custom vocabulary of ``cone, drum, vehicle, traffic sign'' to bootstrap annotations on our captured images. However, category predictions from Detic~\cite{zhou2022detecting} were discarded as due to frequent classification errors. Polygons were simplified using the Vishwalingam-Wyatt algorithm~\cite{visvalingam2017line} to facilitate editing. All object categories were manually assigned, and any additional objects in these images, as well as objects in all other images, were manually segmented and categorized. Finally, all annotations were manually verified by one person.



\noindent \textbf{Fine-Grained Object Annotations.} Objects that are partially blocked by other objects or truncated were labeled as ``occluded''. A few object categories, including arrow boards, TTC signs, and TTC message boards, have additional annotations. For example, arrow board states (``OFF'', ``LEFT'', ``RIGHT'', ``NONE'') is annotated (See Figure~\ref{fig:img_stats} for the distribution of arrow board states). 

TTC sign and TTC message boards generally contain both ``text'' and ``graphics''. We also annotated graphic descriptions (e.g. ``LEFT ARROW'') and associated text for each sign (e.g. ``DETOUR AHEAD''). Additionally, text or graphics were marked as ``occluded'' if the object is partially occluded or truncated by the image boundary. Sign text and graphic descriptions were parsed to identify common types of TTC signs (See Figure~\ref{fig:img_stats}). The distribution of TTC sign graphics and text follows a long-tailed pattern (See Figure~\ref{fig:img_stats}), with 62 and 360 different types annotated, respectively. 

\noindent \textbf{Semantic Segmentation.} We manually segmented roads, sidewalks, and a sparse sampling of bicycle lanes to provide contextual localization for work zone objects.

\subsection{Metric 3D Reconstruction and Pathway Generation from Smartphone Videos}
\label{appendix:recons-pathway}

\noindent \textbf{Leveraging Smartphone-As-Dashcam Videos.} Our work utilizes the MMI Open Data Set~\cite{roadbotics-open}, which contains extensive video footage captured from a Samsung Galaxy S9 smartphone, for which camera intrinsics are known. From this dataset, we extract 30-second video snippets corresponding to our annotated work zones. These snippets are then downsampled to 5 FPS to yield the final set of smartphone images for our 3D reconstruction pipeline.

\noindent \textbf{Leverging 3D Reconstruction As Anchor.} Our primary goal is to produce an accurate, metric-scale, and spatially-aligned 3D reconstruction from the collection of smartphone videos, which have weakly-aligned GPS metadata. With recent advances in Visual Place Recognition~\cite{berton2025megaloc}, it's likely that the weakly-aligned GPS metadata might also be superfluous in the future.

The core of our approach is to anchor our reconstruction to a set of images with high-quality pose information~\cite{vuong2024toward, vuong2025aerialmegadepth}. To achieve this, we use the initial, coarse GPS from each video to query and retrieve nearby Google Street View panoramas. We then generate multiple perspective views from each panorama, following the systematic sampling strategy described in~\cite{vuong2024toward}. These views, along with the panoramas' accurate GPS and pose data from large-scale SfM pipelines~\cite{klingner2013street} that Google Street View is based on, serve as the high-quality georeferencing anchor for our 3D reconstructions.

\noindent \textbf{Feature Matching and SfM.} To reconstruct from this heterogeneous set of smartphone and Street View images, we must establish robust feature matches. As a brute-force all-pairs matching approach is computationally infeasible, we adopt a retrieval-based strategy. We first compute a global descriptor for every image using EigenPlaces~\cite{berton2023eigenplaces}. We then use these features to find the top 20 nearest neighbors for each image using the Faiss library~\cite{douze2024faiss}, efficiently identifying pairs with likely visual overlap. For these candidate pairs, we perform local feature matching by extracting keypoints and descriptors with SuperPoint~\cite{detone2018superpoint} and matching them with LightGlue~\cite{lindenberger2023lightglue}. With this graph of matched images, we perform Structure-from-Motion (SfM) using the global solver GLOMAP~\cite{pan2024global} to recover camera poses and a sparse 3D point cloud. COLMAP~\cite{schoenberger2016sfm} is also applicable but GLOMAP~\cite{pan2024global} is an order-of-magnitude faster.

\begin{figure*}[t!]
    \centering
    \includegraphics[width=1\textwidth]{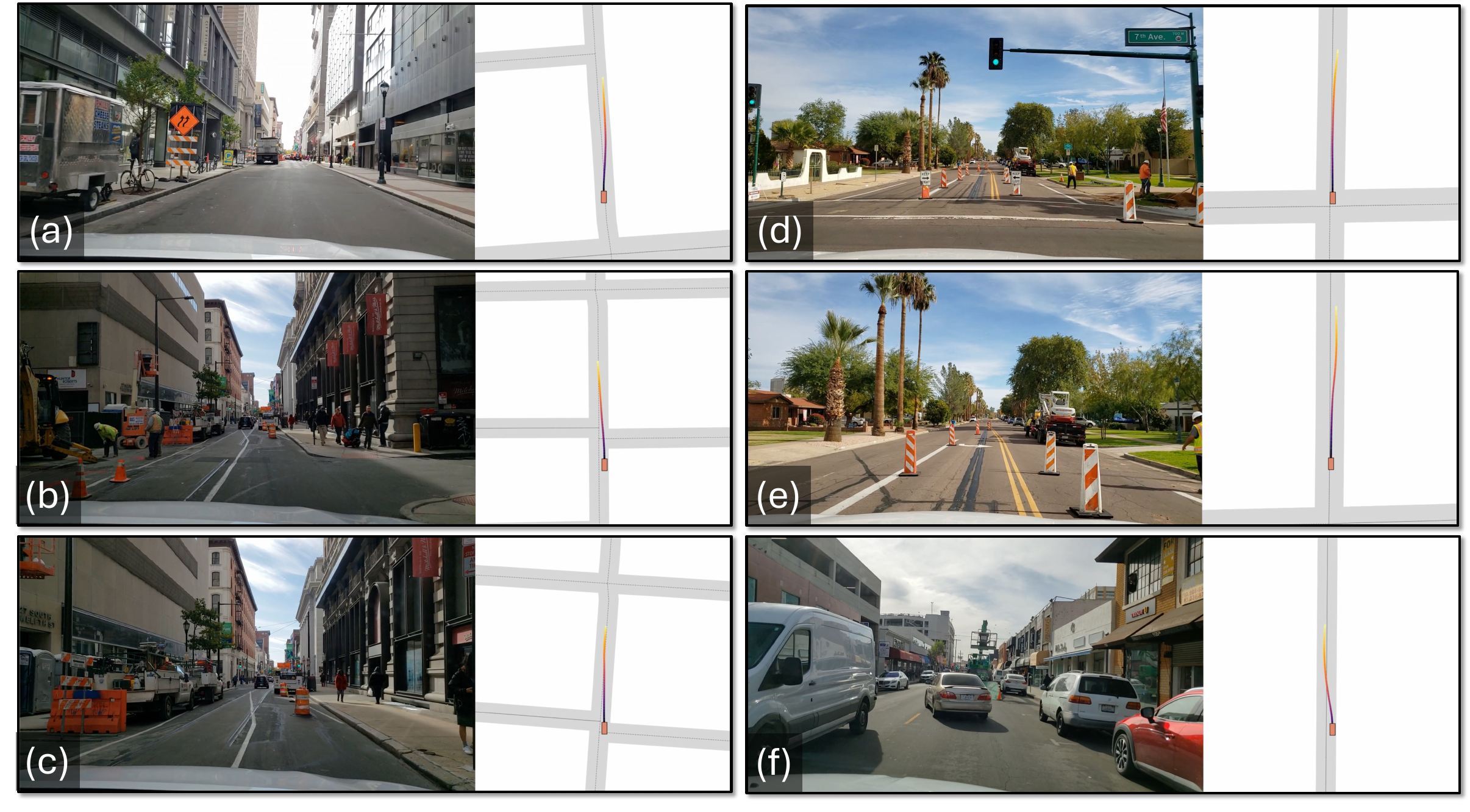}
    \caption{\textbf{Metric Geo-referenced Trajectories from our 3D Reconstruction Pipeline.} We show examples of trajectories obtained from our reconstruction pipeline overlayed on birds-eye-view maps retrieved from OpenStreetMaps. We show some interesting situations for planning which require all aspects of scene perception, the trajectories are shown for a 4 second future horizon. \textbf{(a-c)} Depict a construction zone with two lane changes, first lane change to the right is marked by a TTC sign, while the second lane change is marked with drums. \textbf{(d-e)} Depict a construction zone marked with TTC signs and Vertical Panels. While there exists ``free'' space to navigate to the right most lane (where the workers are), the objects helpfully mark the actual drivable regions. Identifying drivable regions is still challenging for self-driving cars~\cite{news-2024missionlocal, news-nyt2023concrete}. \textbf{(f)} Cones behind a work zone vehicle mark it as a static object blocking the lane. This cue guides the cars to change the lane towards oncoming traffic to pass this work zone vehicle.
    }
    \label{fig:e2e-planning-viz}
\end{figure*}

\noindent \textbf{Georeferencing and Trajectory Generation.} 
A key step is georeferencing the resulting 3D reconstruction such that we align the reconstruction to a real-world coordinate system via a 7-DoF similarity transformation. Following the procedure described in~\cite{vuong2024toward, vuong2024walt3d}, this is computed by minimizing the discrepancy between the recovered poses of the Street View images and their ground-truth GPS data, which we project into an Earth-Centered, Earth-Fixed (ECEF) global coordinate frame. We explicitly discard the noisy GPS from the smartphone videos during this alignment, relying solely on the high-quality Street View data for metric accuracy~\cite{klingner2013street}. The result is a single, georeferenced sparse reconstruction where the poses for all smartphone images are accurately localized, forming precise 3D trajectories. We then fit a ground plane to the reconstruction by using a Mask2Former~\cite{cheng2022masked} semantic segmentation model to identify 3D points corresponding to the road surface. By projecting the 3D camera poses onto this fitted plane, we define the vehicle's 3D path, which is then projected back into the source images to create 2D drivable trajectories. Visualization of our trajectories can be viewed in Figure~\ref{fig:e2e-planning-viz}.

\noindent \textbf{Trajectories to Waypoints.} To standardize the trajectories for our prediction task, we convert them into a fixed number of waypoints. For each sequence, we identify the longest continuous segment of the 2D trajectory that remains on the road. We then fit a spline to this segment and sample 20 equidistant waypoints. For the pathway prediction problem discussed in the main manuscript, the first five waypoints serve as the observed path (input), the final waypoint represents the goal, and the intermediate 14 points constitute the future pathway to be predicted.

\subsection{Other Details}
\label{sup:dataset-details}

\noindent
\textbf{Number of Workzones.}  Counting work zones is challenging, as they could extend for miles. Should such long stretches be considered a single work zone? Additionally, workzones resemble the Ship of Theseus -- they evolve over time while remaining at the same location for months or even years. These spatio-temporal factors make it difficult to define and count work zones accurately.

In our analysis, we counted workzones based on locations alone, clustering images within a 20m radius  as a single work zone, regardless of when they were captured. As a result, \emphdatasetname dataset contains instances of work zones at the same location observed across months or years. Besides, we used the DBSCAN algorithm to cluster workzones images based on the the noisy GPS locations. Consequently, we obtained 5024 clusters at a 20m threshold and 4759 such clusters at a 30m threshold. Based on these results, we estimate that our dataset contains approximately 5,000 work zones.

\noindent
\textbf{More Visualizations and Details.} A full description of all annotated categories is provided in Table~\ref{tbl:supp-object-descrip}, while the distribution of each class across all cities is shown in Table~\ref{tbl:supp-distribution}.

\clearpage

\section{Additional Analysis and Results}
\label{sup:addl-results}

\subsection{Recognizing Work Zones}
\label{sup:addl-recog-results}

\begin{table}[h]
\centering
\resizebox{\linewidth}{!}{
\begin{tabular}{lcccccc}
\hline
 \textbf{Method} &
  \multicolumn{1}{l}{$AP$} &
  \multicolumn{1}{l}{$AP50$} &
  \multicolumn{1}{l}{$AP75$} &
  \multicolumn{1}{l}{$AP_s$} &
  \multicolumn{1}{l}{$AP_m$} &
  \multicolumn{1}{l}{$AP_{l}$} \\ \hline
\multicolumn{7}{c}{\textbf{Open Vocabulary Detectors}}                                  \\ \hline
Grounding DINO (O365)~\cite{liu2024grounding} & 6.6  & 9.5  & 7.0  & 4.0  & 7.1  & 10.5 \\ \hline
\multicolumn{7}{c}{\textbf{Supervised with \emphdatasetname Dataset}}                   \\ \hline
Faster R-CNN~\cite{ren2016faster}             & 25.0 & 42.4 & 25.8 & 12.7 & 30.6 & 36.0 \\
DiffusionDet~\cite{chen2023diffusiondet}      & 31.1 & 50.1 & 32.2 & 18.3 & 30.8 & 42.0 \\
Grounding DINO~\cite{liu2024grounding}        & 37.9 & 54.2 & 39.8 & 21.7 & 39.0 & 51.9 \\
\rowcolor[HTML]{ECF4FF} 
DINO~\cite{zhangdino} &
  \textbf{39.9} &
  \textbf{57.2} &
  \textbf{42.2} &
  \textbf{24.0} &
  \textbf{38.6} &
  \textbf{52.1} \\ \hline
\end{tabular}%
}
\caption{\textbf{Detecting Work Zone Objects.} We train detection models~\cite{chen2023diffusiondet, zhangdino} on the \emphdatasetname dataset using a coarse vocabulary. The open-vocabulary detector~\cite{liu2024grounding} struggle to recognize work zone objects, but incorporating our data significantly improves its performance. Overall, our supervised models achieve substantially better results (\textbf{\txtgreen{+33.3 $AP$}}).}
\label{tbl:sup-det-coarse-grained}
\end{table}

\noindent \textbf{Detecting Work Zone Objects.} As mentioned in Section 3 of the main manuscript, open-vocabulary detectors such as Grounding DINO~\cite{liu2024grounding} follow similar trends to Detic~\cite{zhou2022detecting} and OpenSeeD~\cite{xu2023open}. As shown in Table~\ref{tbl:sup-det-coarse-grained}, supervised models like DiffusionDet~\cite{chen2023diffusiondet} and DINO~\cite{zhangdino} significantly outperform Grounding DINO (\textbf{\txtgreen{+33.3 $AP$}}). Fortunately, fine-tuning Grounding DINO~\cite{liu2024grounding} on our data almost matches the performance to DINO~\cite{zhangdino}. However, DINO~\cite{zhangdino} is still better than Grounding DINO~\cite{liu2024grounding} by \textbf{\txtgreen{+2.0 $AP$}}, likely reflecting the trade-off between a specialized detector and an open-vocabulary detector that generalizes well across a larger number of categories.

\begin{table}[h]
\centering
\resizebox{\linewidth}{!}{%
\begin{tabular}{lcccccc}
\hline
\textbf{Method}                          & $AP$   & $AP50$  & $AP75$  & $AP_s$         & $AP_m$  & $AP_l$  \\ \hline
\multicolumn{1}{c}{}                     & \multicolumn{6}{c}{\textbf{Mapillary (Discovered In-The-Wild)}} \\ \hline
Grounding DINO~\cite{liu2024grounding} \textbf{(pre-trained)}     & 5.1    & 7.9     & 5.3     & 2.2            & 5.3     & 9.1     \\ \hline
DiffusionDet~\cite{chen2023diffusiondet} & 13.1   & 24.3    & 12.5    & 5.2            & 12.6    & 23.4    \\
DINO~\cite{zhangdino}                    & 19.7   & 32.1    & 20.2    & \textbf{10.0}  & 18.0    & 31.8    \\
\rowcolor[HTML]{ECF4FF} 
Grounding DINO~\cite{liu2024grounding} & \textbf{22.8} & \textbf{35.2} & \textbf{23.2} & 7.9           & \textbf{19.6} & \textbf{37.6} \\ \hline
\multicolumn{1}{c}{}                     & \multicolumn{6}{c}{\textbf{BDD100K (Discovered In-The-Wild)}}   \\ \hline
Grounding DINO~\cite{liu2024grounding} \textbf{(pre-trained)} & 8.8    & 13.0    & 9.5     & 6.6            & 10.9    & 11.7    \\ \hline
DiffusionDet~\cite{chen2023diffusiondet} & 18.5   & 33.2    & 17.9    & 12.1           & 20.7    & 27.9    \\
DINO~\cite{zhangdino}                    & 27.3   & 43.0    & 28.2    & 17.0           & 28.8    & 36.2    \\
\rowcolor[HTML]{ECF4FF} 
Grounding DINO~\cite{liu2024grounding} & \textbf{28.5} & \textbf{43.2} & \textbf{29.4} & \textbf{20.1} & \textbf{31.8} & \textbf{38.6} \\ \hline
\end{tabular}%
}
\caption{\textbf{Zero-Shot Detection On Discovered Workzones From BDD100K And Mapillary.} For discovered-in-the-wild work zone images,  fine-tuning the open-vocabulary detector Grounding DINO on our \emphdatasetname dataset improves performance by \textbf{\txtgreen{+17.7 $AP$}} on Mapillary and \textbf{\txtgreen{ +19.7 $AP$}} on BDD100K. Additionally, the supervised detectors DiffusionDet~\cite{chen2023diffusiondet} and DINO~\cite{zhangdino} achieve promising performance. }
\label{tbl:zero-shot-discovered-det}
\end{table}


%

\noindent \textbf{Zero-Shot Detection On Discovered Workzones.} In Section~\pointToMain{~\ref{subsec:recognize}}{3} of the main manuscript, we discovered 969 images in BDD100K~\cite{yu2020bdd100k} and Mapillary~\cite{neuhold2017mapillary} datasets. While detectors trained on the \emphdatasetname dataset facilitated the discovery of work zones around the world, their performance on these in-the-wild images has not been evaluated. To assess generalization, we manually annotated the 969 in-the-wild work zone images discovered in BDD and Mapillary (See Table 4 of the main manuscript for workzone discovery experiments). As shown in Table \ref{tbl:zero-shot-discovered-det}, the open-vocabulary detector Grounding DINO achieves significantly better zero-shot performance after being fine-tuned on our dataset, while supervised detectors also delivers promising zero-shot performance. Interestingly, compared to in-distribution performance (Table~\ref{tbl:sup-det-coarse-grained}) where DINO~\cite{zhangdino} is better than Grounding DINO~\cite{liu2024grounding} by \textbf{\txtgreen{+2.0 $AP$}}, in this case Grounding DINO~\cite{liu2024grounding} shows improved generalization (\textbf{\txtgreen{+3.1 $AP$}}) over DINO~\cite{zhangdino}. 

\begin{table}[h]
\centering
\resizebox{\linewidth}{!}{%
\begin{tabular}{lcccccc}
\hline
\textbf{Method}              & $AP$       & $AP50$     & $AP75$     & $AP_s$      & $AP_m$      & $AP_l$      \\ \hline
                    & \multicolumn{6}{c}{\textbf{Mapillary (Discovered In-The-Wild)}} \\ \hline
Detic~\cite{zhou2022detecting} \textbf{(pre-trained)} & 2.9      & 4.5      & 2.9      & 0.6      & 3.3      & 5.4      \\
Mask R-CNN~\cite{he2017mask}          & 14.4     & 25.4     & 14.2     & 2.8      & 14.1     & 26.9     \\
\rowcolor[HTML]{ECF4FF} 
Mask DINO~\cite{li2023maskdino}           & 21.6     & 35.5     & 22.5     & 6.9      & 19       & 37.2     \\ \hline
                    & \multicolumn{6}{c}{\textbf{BDD100K (Discovered In-The-Wild)}}            \\ \hline
Detic~\cite{zhou2022detecting} \textbf{(pre-trained)} & 3.7      & 5.8      & 4        & 3        & 5.1      & 4        \\
Mask R-CNN~\cite{he2017mask}          & 19.8     & 33.8     & 21       & 12.8     & 23.3     & 28.1     \\
\rowcolor[HTML]{ECF4FF} 
Mask DINO~\cite{li2023maskdino}           & 29.1     & 46.6     & 31.3     & 18.1     & 31.4     & 45.5     \\ \hline
\end{tabular}%
}
\caption{\textbf{Zero-Shot Instance Segmentation on Discovered Workzones from BDD100K and Mapillary.} As we noted in Section\pointToMain{~\ref{subsec:recognize}}{3}, Detic~\cite{zhou2022detecting} performed miserably for discovering work zones. We also observe that Detic's zero-shot performance on work zone images from Mapillary~\cite{neuhold2017mapillary} and BDD100K~\cite{yu2020bdd100k} follows the trends from the main manuscript. Similarly, Mask DINO~\cite{li2023maskdino} performs significantly better on both out-of-distribution datasets.}
\label{tbl:zero-shot-discovered-ins}
\end{table}

\noindent
\textbf{Zero-Shot Segmentation on Discovered Workzones.} We evaluate open-vocabulary detectors and \emphdatasetname supervised models on discovered images (See Section\pointToMain{~\ref{subsec:recognize}}{3}) -- which are out-of-distribution for all the models. Pre-trained Detic performs poorly on both Mapillary (\textbf{\txtred{2.9 AP}}) and BDD100K (\textbf{\txtred{3.7 AP}}), reinforcing our observation that work zone objects are severely underrepresented in foundation model training data. In contrast, models trained on the ROADWork dataset show substantial improvements. Mask DINO~\cite{li2023maskdino} achieves \textbf{\txtgreen{21.6 AP}} on Mapillary and \textbf{\txtgreen{29.1 AP}} on BDD100K, representing gains of \textbf{\txtgreen{+18.7 AP}} and \textbf{\txtgreen{+25.4 AP}} respectively over pre-trained Detic. Even the simpler Mask R-CNN architecture demonstrates significant improvements when trained on our dataset.

\begin{table}[h]
\centering
\resizebox{\linewidth}{!}{%
\begin{tabular}{lcc|cc|cc|cc}
\hline
\textbf{Method}     & $AP$ & $AP75$ & $AP$ & $AP75$ & $AP$ & $AP75$ & $AP$ & $AP75$ \\ \hline
 &
  \multicolumn{2}{c|}{\textbf{Vehicle - Pittsburgh}} &
  \multicolumn{2}{c|}{\textbf{Vehicle - Rural}} &
  \multicolumn{2}{c|}{\textbf{Pittsburgh Bus - Front Cam.}} &
  \multicolumn{2}{c}{\textbf{Pittsburgh Bus - Side Cam.}} \\ \hline
Detic \textbf{(pre-trained)} & 5.1  & 5.9    & 2.4  & 2.5    & 3.6  & 3.4    & 3.7  & 3.8    \\ \hline
Mask R-CNN          & 28.1 & 30.9   & 20.5 & 20.9   & 20   & 20.9   & 20.1 & 21.9   \\
\rowcolor[HTML]{ECF4FF} 
Mask DINO &
  \textbf{38} &
  \textbf{39.6} &
  \textbf{30.1} &
  \textbf{30.0} &
  \textbf{29.4} &
  \textbf{30.4} &
  \textbf{32.6} &
  \textbf{35.6} \\ \hline
\end{tabular}%
}
\caption{\textbf{Zero-Shot Instance Segmentation Results On Other Discovered Work Zone Images.} We evaluate instance segmentation models on additional discovered work zone subsets (See Figure 13) from various sources. Consistent with our prior findings, pre-trained Detic struggles on these specialized subsets, while Mask DINO trained on ROADWork significantly outperforms both pre-trained models and simpler architectures like Mask R-CNN. The performance gap is particularly noticeable in more challenging conditions like rural areas and bus-mounted camera views.}\
\vspace{-0.3in}
\label{tbl:zero-shot-others}
\end{table}

\noindent
\textbf{Zero-Shot Instance Segmentation Results On Other Discovered Work Zone Images.} We further evaluate our models on additional discovered work zone subsets (Vehicle-Pittsburgh, Vehicle-Rural, and Bus-Pittsburgh) to assess generalization under varying conditions. As shown in Table~\ref{tbl:zero-shot-others} pre-trained Detic~\cite{zhou2022detecting} performs miserably across all subsets, with AP values ranging from \textbf{\txtred{2.4}} to \textbf{\txtred{5.1}}. This performance is particularly poor in the Vehicle-Rural subset, where the AP is merely \textbf{\txtred{2.4}}, highlighting the difficulty of segmenting work zones in rural environments. In contrast, models trained on \emphdatasetname show substantially better performance, with Mask DINO achieving the best results across all subsets (\textbf{\txtgreen{+32.9 $AP$}} on Vehicle-Pittsburgh, \textbf{\txtgreen{+27.7 $AP$}} on Vehicle-Rural, \textbf{\txtgreen{+25.8 $AP$}} on Bus-Pittsburgh Front Camera, and \textbf{\txtgreen{+28.9 $AP$}} on Bus-Pittsburgh Side Camera compared to pre-trained Detic). These results further validate our observations from the main manuscript, confirming that foundation models struggle with work zone recognition in diverse conditions, while our \emphdatasetname-trained models generalize effectively across various scenarios and viewpoints.

\begin{table}[h]
\centering
\resizebox{\linewidth}{!}{%
\begin{tabular}{llcccccc}
\hline
\textbf{Training Data}             & \textbf{Test Data}        & $AP$ & $AP50$ & $AP75$ & $AP_s$ & $AP_m$ & $AP_l$ \\ \hline
Obj365                    & \emphdatasetname & 6.6  & 9.5    & 7.0    & 4.0    & 7.1    & 10.5   \\
Obj365 + \emphdatasetname & \emphdatasetname & 37.9 & 54.2   & 39.8   & 21.7   & 39.0   & 51.9   \\ \hline
Obj365                    & Cityscapes       & 34.2 & 50.2   & 35.9   & 13.6   & 36.0   & 56.2   \\
\rowcolor[HTML]{ECF4FF} 
Obj365 + \emphdatasetname & Cityscapes       & 34.4 & 52.2   & 34.2   & 11.7   & 33.4   & 54.0   \\ \hline
Obj365                    & BDD100K          & 23.6 & 40.6   & 23.2   & 9.2    & 27.8   & 49.5   \\
\rowcolor[HTML]{ECF4FF} 
Obj365 + \emphdatasetname & BDD100K          & 23.7 & 40.9   & 22.8   & 8.4    & 27.2   & 50.0   \\ \hline
\end{tabular}%
}
\caption{\textbf{Does fine-tuning a open-vocabulary foundation model on \emphdatasetname cause overfitting?} Large foundation Models are prone to overfitting when trained on small datasets. To assess whether our dataset is large enough to mitigate overfitting, we finetune Grounding DINO~\cite{liu2024grounding} and evaluate it on common driving datasets including Cityscapes~\cite{cordts2016cityscapes} and BDD100K~\cite{yu2020bdd100k}, using their respective categories. We observe that  fine-tuning significantly improves performance on \emphdatasetname dataset (\textbf{\txtgreen{+31.3 $AP$}}), while performance on Cityscapes (\textbf{\txtgreen{+0.2 $AP$}}) and BDD100K (\textbf{\txtgreen{+0.1 $AP$}}) does not degrade.}
\label{tbl:sup-grounding-dino-forgetting}
\end{table}

\noindent \textbf{Does fine-tuning a open-vocabulary foundation model on \emphdatasetname cause overfitting?} Large-scale foundation open-vocabulary model trained on millions of images may forget previously learned distributions when fine-tuned on additional data~\cite{ilharco2022patching}. This effect is more pronounced when the dataset used for fine-tuning is small, leading to overfitting on the target data. To assess whether if \emphdatasetname dataset is large enough to mitigate overfitting, we evaluate Grounding DINO~\cite{liu2024grounding} on common driving datasets Cityscapes~\cite{cordts2016cityscapes} and BDD100K~\cite{yu2020bdd100k} with their label set as the vocabulary. We then finetune the model on \emphdatasetname, and re-evaluate the detector on the same datasets with their vocabulary. As shown in Table~\ref{tbl:sup-grounding-dino-forgetting}, fine-tuning significantly improves performance on \emphdatasetname (\textbf{\txtgreen{+31.3 $AP$}}), while performance marginally improves on Cityscapes~\cite{cordts2016cityscapes} (\textbf{\txtgreen{+0.2 $AP$}}) or BDD100K~\cite{yu2020bdd100k} (\textbf{\txtgreen{$AP$}}). We hypothesize that this is due to the small domain gap between \emphdatasetname dataset and common driving datasets. Hence, image features might remain consistent even when fine-tuned on our data. We leave further exploration of this phenomenon for future work.

\begin{table}[h]
\centering
\resizebox{\linewidth}{!}{
\begin{tabular}{lcccccc}
\hline
\textbf{Supervision} &
  \multicolumn{1}{l}{$AP$} &
  \multicolumn{1}{l}{$AP50$} &
  \multicolumn{1}{l}{$AP75$} &
  \multicolumn{1}{l}{$AP_s$} &
  \multicolumn{1}{l}{$AP_m$} &
  \multicolumn{1}{l}{$AP_{l}$} \\ \hline
\multicolumn{7}{c}{\textbf{Psuedo-Segmentations from SAM~\cite{kirillov2023segany}}}                                  \\ \hline
Bbox                  & 22.6          & 44.7          & 20.9          & 14.3          & 29.6          & 30.9          \\
Bbox + 5 pts          & 23.3          & 44.9          & 22.1          & 17.7          & 29.6          & 30.4          \\
Bbox + 10 pts         & 23.5          & 45.6          & 22.4          & 15.3          & 30.0          & 30.2          \\ \hline
\rowcolor[HTML]{ECF4FF} 
\textbf{Ground Truth} & \textbf{27.6} & \textbf{47.2} & \textbf{29.1} & \textbf{18.7} & \textbf{33.5} & \textbf{35.9} \\ \hline
\end{tabular}%
}
\caption{\textbf{Are Manual Segmentations Still Needed? Results with Boundary IOU.} We train instance segmentation models~\cite{he2017mask} with varying levels of supervision from the \emphdatasetname dataset using SAM~\cite{kirillov2023segany}. Unlike the results in the main paper, we evaluate performance using Boundary IOU~\cite{cheng2021boundary}, a metric more sensitive to boundary errors than standard IOU. We observe a larger improvement using Boundary IOU at higher thresholds (\textbf{\txtgreen{+6.7 $AP75_{IOU(B)}$}}), which indicates boundary quality improvements with manual annotations compared to psuedo ground truth annotations.}
\label{tbl:sam-results-boundary}
\end{table}

\noindent \textbf{Are Manual Segmentations Still Needed?} We posited in Section 3 of the main manuscript that some of the object categories in \emphdatasetname exhibit irregular shapes. We present additional results in Table~\ref{tbl:sam-results-boundary}, computing AP using the Boundary IOU~\cite{cheng2021boundary} metric ($AP_{IOU(B)}$). This metric penalizes boundary errors more strictly, making it more suitable for evaluating segmentation quality, particularly for irregularly shaped work zone objects such as arrow boards and work vehicles (e.g., ``cranes''). Compared to the results in Section 3 of the main manuscript, we find that the performance gap at tighter thresholds is even more pronounced (\textbf{\txtgreen{+6.7 $AP75_{IOU(B)}$}}) compared to \textbf{\txtgreen{+5.3 $AP75$}} from Table 3 of the main manuscript. This further underscores that manual ground-truth masks yield higher quality boundaries than those predicted by SAM~\cite{kirillov2023segany}, reinforcing the need for manual segmentations of rare work zone objects.

\subsubsection{Other Interesting Recognition Scenarios}
\label{sup:other-interesting-scenarios}

\emphdatasetname dataset enables the study of various scene understanding challenges beyond those considered of the main manuscript.

\begin{figure}[h]
    \centering
    \includegraphics[width=\linewidth]{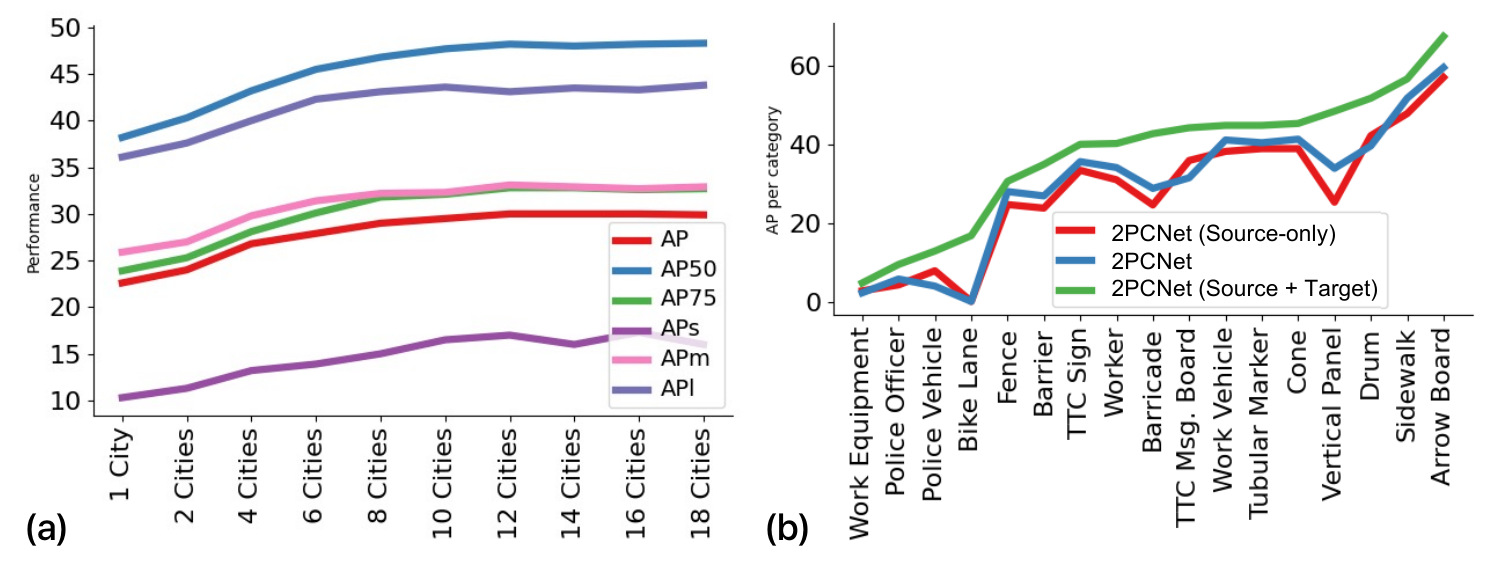}
    \caption{\textbf{Adapting to New Geographies.} (a) Starting from Pittsburgh, we progressively add data from new cities to train a detector, leading to significant accuracy gains (\textbf{\txtgreen{+7.4 $AP$}} with all cities). (b) Geographic adaptation remains challenging. Training on source data only serves as our \textbf{\txtdeepred{baseline}}, while training on both source and target data represents the \textbf{\txtgreen{upper bound}}. \textbf{\txtblue{Adaptation methods}} such as 2PCNet~\cite{kennerley20232pcnet} provide limited improvement over the baseline. For example, the upper-bound gap of adaptation method for ``barricade'' (\textbf{\red{-13.8 $AP_{50}$}}) and for ``vertical panel'' (\textbf{\red{-20.7 $AP_{50}$}}) is very large.}
    \label{fig:geo-adapt-fig}
\end{figure}

\noindent \textbf{Adapting to New Geographies.} While we discovered work zones in new geographies (Figure 5 of the main manuscript), does  our recognition model maintain the same performance? Domain adaptation methods have explored geographic adaptation, but mainly across countries~\cite{wang2020train, varma2019idd, kalluri2023geonet, zheng2024instance, hoyer2022daformer} and mostly for common objects like cars~\cite{wang2020train}. Obtaining supervised data for new geographies , such as new cities in our case, is expensive to scale. We make two observations: (a) A geographic domain gap exists in our data, and (b) state-of-the-art adaptation methods do not address this gap.

To demonstrate these observations, we conduct a simple experiment.  We train work zone detector using data from Pittsburgh and test it on all cities. After that, we add data from the city with most samples and retrain the model. Accuracy improves by \textbf{\txtgreen{+1.4 $AP$}}. We continue adding data from other cities, leading to a final improvement of \textbf{\txtgreen{+7.4 $AP$}} (Figure~\ref{fig:geo-adapt-fig} (a)).


Next, we consider the unsupervised domain adaptation problem. We treat data from Pittsburgh as source domain, where both images and labels are available during training. Data from other cities (excluding Pittsburgh) forms the target domain, where only images are available during training while performing adaptation. We evaluate the model on the target domain.

Following state-of-the-art adaptation methods~\cite{kennerley20232pcnet, zheng2024instance}, we use a Faster R-CNN Resnet50 backbone for training, assuming different levels of available target domain data. Training on source data only is our baseline (\textbf{\txtdeepred{red}} in Figure~\ref{fig:geo-adapt-fig} (b)), whereas upper bound is trained on both labeled source and labeled target data (depicted in \textbf{\txtgreen{green}}). State-of-the-art adaptation methods~\cite{kennerley20232pcnet, zheng2024instance} (depicted in \textbf{\txtblue{blue}}) do not significantly improve adaptation over baseline. Compared to the upper bound in Figure~\ref{fig:geo-adapt-fig} (b), performance gaps remain for heavily represented objects like cones (\textbf{\red{-5.4 $AP_{50}$}}) and drums (\textbf{\red{-12.6 $AP_{50}$}}), and also for rare objects like barricades (\textbf{\red{-13.8 $AP_{50}$}}) and vertical panels (\textbf{\red{-20.7 $AP_{50}$}}). Our \emphdatasetname dataset highlights the geographic domain gap problem, underscoring the need for new algorithms to bridge this gap.


\begin{table}[h]
\centering
\resizebox{1.0\linewidth}{!}{
\small
\begin{tabular}{lcccccc}
\hline
\textbf{Method}                      & $AP$          & $AP50$        & $AP75$ & $AP_s$ & $AP_m$ & $AP_l$ \\ \hline
CS Psuedo Labels                     & 35.3          & 62.2          & 35.4   & \textbf{17.2}   & 36.9   & 50.6   \\
\rowcolor[HTML]{ECF4FF} 
UniDetector~\cite{wang2023detecting} & \textbf{37.3} & \textbf{63.6} & N/A    & 14.6   & \textbf{37.9}   & \textbf{56.1}   \\ \hline
Cityscapes (Pretrained)              & 40.3          & 65.3          & 42.1   & 17.2   & 40.9   & 61.4  \\ \hline
\end{tabular}%
}
\captionof{table}{\textbf{Label Unification.} 
We train a bounding box detector on unified Cityscapes~\cite{cordts2016cityscapes} and \emphdatasetname label space by (a) pseudo-labeling our dataset using a pretrained Cityscapes~\cite{cordts2016cityscapes} (CS) model (b) via UniDetector~\cite{wang2023detecting}. Testing on Cityscapes~\cite{cordts2016cityscapes} and compared to a pretrained model solely trained on Cityscapes~\cite{cordts2016cityscapes} labels, we observe significant degradation when trained on unified label space while UniDetector~\cite{wang2023detecting} improves the performance over naive pseudo-labeling.}
\label{table:unidet}
\end{table}

\noindent \textbf{Label Unification.} Suppose we aim to train a unified detector that detects both common objects like cars from a common driving dataset and rare objects from the \emphdatasetname dataset simultaneously. This requires label unification. While practical, unifying the label space is a challenging task~\cite{zhao2020object, wang2023detecting} -- training a model on a unified set of categories reduces performance compared to specialized models individually trained on each dataset. One reason is due to the presence of \textit{unlabeled} instances of a particular category in the unified dataset, another could be due concept overlaps between two labels in the unified dataset. To assess this observation, we consider the Cityscapes~\cite{cordts2016cityscapes} bounding box dataset in addition to our \emphdatasetname dataset, using UniDetector~\cite{wang2023detecting} with a Faster R-CNN model. We train our model on the unified label space of Cityscapes (common objects) and \emphdatasetname dataset (long-tailed objects) -- (a) by employing a pre-trained detector (see Section 3) to pseudo-label the \emphdatasetname dataset. (b) employing the method proposed by UniDetector~\cite{wang2023detecting}. When testing our unified models on the Cityscapes validation set, we observe a considerable drop in performance (\txtdeepred{-5.0 AP}) when naive pseudo-labeling is used. However, UniDetector~\cite{wang2023detecting} closes the gap with the Cityscapes pre-trained model by improving the performance by \textbf{\txtgreen{+2.0 $AP$}} over naive pseudo-labeling.


\subsection{Analyzing Work Zones}
\label{sup:addl-analyzing-work-zones}

\begin{table}[h]
\centering
\resizebox{\linewidth}{!}{%
\begin{tabular}{lrcccc}
\hline
Pretrained                                                       & \textbf{Size} & \textbf{BLEU@4} & \textbf{METEOR} & \textbf{ROUGE} & \textbf{CIDEr} \\ \hline
LLaVA-1.5~\cite{liu2024improved}                         & 7B  & 0.4  & 11.0 & 9.4           & 0              \\
LLaVA-1.5~\cite{liu2024improved}                         & 13B & 0.3  & 9.8  & 8.0           & 0              \\
LLaVA-NEXT~\cite{liu2024improved,li2024llavanext-strong} & 13B & 0.2  & 9.4  & 6.9           & 0              \\
LLaVA-NEXT~\cite{liu2024improved,li2024llavanext-strong} & 34B & 0.3  & 9.3  & 6.9           & 0              \\ \hline
\multicolumn{1}{c}{\textbf{Fine-tuned}}                  &     &      &      &               &                \\ \hline
LLaVA-1.5~\cite{liu2024improved}                         & 7B  & 27.0 & 24.7 & 48.0          & 112.1          \\
LLaVA-1.5~\cite{liu2024improved}                         & 13B & 27.7 & 25.1 & \textbf{48.6} & 113.1          \\
LLaVA-NEXT~\cite{li2024llavanext-strong}                 & 13B & 28.2 & 25.3 & 48.3          & \textbf{116.4} \\
\rowcolor[HTML]{ECF4FF} LLaVA-NEXT~\cite{li2024llavanext-strong} & 34B           & \textbf{28.4}   & \textbf{26.1}   & 47.2           & 113.2          \\ \hline
\end{tabular}%
}
\caption{\textbf{Newer and Larger Vision-Language Models (VLMs).} Consistent with the poor performance trends in Section 5 of the main manuscript, larger pretrained VLMs (13B–34B parameters) also fail to describe work zones. Switching to a newer generation of VLMs~\cite{li2024llavanext-strong} does not improve performance, reinforcing the under-representation of work zones in existing large-scale training datasets. Fine-tuning helps, but even a 34B model provides only a marginal improvement (\textbf{\txtgreen{+1.4 METEOR}}).}
\label{tbl:sup-vlm-results}
\end{table}

\noindent \textbf{Newer and Larger Vision-Language Models (VLMs).} We performed our experiments in Section 5 of the main manuscript using LLaVA-1.5-7B. However, two key  questions arise: \textbf{(a)} Do larger VLMs also struggle with work zones descriptions? \textbf{(b)} Are newer VLMs such as LLaVA-NEXT~\cite{li2024llavanext-strong} better at describing work zones than older models? 

As shown in Table~\ref{tbl:sup-vlm-results}, pre-trained VLMs of all sizes perform poorly on work zones unless they are fine-tuned on \emphdatasetname dataset. Unfortunately, even the larger LLaVA-NEXT-34B model provides only a marginally performance gain (e.g. \textbf{\txtgreen{+1.4 METEOR}}) over LLaVA-1.5-7B~\cite{liu2024improved}. Moreover, newer VLMs like LLaVA-NEXT~\cite{li2024llavanext-strong} do not significantly outperform the previous generation of VLMs, likely because they still lack exposure to work zone images. Our \emphdatasetname dataset fills that gap, advancing high-level scene understanding in work zones.

\begin{table}[h]
\centering
\resizebox{\linewidth}{!}{%
\begin{tabular}{llcccc}
\hline
\textbf{Methods} & \textbf{Dataset}         & \textbf{BLEU@4} & \textbf{METEOR} & \textbf{ROUGE} & \textbf{CIDER} \\ \hline
LLaVA-1.5-7B     & LLaVA                    & 4.7             & \textbf{18.1}   & 18.9           & 0              \\
\rowcolor[HTML]{ECF4FF} 
LLaVA-1.5-7B     & LLaVA + \emphdatasetname & \textbf{12.6}   & 16.9            & \textbf{37.5}  & \textbf{59.0}  \\ \hline
\end{tabular}%
}
\caption{\textbf{Does fine-tuning a vision-language foundation model on \emphdatasetname cause overfitting?} Large-scale vision-language models trained on millions of images risk overfit and catastrophic forgetting of prior learned distributions when trained on small target datasets. We test this by evaluating LLaVA-7B~\cite{liu2024improved} models on COCO-Captions~\cite{chen2015microsoft}. We evaluate using the pretrained model and model fine-tuned on \emphdatasetname data. Surprisingly, the fine-tuned model does not degrade in performance and instead shows significant improvements (\textbf{\txtgreen{+18.6 ROUGE}}).}
\label{tbl:sup-vlm-finetuned-forgetting}
\end{table}

\noindent \textbf{Does fine-tuning a vision-language foundation model on \emphdatasetname cause overfitting?} In Section~\ref{sup:addl-recog-results} we asked if fine-tuned open vocabulary models forget previously learned distributions when trained on our \emphdatasetname dataset. A similar question applies to vision-language foundation models, which we investigate here. We evaluate two models on COCO-Captions~\cite{chen2015microsoft}, a dataset that provides captions for many real-world images.  To test our hypothesis, we evaluate two models (a) a pretrained LLaVA-1.5-7B~\cite{liu2024improved} model (b) a LLaVA-1.5-7B additionally fine-tuned on the \emphdatasetname dataset. The input prompt to both the model is \textit{``Describe the given image in detail.''} Our findings are reported in Table~\ref{tbl:sup-vlm-finetuned-forgetting}. Surprisingly, contrary to our expectations, the fine-tuned model performs better on almost all metrics. Further analysis suggests that the captioning style of COCO-Captions~\cite{chen2015microsoft} validation set is more similar to the \textit{terse} and \textit{direct} scene descriptions of the \emphdatasetname dataset, whereas the original LLaVA-1.5-7B~\cite{liu2024improved} training data is more \textit{detailed} and \textit{flowery}. We hypothesize that fine-tuning on our \emphdatasetname dataset improved model alignment for captioning tasks. We leave further investigation to future work.



\begin{table*}[]
\centering
\resizebox{0.8\textwidth}{!}{%
\begin{tabular}{lcccccccc}
\hline
\textbf{Method} &
  \multicolumn{2}{c}{All Paths} &
  \multicolumn{2}{c}{Low Curvature} &
  \multicolumn{2}{c}{Medium Curvature} &
  \multicolumn{2}{c}{High Curvature} \\ \hline
 &
  ADE &
  FDE &
  ADE &
  FDE &
  ADE &
  FDE &
  ADE &
  FDE \\ \hline
YNet~\cite{mangalam2021goals} w/ Pretrained Segm.~\cite{cordts2016cityscapes} &
  31.28 &
  102.7 &
  28.28 &
  95.92 &
  29.84 &
  102.39 &
  40.76 &
  113.38 \\
\rowcolor[HTML]{ECF4FF} 
YNet~\cite{mangalam2021goals} w/ \emphdatasetname Segm. &
  \textbf{22.68} &
  \textbf{80.78} &
  \textbf{22.41} &
  \textbf{75.33} &
  \textbf{21.82} &
  \textbf{83.28} &
  \textbf{30.21} &
  \textbf{84.58} \\ \hline
\end{tabular}%
}
\caption{\textbf{Pathway Prediction in Images.} We employ YNet~\cite{mangalam2021goals} with a segmentation model trained on Cityscapes~\cite{cordts2016cityscapes} as our baseline, and train a segmentation model with \emphdatasetname dataset, and we observe that work zone object segmentations improve pathway and goal predictions. Displacement Error (ADE) and Final Displacement Error (FDE) captures the error of predicted pathway and goal from the ground truth pathway and goal respectively. We also report results for different thresholds of average curvatures, hypothesizing that it is more difficult to navigate workzones where pathways are more irregular. We do observe that displacement errors of both predicted pathway and goal is higher at the higher curvature threshold.}
\label{tbl:pathways-expt}
\end{table*}

\subsection{Driving through Work Zones}
\label{sup:addl-driving-results}

\noindent
\textbf{Metric $AE\% < \theta$ vs Pixel level metrics~\cite{mangalam2021goals}.} ~\cite{mangalam2021goals} presents results on pixel level metrics like Average Displacement Error (ADE) and Final Displacement Error (FDE), we also report those metrics. However, we believe $AE\% < \theta$ measures model performance more fairly in autonomous driving situations. This is because pixel level metrics like average displacement error~\cite{mangalam2021goals} but do not account for camera's field of view in the ego-car's viewpoint.  We have access to the camera instrinsics $K$ and the angular error is computed by finding the angle between ground truth point $p$ and predicted point $\hat{p}$ in pixel coordinates,

\begin{equation*}
    AE(p, \hat{p}) = \cos^{-1} \left( \frac{ (K^{-1}p) \cdot (K^{-1}\hat{p}) }{\norm{K^{-1}p}\norm{K^{-1}\hat{p}}} \right)
\end{equation*}

\noindent
Now, we define $AE\% < \theta$ as the percentage of predictions whose angular error is within a threshold $\theta$. Do note horizontal field of view of our images are around $\ang{50}$.

\noindent
\textbf{Pathway Prediction Results with Pixel Level metrics.}   Nevertheless, like~\cite{mangalam2021goals}, we also report pixel level displacement metrics. Table~\ref{tbl:pathways-expt} shows goal and pathway results, Final Displacement Error (FDE) is the pixel error between the predicted goal and the actual goal while Average Displacement Error (ADE) is the error between predicted pathway and actual pathway. We observe that \emphdatasetname improves  both FDE (\textbf{\txtgreen{-21.3\%}}) and ADE (\textbf{\txtgreen{-27.5\%}}). We also bin the pathways in terms of curvature, and observe that paths with higher curvature are difficult to predict, however, model trained on \emphdatasetname dataset improves FDE (\textbf{\txtgreen{-25.4\%}}) and ADE (\textbf{\txtgreen{-25.9\%}}) in those cases. 

\begin{figure*}[t!]
    \centering
    \includegraphics[width=1\textwidth]{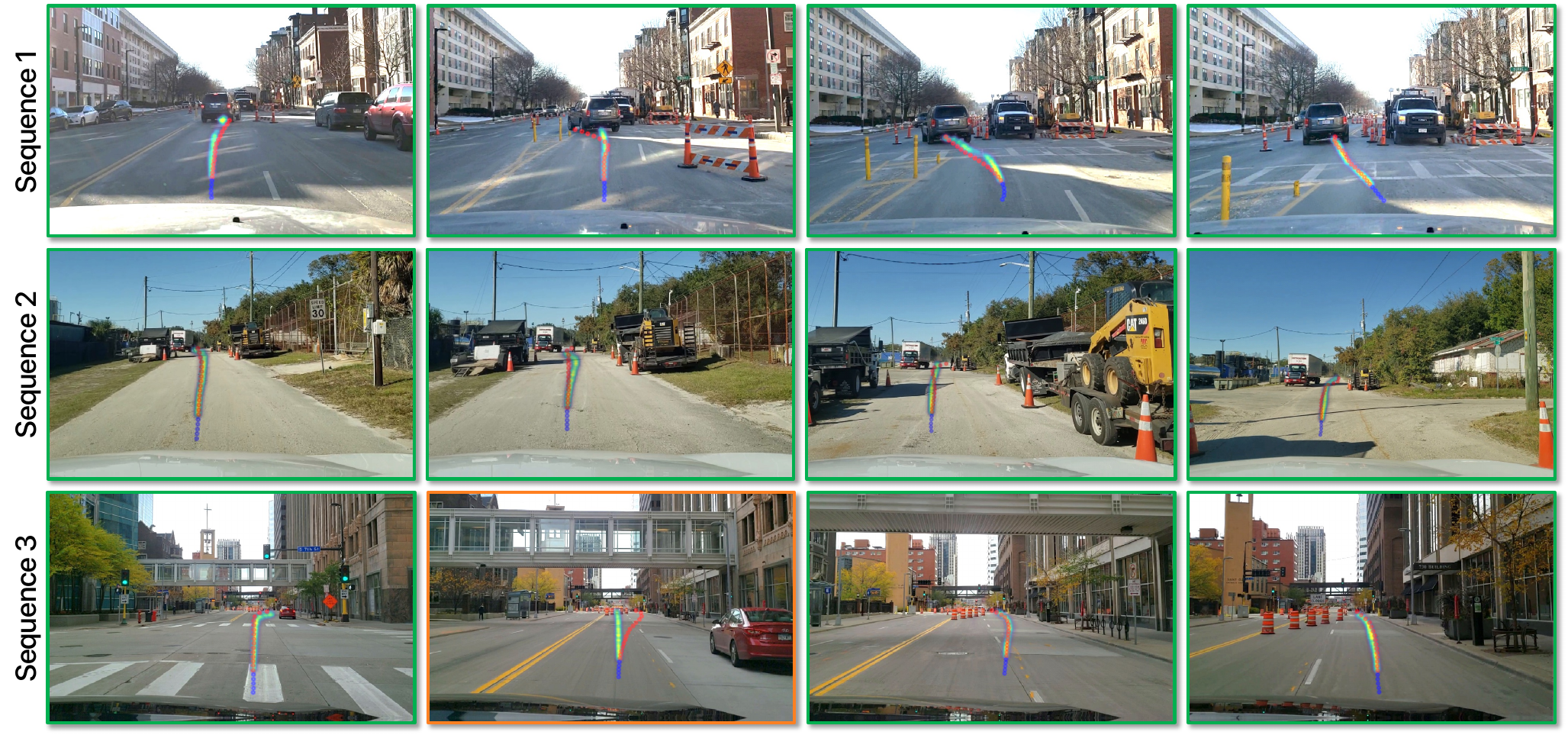}
    \caption{\textbf{Pathway Prediction for Work Zone Image Sequences.} We show examples of trajectory heatmaps predicted by YNet~\cite{mangalam2021goals} for video sequences. Input to the model is the image and \textbf{\textcolor{blue(ryb)}{observed pathway}}, also shown is the \textbf{\textcolor{cadmiumred}{future pathway}} (computed from actual driving). 
    Frames are outlined indicating \textbf{\textcolor{ourmediumgreen}{plausible pathway heatmap}} and \textbf{\txtorange{colliding pathway heatmap}}. \textit{(Sequence 1)} Following vehicles is a  learned cue. \textit{(Sequence 2)} Exploiting available free space is also learned. \textit{(Sequence 3)} Even if the initial goal is plausible, model predicts an unsafe trajectory that would collide with work zone objects. Later, model course-corrects the trajectory when closer to work zone objects.
    }
    \label{fig:pathway-video}
\end{figure*}

\noindent
\textbf{Visual Results} Figure~\ref{fig:pathway-video} shows predictions in a sequence -- we observe that the trajectory heatmap is dynamic and stochastic employing different scene level cues while forecasting trajectories (such as locations of other vehicles navigating the same work zone or the available free space in the work zone). Figure~\ref{fig:pathway-bad} shows some of the failure cases. For instance, Figure~\ref{fig:pathway-bad} (c) shows predicting multiple goals is difficult and the models fails at an intersection. 

\section{Implementation Details}

\noindent
\textbf{Detecting Work Zone Objects.} We employ the pre-trained open vocabulary models~\cite{xu2023open, zhou2022detecting, liu2024grounding} as is and follow their custom vocabulary protocol. For training Mask R-CNN~\cite{he2017mask}, we use the mmdetection~\cite{mmdetection} library initialized with COCO~\cite{lin2014microsoft} weights. We use the default model zoo parameters with the 1x schedule. For DINO~\cite{zhangdino}, Mask DINO~\cite{li2023maskdino} and DiffusionDet~\cite{chen2023diffusiondet}, we use their official codebases and weights. We employ the simple copy-paste implementation from mmdetection~\cite{mmdetection} and use the default parameters.

\noindent
\textbf{Adapting to New Geographies} We follow the same pre-train and adaptation protocol described in 2PCNet~\cite{kennerley20232pcnet}.

\noindent
\textbf{Generating Work Zone Descriptions.} To circumvent memory constraints, we train models via low rank adaptation (LORA)~\cite{hu2022lora}. We also hypothesized utilzing object predictions as context would improve description quality. We compose the coarse vocabulary work zone object detector (from Section 3 in the main paper) to align our descriptions. We employ rank $R = 128$ and alpha $\alpha = 256$ while performing LORA fine-tuning on LLaVA-7B~\cite{liu2023llava} for 4 epochs. We keep the rest of the parameters as is, following LLaVA's training schedule. The model prompt to generate descriptions is \textit{``You are the planner of an autonomous vehicle, ONLY describe the workzone in the scene identifying and describing the spatial relationship of relevant objects to plan and navigate a route''}.  While training with additional object context, we use ground truth to append a programmatic prompt for each object -- \textit{``(\texttt{object\_category}: \texttt{confidence}) at \texttt{[(x1, y1), (x2, y2)]}''}. While testing with additional object context, we use detector predictions.

\begin{figure*}
    \centering
    \includegraphics[width=\linewidth]{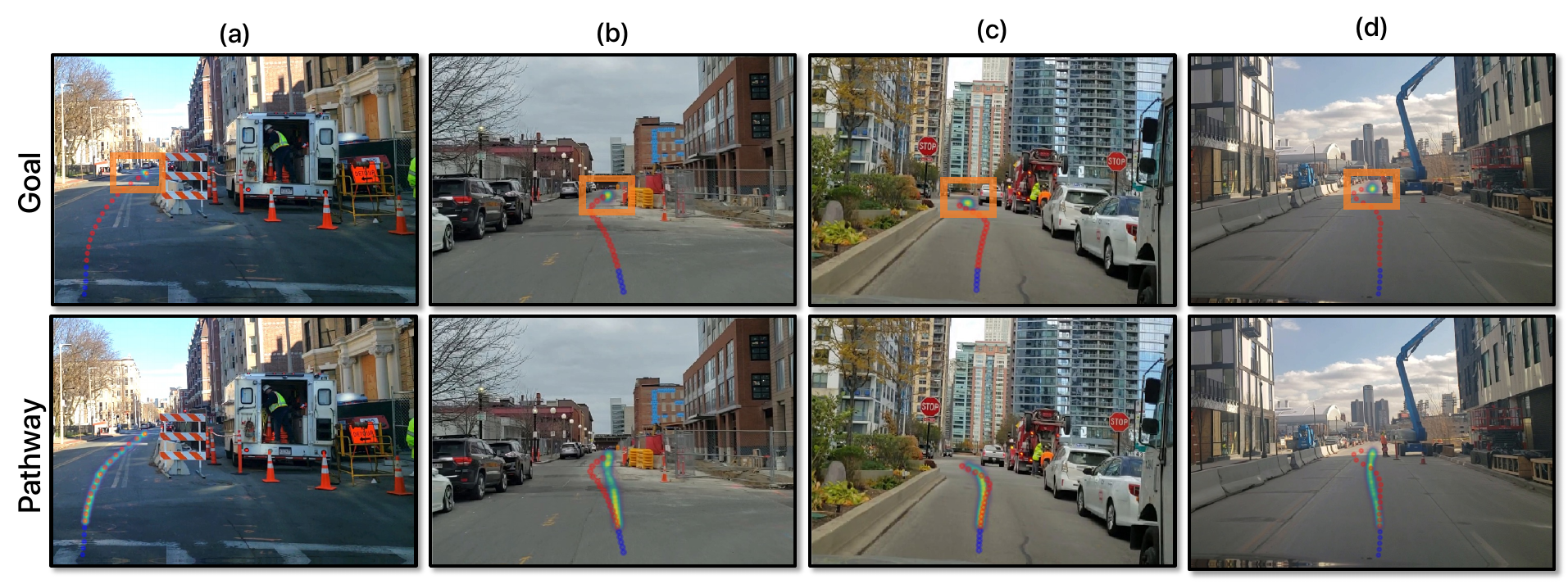}
    \caption{\textbf{Pathway Prediction in Work Zone Images.} We show examples of goal and trajectory heatmaps predicted by YNet~\cite{mangalam2021goals}. Input to the model is the image and \textbf{\textcolor{blue(ryb)}{observed pathway}}, also shown is the \textbf{\textcolor{cadmiumred}{future pathway}} (both computed from actual driving videos). Top row shows the predicted goal heatmaps while the bottom row shows the predicted pathway heatmaps, conditioned on a sampled goal. We observe that the predicted goal heatmap (marked with an \textbf{\txtorange{orange box}} for clarity) is close to the ground truth goal, and the predicted pathway is plausible.}
    \label{fig:pathway-good-extended}
    \vspace{-0.2in}    
\end{figure*}

\begin{figure*}[t!]
    \centering
    \includegraphics[width=1\textwidth]{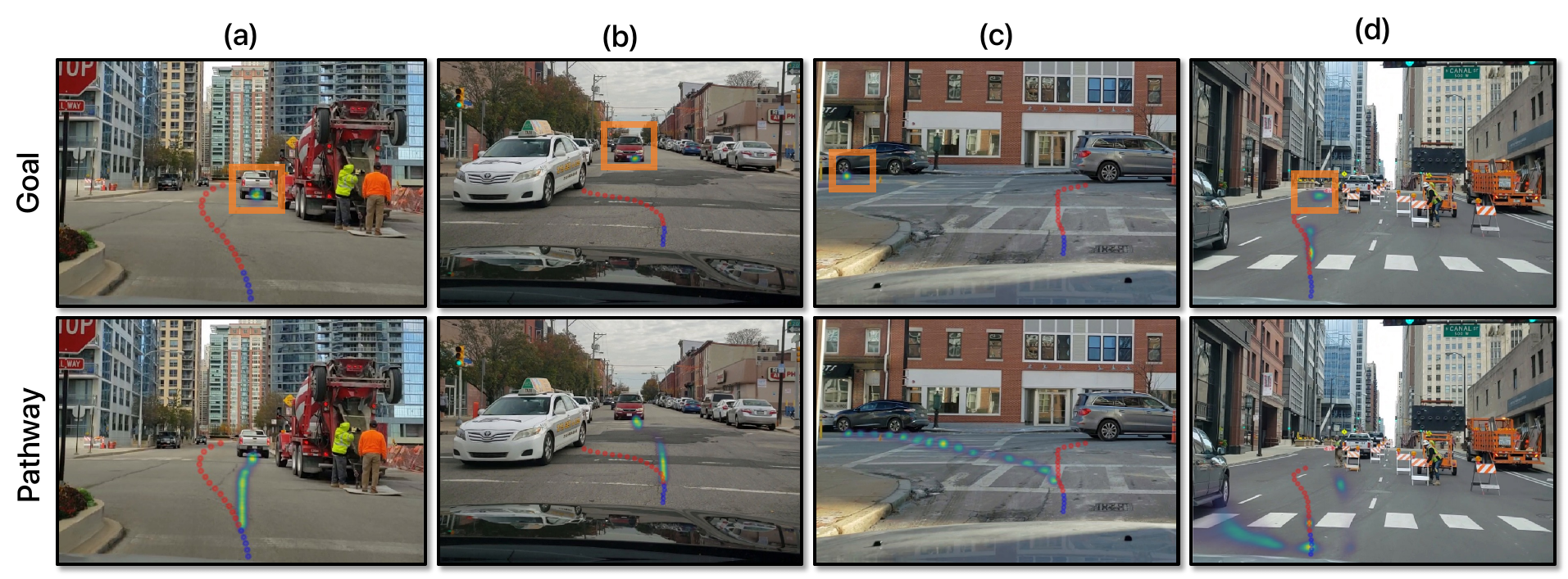}
    \caption{\textbf{Pathway Prediction in Work Zone Images: Failure cases.} We show examples of goal and trajectory heatmaps predicted by YNet~\cite{mangalam2021goals} where the model fails. Input to the model is the image and \textbf{\textcolor{blue(ryb)}{observed pathway}}, also shown is the \textbf{\textcolor{cadmiumred}{future pathway}} (both computed from actual driving). Top row shows the predicted goal heatmaps (marked with an \textbf{\txtorange{orange box}} for clarity) while the bottom row shows the predicted pathway heatmaps, conditioned on a sampled goal. We observe, (a-b) the model selects vehicles in front as goal without considering global semantics. (c) Modelling multimodality of goals is a challenge, model is unable to predict all goals at an intersection. (d) Even if the goal is valid, the pathway prediction fails for heavily blocked work zones.}
    \label{fig:pathway-bad}
    \vspace{-0.15in}
\end{figure*}

\clearpage

\begin{table*}
\centering
\caption{Description and Examples of Roadwork Objects in \emphdatasetname Dataset.}
\begin{tabular}{r p{7cm} P{7cm}}
\hline
\textbf{Object Name} & \textbf{\thead{Description}} & \textbf{Examples} \\ \hline
Cone & 
A cone shaped marker. Usually orange in color, but may be yellow, lime green, blue, red, pink or white. One or more white or retro-reflective collars around the top. May have four flat sides instead of a cone shape. & 
\vspace{-0.2cm} \includegraphics[width=5.cm]{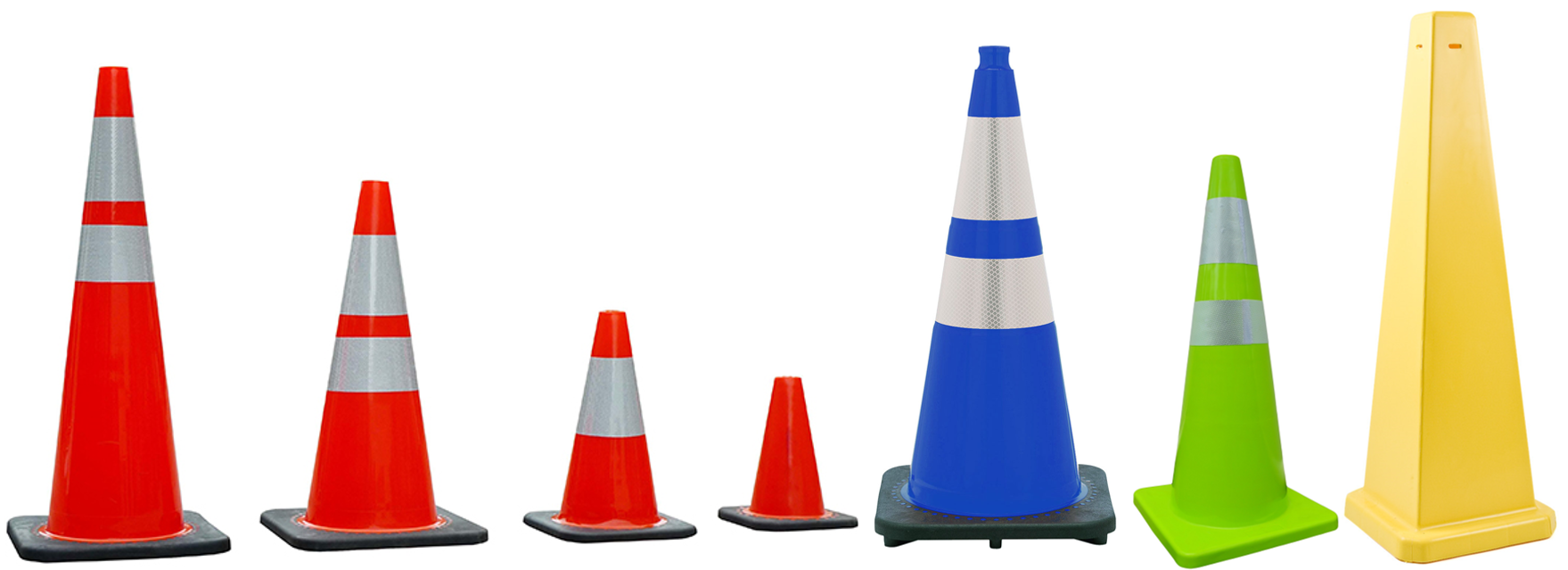}            \\
Vertical Panel &
Rectangular shaped marker. Orange or white with alternating orange and white retro-reflective stripes sloping at an angle. May have text over downward sloping stripes or text and graphics instead of downward sloping stripes. May have light on top. &
\vspace{-0.2cm} \includegraphics[width=5.cm]{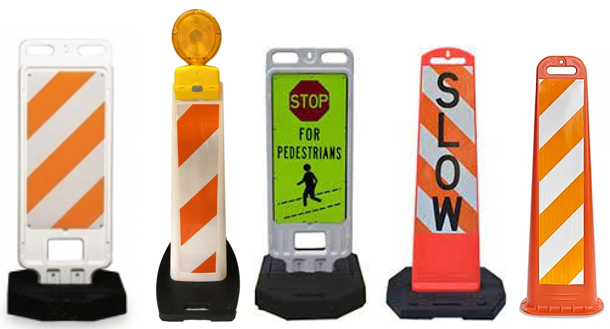}            \\
Tubular Marker &
Long and round tube shaped markers. Predominately orange in color. Typically white or green when used for protected bike lanes. Top may have white or retro-reflective bands on top. Top may become flattened or a loop. &         
\vspace{-0.2cm} \includegraphics[width=5.cm]{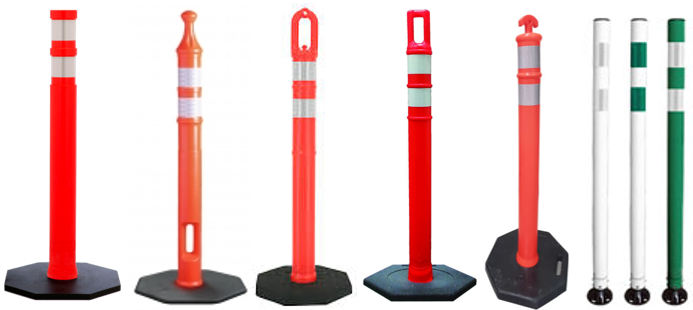}  \\
Work Vehicle         & 
Heavy duty and light duty vehicles that are driven and operated in order to perform roadwork related functions. Also includes traffic control vehicles and passengers vehicles that may be modified for use on the road and in work zones.  & 
 \vspace{-0.0cm} \includegraphics[width=5.5cm]{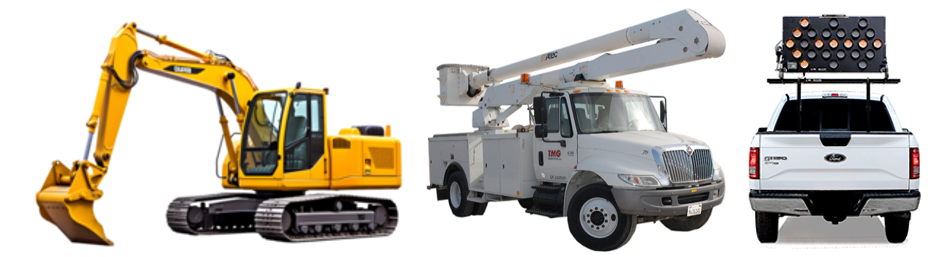}  \\
TTC Sign             &
Placed temporarily in and around work zones to increase motorist and pedestrian awareness and provide information about work zones. Usually orange, but can also be white or yellow. &
\vspace{-0.2cm} \includegraphics[width=5.5cm]{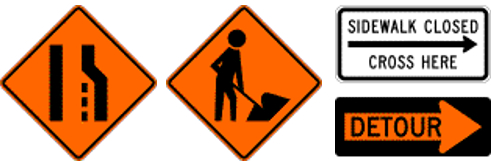}  \\
Drum                 & Bright orange cylindrical object with horizontal retro-reflective orange and white stripes around the circumference. May have a warning light or a temporary traffic control sign mounted on top. &
\vspace{-0.2cm} \includegraphics[width=5.0cm]{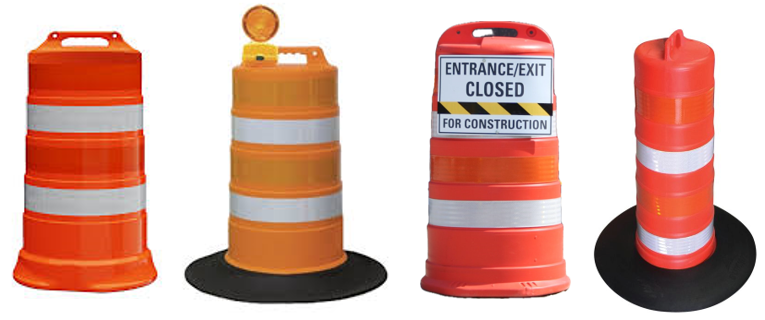} \\
&
&
\emph{Table continued on following page.}\\
 \hline
\label{tbl:supp-object-descrip}
\end{tabular}
\end{table*}

\begin{table*}
\centering
\caption*{Table \ref{tbl:supp-object-descrip}. Description and Examples of Roadwork Objects in \emphdatasetname Dataset, Continued.}
\begin{tabular}{r p{7cm} P{7cm}}
\hline
\textbf{Object Name} & \textbf{\thead{Description}} & \textbf{Examples} \\ \hline
Barricade            &                                          
Marker often used to indicate road or sidewalk closer or used as a channeling device. Consists of one to three horizontal boards with alternating orange and white retro-reflective stripes sloping at an angle. Single board barricades, commonly referred to as saw horse or roadblock horse, are often painted in a single color when used by local municipalities and police departments. May have a mounted warning light and/or temporary traffic control sign. &                   
\vspace{-0.2cm} \includegraphics[width=4.5cm]{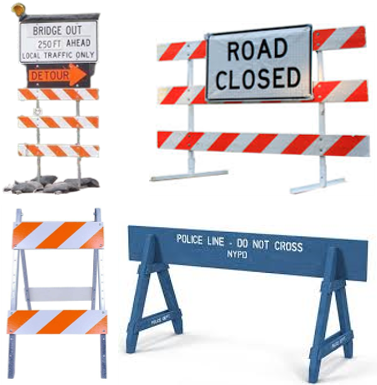} \\
Barrier              &
Longitudinal channeling device used as a temporary traffic control device for merging traffic, closing roads, and to provide guidance and warning. Also used to protect workers in a work zone. Made of concrete, plastic, or metal. May be solid (e.g., concrete barriers on highway median) or have open vertical space.&  
\vspace{-0.2cm} \includegraphics[width=4.25cm]{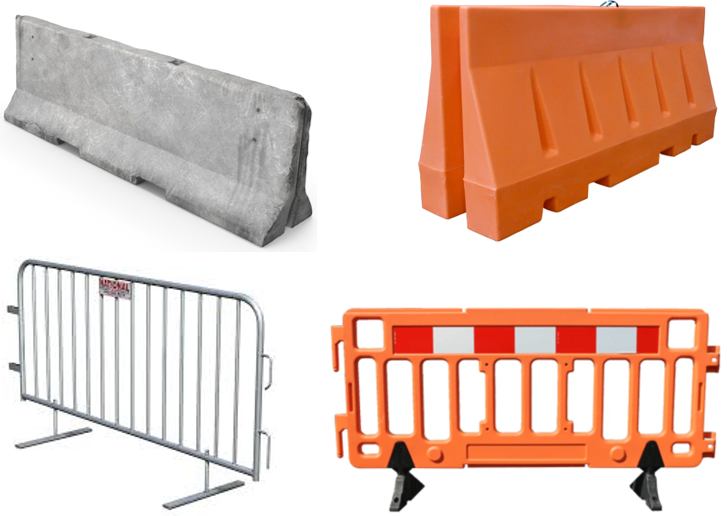} \\           
Worker               &                                          
People that performing duties related to their job in the road environment. Workers may be within a confined roadwork zone or in the area outside of a work zone. Workers may be operating or inside of a vehicle. Usually identifiable by a high visibility vest and hard hat.
&
\vspace{-0.2cm} \includegraphics[width=5.0cm]{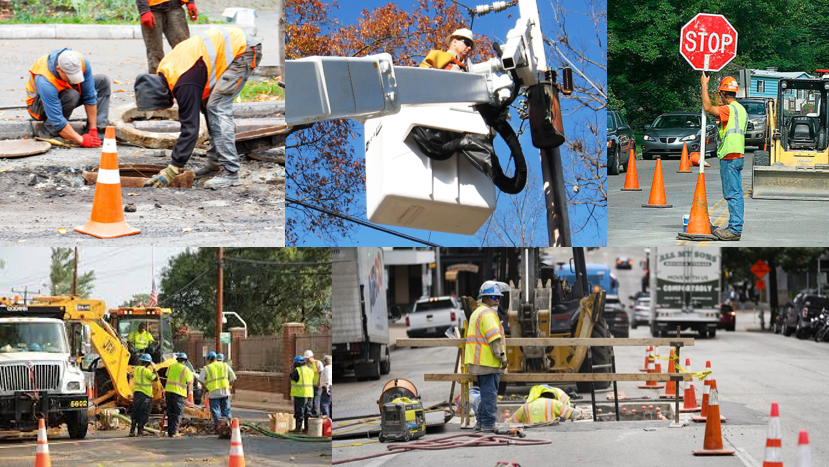} \\
Fence                &
Temporary structure used around a work zone. Usually a temporary chain link fence or safety fence (usually orange). Chain link fence may have privacy screen and mounted on top of a barrier.&        
\vspace{0.0cm} \includegraphics[width=5.5cm]{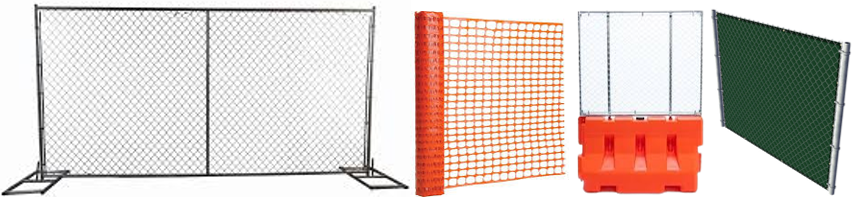} \\
Work Equipment       &
Broadly encapsulates equipment (not including work vehicles) commonly found in roadwork zones. Includes manual and power equipment whether. May be actively in use by worker. & 
\vspace{-0.1cm} \includegraphics[width=5.5cm]{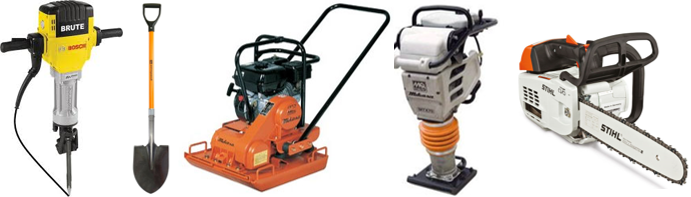} \\ 
&
&
\emph{Table continued on following page.}\\
\hline
\end{tabular}
\end{table*}

\begin{table*}
\centering
\caption*{Table \ref{tbl:supp-object-descrip}. Description and Examples of Roadwork Objects in \emphdatasetname Dataset, Continued.}
\begin{tabular}{r p{7cm} P{7cm}}
\hline
\textbf{Object Name} & \textbf{\thead{Description}} & \textbf{Examples} \\ \hline
Arrow Board          &
Digital sign with a matrix of elements capable of displaying static, sequential, or flashing arrows used for providing warning and directional information to assist with merging and directing road users through or around roadwork zone. Usually on a dedicated trailer or may be mounted on a vehicle. &            
\vspace{0.5cm} \includegraphics[width=5.5cm]{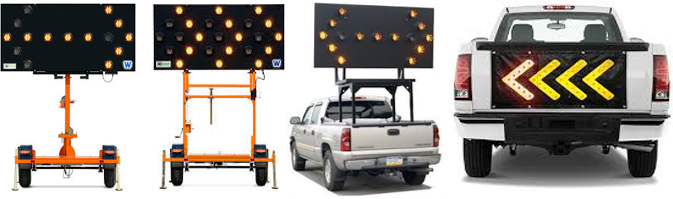} \\
TTC Message Board        &
Digital sign with the flexibility of displaying static, sequential, or flashing messages and symbols. Primarily used to advise road users of unexpected situations, displaying real-time information, and providing information to assist in decision making. Usually on a dedicated trailer or may be mounted on a vehicle. &                   
\vspace{0.2cm} \includegraphics[width=5.5cm]{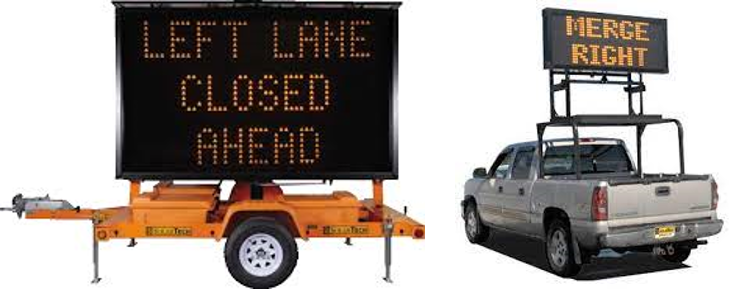} \\
Police Vehicle       &
A vehicle used by police and law enforcement to respond to service calls. Usually a sedan, sports utility vehicle, or pick-up truck fitted with a light bar. Paint color and markings vary between states and municipalities. &
\vspace{0.2cm} \includegraphics[width=5.5cm]{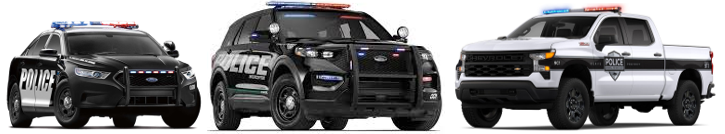} \\
Police Officer       &
Uniformed officers are often in the area of work zones to help manage traffic around work sites. May be wearing high visibility vest or safety sash belt. &                  
\vspace{-0.2cm} \includegraphics[width=5.5cm]{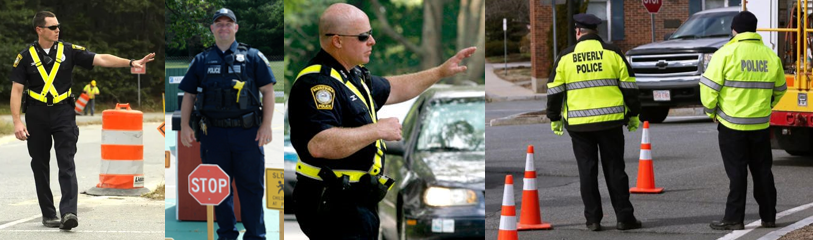} \\
\hline
\end{tabular}
\end{table*}

\clearpage

\begin{table*}[!h]
\centering
\caption{Number of annotated object instances for each city in the dataset. The categories are ordered by their totals in all the cities.}
\begin{tabular}{lrrrrrrrrrr}
\hline
                           & {\small \textbf{\makecell{Bike \\ Lane}}} & {\small \textbf{\makecell{Other \\ Roadwork \\ Objects}}} & {\small \textbf{\makecell{Police \\ Officer}}} & {\small \textbf{\makecell{Police \\ Vehicle}}} & {\small \textbf{\makecell{TTC \\ Message \\ Board}}} & {\small \textbf{\makecell{Arrow \\ Board}}} & {\small \textbf{\makecell{Work \\ Equipment}}} & {\small \textbf{Fence}} & {\small \textbf{Worker}} & {\small \textbf{Total}} \\
                           \hline
{\small \textbf{Boston, MA}}        & 16        & 27                     & 27             & 21             & 18                & 15          & 108            & 268   & 263    & 763   \\
{\small \textbf{Charlotte, NC}}     & 0         & 0                      & 9              & 12             & 10                & 11          & 29             & 113   & 210    & 394   \\
{\small \textbf{Chicago, IL}}       & 0         & 0                      & 3              & 3              & 12                & 26          & 16             & 22    & 106    & 188   \\
{\small \textbf{Columbus, OH}}      & 0         & 0                      & 3              & 6              & 5                 & 15          & 19             & 77    & 79     & 204   \\
{\small \textbf{Denver, CO}}        & 40        & 13                     & 5              & 3              & 9                 & 105         & 29             & 223   & 157    & 584   \\
{\small \textbf{Detroit, MI}}       & 0         & 0                      & 1              & 2              & 9                 & 95          & 38             & 381   & 113    & 639   \\
{\small \textbf{Houston, TX}}       & 0         & 0                      & 0              & 3              & 11                & 23          & 12             & 38    & 63     & 150   \\
{\small \textbf{Indianapolis, IN}}  & 0         & 0                      & 0              & 6              & 7                 & 16          & 37             & 64    & 57     & 187   \\
{\small \textbf{Jacksonville, FL}}  & 0         & 0                      & 0              & 1              & 0                 & 3           & 6              & 10    & 36     & 56    \\
{\small \textbf{Los Angeles, CA}}   & 0         & 0                      & 6              & 8              & 19                & 96          & 26             & 203   & 433    & 791   \\
{\small \textbf{Minneapolis, MN}}   & 0         & 0                      & 2              & 2              & 2                 & 0           & 20             & 69    & 38     & 133   \\
{\small \textbf{New York City, NY}} & 0         & 0                      & 7              & 3              & 1                 & 6           & 116            & 59    & 126    & 318   \\
{\small \textbf{Philadelphia, PA}}  & 0         & 0                      & 3              & 12             & 4                 & 12          & 38             & 117   & 155    & 341   \\
{\small \textbf{Phoenix, AZ}}       & 0         & 0                      & 1              & 0              & 0                 & 6           & 10             & 39    & 57     & 113   \\
{\small \textbf{Pittsburgh, PA}}            & 40        & 49                     & 22             & 22             & 83                & 220         & 308            & 707   & 930    & 2381  \\
{\small \textbf{San Antonio, TX}}   & 2         & 3                      & 42             & 24             & 7                 & 10          & 7              & 174   & 350    & 619   \\
{\small \textbf{San Francisco, CA}} & 0         & 0                      & 3              & 2              & 6                 & 39          & 23             & 79    & 251    & 403   \\
{\small \textbf{Washington, DC}} & 0         & 0                      & 6              & 20             & 34                & 58          & 72             & 136   & 178    & 504   \\ \hline
{\small \textbf{Total}}             & 98        & 106                    & 143            & 150            & 266               & 831         & 973            & 3171  & 4060   &  9798     \\ \hline
\end{tabular}
\label{tbl:supp-distribution}
\end{table*}

\begin{table*}[!h]
\centering
\begin{tabular}{lrrrrrrrrrr}
\hline
                            & \textbf{Barrier} & \textbf{Barricade} & \textbf{Drum} & \textbf{TTC Sign} & \textbf{\makecell{Work \\ Vehicle}} & \textbf{\makecell{Tubular \\ Marker}} & \textbf{\makecell{Vertical \\ Panel}} & \textbf{Cone} & \textbf{Total} \\  \hline
\textbf{Boston, MA}        & 568              & 169                & 594           & 225               & 845                   & 3591                    & 13                      & 1565          & 7570  \\
\textbf{Charlotte, NC}     & 98               & 117                & 448           & 155               & 334                   & 714                     & 0                       & 757           & 2623  \\
\textbf{Chicago, IL}       & 35               & 178                & 194           & 44                & 172                   & 40                      & 0                       & 386           & 1049  \\
\textbf{Columbus, OH}      & 65               & 106                & 430           & 193               & 131                   & 256                     & 68                      & 314           & 1563  \\
\textbf{Denver, CO}        & 149              & 186                & 183           & 429               & 355                   & 487                     & 1004                    & 1663          & 4456  \\
\textbf{Detroit, MI}       & 317              & 227                & 1042          & 206               & 564                   & 304                     & 1                       & 231           & 2892  \\
\textbf{Houston, TX}       & 78               & 95                 & 598           & 169               & 126                   & 57                      & 19                      & 111           & 1253  \\
\textbf{Indianapolis, IN}  & 58               & 19                 & 194           & 52                & 120                   & 43                      & 1                       & 344           & 831   \\
\textbf{Jacksonville, FL}  & 10               & 29                 & 15            & 31                & 57                    & 3                       & 0                       & 175           & 320   \\
\textbf{Los Angeles, CA}   & 237              & 768                & 23            & 703               & 864                   & 128                     & 11                      & 584           & 3318  \\
\textbf{Minneapolis, MN}   & 118              & 214                & 364           & 248               & 108                   & 409                     & 1                       & 195           & 1657  \\
\textbf{New York City, NY} & 119              & 60                 & 115           & 71                & 190                   & 49                      & 10                      & 448           & 1062  \\
\textbf{Philadelphia, PA}  & 184              & 136                & 396           & 187               & 299                   & 21                      & 0                       & 695           & 1918  \\
\textbf{Phoenix, AZ}       & 48               & 75                 & 0             & 95                & 58                    & 3                       & 245                     & 81            & 605   \\
\textbf{Pittsburgh, PA}            & 1214             & 1857               & 471           & 4071              & 2265                  & 1328                    & 6906                    & 6332          & 24444 \\
\textbf{San Antonio, TX}   & 77               & 224                & 915           & 485               & 400                   & 491                     & 218                     & 1202          & 4012  \\
\textbf{San Francisco, CA} & 171              & 133                & 2             & 124               & 419                   & 211                     & 0                       & 1326          & 2386  \\
\textbf{Washington, DC} & 250              & 24                 & 606           & 214               & 248                   & 49                      & 4                       & 1432          & 2827  \\ \hline
\textbf{Total}             & 4137             & 5167               & 6632          & 8242              & 8258                  & 8859                    & 10725                   & 20261         &  72281      \\ \hline
\end{tabular}
\end{table*}

\end{document}